\documentclass{article}

\usepackage[preprint]{neurips_2025}

\usepackage{textcomp}  
\usepackage{marvosym}  

\usepackage[utf8]{inputenc} 
\usepackage[T1]{fontenc}    
\usepackage{hyperref}       
\usepackage{url}            
\usepackage{booktabs}       
\usepackage{pdfpages}      
\usepackage{amsfonts}       
\usepackage{nicefrac}       
\usepackage{microtype}      
\usepackage{xcolor}         
\usepackage{multirow}
\usepackage{graphicx}
\usepackage{adjustbox}
\usepackage{amsmath}
\usepackage[normalem]{ulem}
\useunder{\uline}{\ul}{}
\usepackage{wrapfig}
\usepackage{multirow}     
\usepackage{colortbl}     
\usepackage{graphicx}     
\usepackage{booktabs}     
\usepackage{xcolor}       
\usepackage{caption} 
\usepackage{subcaption} 
\usepackage{amsmath}    
\usepackage{amssymb}    
\usepackage{bm}         

\title{Native Visual Understanding: Resolving Resolution Dilemmas in Vision-Language Models}

\author{
\textbf{Junbo Niu}$^{*1,2}$ ~
\textbf{Yuanhong Zheng}$^{*3 \dag}$ ~
\textbf{Ziyang Miao}$^{4 \dag}$ ~
\textbf{Hejun Dong}$^1$ \\
\textbf{Chunjiang Ge}$^5$ ~
\textbf{Hao Liang}$^{1}$ ~
\textbf{Ma Lu}$^{1}$ ~
\textbf{Bohan Zeng}$^{1}$ ~
\textbf{Qiahao Zheng}$^{1}$ ~\\
\textbf{Conghui He}$^{2}$ ~
\textbf{Wentao Zhang}$^1$\\
$^1$\ Peking University
$^2$\ Shanghai AI Laboratory \\
$^3$\ Shandong University
$^4$\ Beihang University
$^5$\ Tsinghua University
}

\begin{document}
{\let\thefootnote\relax\footnotetext{\noindent* indicates equal contribution. $\dag$ indicates intern at DCML, Peking University}}

\maketitle

\begin{abstract}

Vision-Language Models (VLMs) face significant challenges when dealing with the diverse resolutions and aspect ratios of real-world images, as most existing models rely on fixed, low-resolution inputs. While recent studies have explored integrating native resolution visual encoding to improve model performance, such efforts remain fragmented and lack a systematic framework within the open-source community. Moreover, existing benchmarks fall short in evaluating VLMs under varied visual conditions, often neglecting resolution as a critical factor. To address the \textbf{"Resolution Dilemma"} stemming from both model design and benchmark limitations, we introduce \textbf{RC-Bench}, a novel benchmark specifically designed to systematically evaluate VLM capabilities under extreme visual conditions, with an emphasis on resolution and aspect ratio variations. In conjunction, we propose \textbf{NativeRes-LLaVA}, an open-source training framework that empowers VLMs to effectively process images at their native resolutions and aspect ratios. Based on RC-Bench and NativeRes-LLaVA, we conduct comprehensive experiments on existing visual encoding strategies. The results show that \textbf{Native Resolution Visual Encoding} significantly improves the performance of VLMs on RC-Bench as well as other resolution-centric benchmarks. Code is available at \url{https://github.com/Niujunbo2002/NativeRes-LLaVA}.
\end{abstract}

\section{Introduction}
The advancement of artificial intelligence has propelled Large Language Models (LLMs) \cite{devlin2019bert, yang2024qwen2, touvron2023llama, team2023internlm, achiam2023gpt} beyond traditional language-only paradigms, striving to align with the multimodal characteristics of the real world \cite{2023llava1.6, liu2023visual, bai2025qwen2, chen2024internvl, zhang2024internlm}. However, the vast diversity of visual data, encompassing extreme aspect ratios (e.g., 16:1 panoramas, 1:8 vertical documents) and resolutions across a wide spectrum (from ultra-low thumbnail icons to ultra-high 8K medical imaging), fundamentally challenges Vision-Language Models (VLMs) in their ability to achieve human-level visual perception.

A critical component in this process is the visual encoding strategy. However, the vision encoders \cite{radford2021learning, zhai2023sigmoidlosslanguageimage} in most VLMs perceive images at a fixed resolution (e.g., $336 \times 336$ pixels \cite{radford2021learning}). Building on this, recent studies have begun investigating how to achieve dynamic resolution visual encoding using vision encoders with fixed input sizes, primarily through three approaches: 1) upsampling methods \cite{shi2024we}, 2) cropping-based methods \cite{2023llava1.6, chen2024internvl, wu2024deepseek}, and 3) hybrid vision encoders methods \cite{tong2024cambrian, li2024mini, li2025eagle} (i.e., combining high-resolution and low-resolution encoders). Furthermore, VLMs employing native resolution vision encoders \cite{dehghani2023patchnpacknavit} have also begun to emerge \cite{wang2024qwen2, team2025kimi, guo2025seed15vltechnicalreport}, offering a novel path by preserving the original resolution and aspect ratio of images. Despite the continuous emergence of new visual encoding strategies in the VLM domain, existing multimodal benchmarks often focus solely on image content, designing tasks and question-answer pairs around it. In this process, the crucial attribute of image resolution is often overlooked. These benchmarks primarily assess overall performance on generic image content (e.g., correctness of answers regarding specific image content), lacking detailed analysis of how specific image details (such as resolution distribution and aspect ratio) impact VLM answer accuracy. Nor do they adequately consider the distributional balance of resolution and aspect ratios in selected images.

Although several Native-Resolution VLMs have recently emerged \cite{wang2024qwen2, team2025kimi, guo2025seed15vltechnicalreport}, their open-source ecosystem remains significantly fragmented. These works often provide only limited fine-tuning interfaces based on frameworks like LLaMA-Factory \cite{zheng2024llamafactory} or Swift \cite{zhao2025swift}, lacking modular training architectures akin to LLaVA \cite{liu2023visual} (e.g., pluggable vision encoder replacement mechanisms, multi-stage pre-training pipelines), as shown in Table \ref{tab:openness}. This poses challenges for open-ended research in the current open-source community and academia. More critically, existing research lacks a systematic ablation framework for resolution-centric evaluation of dynamic visual encoding strategies, such as aspect ratio sensitivity and adaptability to extreme resolutions. This theoretical gap presents a dual dilemma for researchers and developers: on the one hand, academia lacks a convenient, modular native resolution training framework to support experimentation; on the other, there is a shortage of reliable benchmarks to guide the design paradigm of novel dynamic visual encoding strategies.

To resolve the aforementioned dilemmas, we first conducted a detailed analysis of existing multimodal benchmarks \cite{liu2024mmbench, fu2023mme, li2024seed2plus, yu2023mm, liu2024ocrbench} from the perspective of data distribution in terms of resolution and aspect ratio. Second, based on experimental results and manual case analysis, we categorized these benchmarks according to their sensitivity to visual encoding strategies into two main types: Semantic-Centric (SC) \cite{liu2024mmbench, fu2023mme} and Resolution-Centric (RC) \cite{liu2024ocrbench, masry2022chartqa}. RC-type benchmarks will serve as the primary evaluation subjects in our experiments. Given that most current multimodal benchmarks overlook the critical attribute of image resolution, we believe there is an urgent need to establish a new benchmark specifically tailored for evaluating visual encoding strategies, one that fully reflects the diversity of visual features. To tackle these issues, we introduce \textbf{RC-Bench}, a \textbf{R}esolution-\textbf{C}entric \textbf{Bench}mark. This benchmark is designed to thoroughly evaluate the capabilities of VLMs' visual encoding and resolution processing strategies under diverse visual conditions.

We propose \textbf{NativeRes-LLaVA}, a native visual encoding training framework capable of effectively handling images of varying resolutions and aspect ratios. By preserving the original resolution and aspect ratio of input images, this method preserves maximum image detail, thereby outperforming existing mainstream visual encoding strategies. We have open-sourced its end-to-end training system, which includes a complete toolkit for joint optimization of vision encoder-LLM, and cross-resolution generalization capability assessment, supporting elastic training and inference with pure native, controllable resolutions. 

In summary, our contributions are as follows:(1) Identified critical resolution dilemmas in the domain of VLMs. (2) Developed RC-Bench, the first benchmark to systematically evaluate visual encoding strategies under extreme visual conditions (encompassing various resolutions and aspect ratios), which guides the design paradigm of novel dynamic visual encoding strategies. (3) Proposed and open-sourced NativeRes-LLaVA, a framework enabling robust native resolution visual encoding by preserving crucial image details, demonstrating superiority over existing methods and advancing open-source capabilities in this domain. (4) We empirically validated the effectiveness of NativeRes-LLaVA across multiple benchmarks, including RC-Bench, and provided comprehensive analyses of model behavior under various visual encoding strategies, effectively resolving the resolution dilemmas in VLMs.

\vspace{-10pt}
\begin{table*}[h]
\renewcommand\arraystretch{1.3}
\centering
\small
\begin{adjustbox}{%
  max width=1\textwidth,
  max height=1\textheight,  
  center
}
\begin{tabular}{@{}l c c c c c c c c c@{}}
\hline    
\toprule
\multirow{2}{*}{\textbf{Models}} 
& \multirow{2}{*}{\begin{tabular}[l]{@{}c@{}}\textbf{NativeRes-Training } \\ \textbf{Codebase}\end{tabular}} 
& \multirow{2}{*}{\begin{tabular}[l]{@{}c@{}}\textbf{Sequence Packing} \\ \textbf{Scripts}\end{tabular}} 
& \multirow{2}{*}{\begin{tabular}[l]{@{}c@{}}\textbf{Pre-Training} \\ \textbf{Codebase}\end{tabular}} 
& \multirow{2}{*}{\begin{tabular}[l]{@{}c@{}}\textbf{Base Model} \\ \textbf{Checkpoint}\end{tabular}} 
& \multirow{2}{*}{\begin{tabular}[l]{@{}c@{}}\textbf{SFT-Training} \\ \textbf{Codebase}\end{tabular}} 
& \multirow{2}{*}{\begin{tabular}[l]{@{}c@{}}\textbf{Instruct Model} \\ \textbf{Checkpoint}\end{tabular}} 
& \multirow{2}{*}{\begin{tabular}[l]{@{}c@{}}\textbf{Flexibly Changing} \\ \textbf{Modules}\end{tabular}} 
& \multirow{2}{*}{\begin{tabular}[l]{@{}c@{}}\textbf{Resolution} \\ \textbf{Strategy}\end{tabular}} \\
    & & & & & & & & \\ 
\midrule
LLaVA \cite{liu2023visual}           & \cellcolor{gray!30} None  & \cellcolor{gray!30} None  & \cellcolor{green!30} Open  & \cellcolor{green!30} Open  & \cellcolor{green!30} Open  & \cellcolor{green!30} Open  & \cellcolor{green!30} Open  & Fixed      \\ \midrule
Cambrian-1 \cite{tong2024cambrian}   & \cellcolor{gray!30} None  & \cellcolor{gray!30} None  & \cellcolor{red!30} Closed  & \cellcolor{green!30} Open  & \cellcolor{red!30} Closed  & \cellcolor{green!30} Open  & \cellcolor{green!30} Open  & Hybrid       \\ \midrule
LLaVA-OneVision \cite{li2024llava} & \cellcolor{gray!30} None  & \cellcolor{gray!30} None  & \cellcolor{green!30} Open  & \cellcolor{green!30} Open  & \cellcolor{green!30} Open  & \cellcolor{green!30} Open  & \cellcolor{green!30} Open  & Crop       \\ \midrule
Seed1.5-VL \cite{guo2025seed15vltechnicalreport}       & \cellcolor{red!30} Closed  & \cellcolor{red!30} Closed  & \cellcolor{red!30} Closed  & \cellcolor{red!30} Closed  & \cellcolor{red!30} Closed  & \cellcolor{red!30} Closed  & \cellcolor{red!30} Closed  & Native     \\ \midrule
Kimi-VL \cite{team2025kimi}         & \cellcolor{red!30} Closed  & \cellcolor{red!30} Closed  & \cellcolor{red!30} Closed  & \cellcolor{red!30} Closed  & \cellcolor{green!30} Open  & \cellcolor{green!30} Open  & \cellcolor{red!30} Closed  & Native     \\ \midrule
Qwen2-VL \cite{wang2024qwen2}        & \cellcolor{red!30} Closed  & \cellcolor{red!30} Closed  & \cellcolor{red!30} Closed  & \cellcolor{green!30} Open  & \cellcolor{green!30} Open  & \cellcolor{green!30} Open  & \cellcolor{red!30} Closed  & Native     \\ \midrule
NativeRes-LLaVA  & \cellcolor{green!30} Open  & \cellcolor{green!30} Open  & \cellcolor{green!30} Open  & \cellcolor{green!30} Open  & \cellcolor{green!30} Open  & \cellcolor{green!30} Open  & \cellcolor{green!30} Open  & Native     \\ \midrule
\end{tabular}
\end{adjustbox}
\caption{Comparisons of openness and capabilities across different VLMs. }
\label{tab:openness}
\vspace{-20pt}
\end{table*}

\section{Related works}
\textbf{Large vision language models.}
While Large Language Models (LLMs) have achieved significant success in natural language processing (NLP), demonstrated by models like BERT \cite{koroteev2021bert} and GPT-3 \cite{floridi2020gpt}, they inherently lack the ability to process visual information. This deficiency limits their capacity for an intuitive understanding of the world. To address this, researchers leveraged advancements in visual models (e.g., ViT \cite{dosovitskiy2020image}) and training techniques (e.g., CLIP \cite{radford2021learning}) to endow LLMs with visual capabilities. A common approach involves employing an MLP layer to project visual features into the language embedding space, effectively bridging the modalities. Through extensive data collection and fine-tuning, numerous Vision-Language Models (VLMs) have been developed (e.g., LLaVA \cite{liu2023visual}, Qwen2.5-VL \cite{bai2025qwen2}, InternVL \cite{chen2024internvl}), which demonstrate strong performance on a variety of visual tasks.

\textbf{Dynamic resolution visual encoding strategy.}
With the growing demand for various fine-grained visual tasks, various strategies have emerged to improve the handling of visual input of dynamic resolution \cite{bai2025qwen2, 2023llava1.6, chen2024internvl, wu2024deepseek, guo2025seed15vltechnicalreport, team2025kimi, chen2025ocean, li2024mini, luo2024feast, li2025eagle, tong2024cambrian}. These approaches can be categorized into four primary methods: 1) Upsampling Methods, such as Qwen-VL \cite{wang2024qwen2} and S² extension \cite{shi2024we}, which increase the resolution of positional encodings and integrate multi-scale image representations. 2) Cropping-based methods, used in models like LLaVA-NeXT \cite{2023llava1.6}, InternVL \cite{chen2024internvl} and Deepseek-VL2 \cite{wu2024deepseek}, splitting the high-res image into smaller tiles and processing each tile separately with a pre-trained low-resolution vision model. 3) Hybrid vision encoders, combining high-resolution and low-resolution encoders to encode an image at different resolution \cite{li2024mini, tong2024cambrian, li2025eagle, luo2024feast}.4) Native Resolution Encoding, which directly processes dynamic-resolution images with a native-resolution ViT, as seen in models like Qwen-2VL \cite{bai2025qwen2}, Kimi-VL \cite{team2025kimi}, OceanOCR \cite{chen2025ocean}, and Seed1.5-VL \cite{guo2025seed15vltechnicalreport}. 
Furthermore, the absence of a comprehensive, resolution-centric comparison of these visual encoding strategies has led to a lack of clear guidance for research in dynamic resolution visual encoding.


\textbf{Evaluations of VLMs.}
Recent advancements in vision-language models (VLMs) have spurred the development of numerous multimodal benchmarks designed to evaluate their performance across a wide array of tasks \cite{liu2024mmbench, fu2023mme, li2024seed2plus, yu2023mm, liu2024ocrbench, mathew2021docvqa, mathew2022infographicvqa, lu2023mathvista, kembhavi2016diagram, wang2025divide, niu2025ovo}.For example, benchmarks such as the MMBench series \cite{liu2024mmbench}, SEED \cite{li2024seed2plus}, and MME \cite{fu2023mme} primarily assess the capabilities of VLMs in contextual and domain-specific understanding. In contrast, other benchmarks, including OCRBench \cite{liu2024ocrbench}, DocVQA \cite{mathew2021docvqa}, ChartQA \cite{masry2022chartqa}, and TextVQA \cite{singh2019towards}, place a greater emphasis on fine-grained image understanding, and this poses a requirement on the ability of VLMs to process high-resolution images. HR-Bench \cite{wang2025divide} takes high resolution into account and increases the image resolution to $4k$, but does not consider its aspect ratio. These benchmarks provide critical insights into how well VLMs understand the relationships between visual elements and natural language descriptions. However, existing evaluations of VLMs still suffer from many dilemmas. Notably, many current multimodal benchmarks do not fully consider the core visual complexities inherent in the real world, such as the requirement of resolution adaptation and handling of extreme aspect ratios.

\section{Resolution dilemmas}
\label{subsec:Res-dilemmas}
This section explores resolution-related challenges in the vision-language model (VLM) domain from two complementary perspectives: \textbf{Benchmarks} and \textbf{VLMs}. We then present a detailed account of the construction process for our proposed RC-Bench, with a full statistical analysis available in the Appendix\ref{sec:rcbench}.
\subsection{Analysis of benchmarks}
\label{subsec:Bench Dilemmas}
\begin{figure}[!t]
    \centering
    \vspace{-20pt}
    \includegraphics[width=\linewidth]{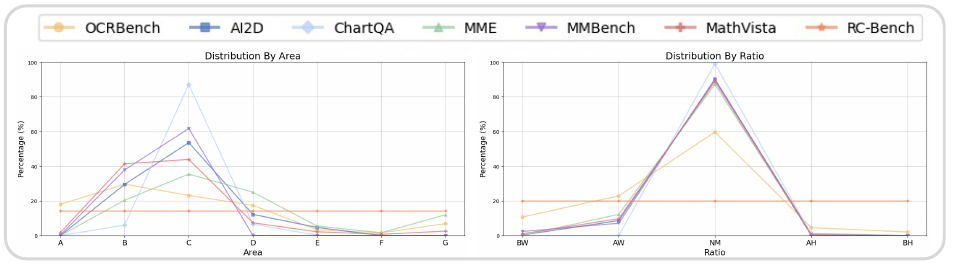}
    \caption{\footnotesize
    \textbf{Left: }The images distribution by Area(pixels). \textbf{Right: }The images distribution by Ratio.}
    \label{img: Area and Ratio}
\vspace{-20pt}
\end{figure}

\textbf{Are multimodal benchmarks truly designed with resolution in mind?} 
To delve into this dilemma, we first analyze the image resolution from the perspective of data distribution. We divide the images into seven different intervals with $384 \times 384$ pixels as the basic resolution unit, representing different resolution distributions(from small to large: A to G). Then, we analyze the aspect ratio of the image. We divide the aspect ratio (defined as the ratio of width to height) into five intervals: medium (NM: $2:1 \sim 1:2$), wide (AW: $4:1 \sim 2:1$), very wide (BW: $>4:1$), high (AH: $1:2 \sim 1:4$) and very high (BH: $<1:4$). The selected analysis results are shown in Figure \ref{img: Area and Ratio}.

Based on the above analysis, we observe that existing benchmarks\cite{liu2024ocrbench, kembhavi2016diagram, masry2022chartqa, fu2023mme, liu2024mmbench, lu2023mathvista} have limited demand for models to dynamically adjust visual perception resolution. Most benchmarks (such as MME\cite{fu2023mme}, MMBench\cite{liu2024mmbench}, etc.) are mainly composed of images with a relatively single resolution distribution. Similarly, the distribution of their image aspect ratios is also relatively concentrated. Although these benchmarks provide high-quality and challenging question-answer (QA) pairs, they are insufficient in simulating the inherent diversity of real-world visual data, especially in images covering extreme resolutions and various aspect ratios. In addition, the evaluation methods of these benchmarks also focus on the overall understanding of general image content (for example, judging whether the model's answer to the image content is correct), but lack a detailed exploration of how specific image properties (such as resolution distribution, aspect ratio) affect the accuracy of VLM responses. This limitation makes it difficult to fully evaluate the flexibility and robustness of current VLMs' visual encoding strategies.

\textbf{Are multimodal benchmarks truly designed to require a clear and detailed view of the images?}

\begin{wrapfigure}{r}{0.45\textwidth}
\vspace{-15pt}
  \centering
  \includegraphics[width=0.45\textwidth]{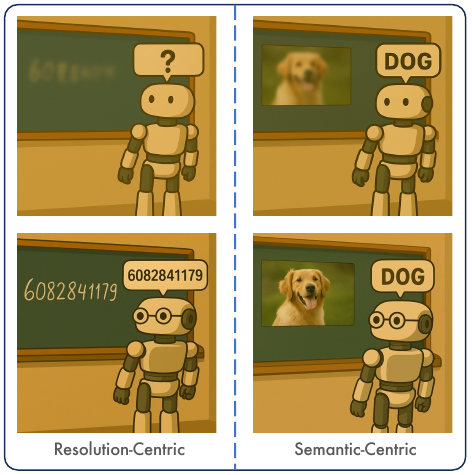}
  \caption{An illustration of the distinction between \textbf{Resolution-Centric (RC)} and \textbf{Semantic-Centric (SC)} visual tasks. RC tasks, such as reading the numbers on the left, necessitate high-resolution input. In contrast, SC tasks, such as recognizing the dog on the right, are less dependent on visual clarity and can be performed even with lower-resolution images.}
  \label{fig:RCvsSC-4o}
\vspace{-15pt}
\end{wrapfigure}

\begin{figure}[!t]
    \centering
    \vspace{-20pt}
    \includegraphics[width=\linewidth]{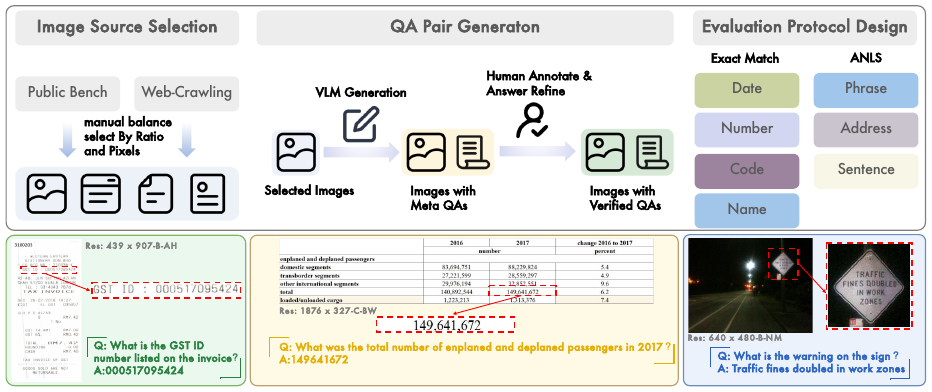}
    \caption{
    \textbf{Above: }Generation pipeline of RC-Bench. The details will be discussed in Section\ref{subsec:RC-Bench}. \textbf{Below: }Examples of three QAs in RC-Bench. More samples are shown in the Appendix \ref{sec:supp-Demo}.}
    \label{RC-Bench Pipeline}
\vspace{-10pt}
\end{figure}

To illustrate this problem, let's borrow a real-life analogy: when a mild myopic person is not wearing glasses, he can still successfully complete most daily activities (such as picking up objects, traveling, etc.) despite the blurry visual input. These tasks require relatively low visual clarity. However, when faced with tasks that require the recognition of subtle visual information (such as reading a blackboard at a distance), the lack of visual clarity will significantly affect the task performance, and glasses are needed to improve the visual resolution. This phenomenon inspires us to divide most visual tasks into two categories based on their dependence on the resolution of visual input: \textbf{Semantic-Centric (SC)} tasks and pixel-level detail \textbf{Resolution-Centric (RC)} tasks. The former, such as object recognition or direction positioning, can still be effectively completed even when the visual input is relatively blurry; the latter, such as recognizing fine text (similar to reading a blackboard in the above example), as illustrated in Figure~\ref{RC-Bench Pipeline} , is highly dependent on high-definition visual input, and its performance will significantly decrease as the clarity decreases. To aid understanding, we include an illustration in Figure \ref{fig:RCvsSC-4o}. Similarly, through manual case analysis of the benchmark, we found that the problems in the field of VLMs can also be divided into these two categories. We systematically adjusted the components of VLM \cite{liu2023visual} and resolution processing strategies to analyze the sensitivity of the selected benchmark to the input image resolution. We categorized these benchmarks according to their sensitivity to visual encoding strategies into two main types: Semantic-Centric (SC) type and Resolution-Centric (RC) type, as presented in Table \ref{Bench-Types}.

\begin{table}[h]
\vspace{-15pt}
\caption{Multimodal benchmarks for Resolution-Centric and Semantic-Centric tasks, including TextVQA (VQA$^\mathrm{T}$), OCRBench (OCR), DocVQA (VQA$^\mathrm{D}$),  ChartQA (VQA$^\mathrm{C}$), InfographicVQA (VQA$^\mathrm{I}$), HR-Bench (HR), MM-Vet (MMV), AI2D, MathVista (Math), SEED-Bench-Image (SEED), MME-Cognition (MME$^\mathrm{C}$), MME-Perception (MME$^\mathrm{P}$), MMBench-CN (MMB$^\mathrm{C}$), MMBench-EN (MMB$^\mathrm{E}$) and POPE.}
\vspace{3pt}
\resizebox{\columnwidth}{!}{%
\begin{tabular}{@{}ccccccc|cccccccc@{}}
\toprule
\multicolumn{7}{c|}{\multirow{2}{*}{\textbf{Resolution-Centric}}} & \multicolumn{8}{c}{\multirow{2}{*}{\textbf{Semantic-Centric}}} \\
\multicolumn{7}{c|}{} & \multicolumn{8}{c}{} \\
VQA$^\mathrm{T}$ & OCR & VQA$^\mathrm{D}$ & VQA$^\mathrm{C}$ & VQA$^\mathrm{I}$ & HR & MMV & AI2D & Math & SEED & MME$^\mathrm{C}$ & MME$^\mathrm{P}$ & MMB$^\mathrm{C}$ & MMB$^\mathrm{E}$ & POPE \\ \bottomrule
\end{tabular}%
}
\label{Bench-Types}
\end{table}

\begin{wrapfigure}{r}{0.45\textwidth}
\vspace{-15pt}
  \centering
  \includegraphics[width=0.45\textwidth]{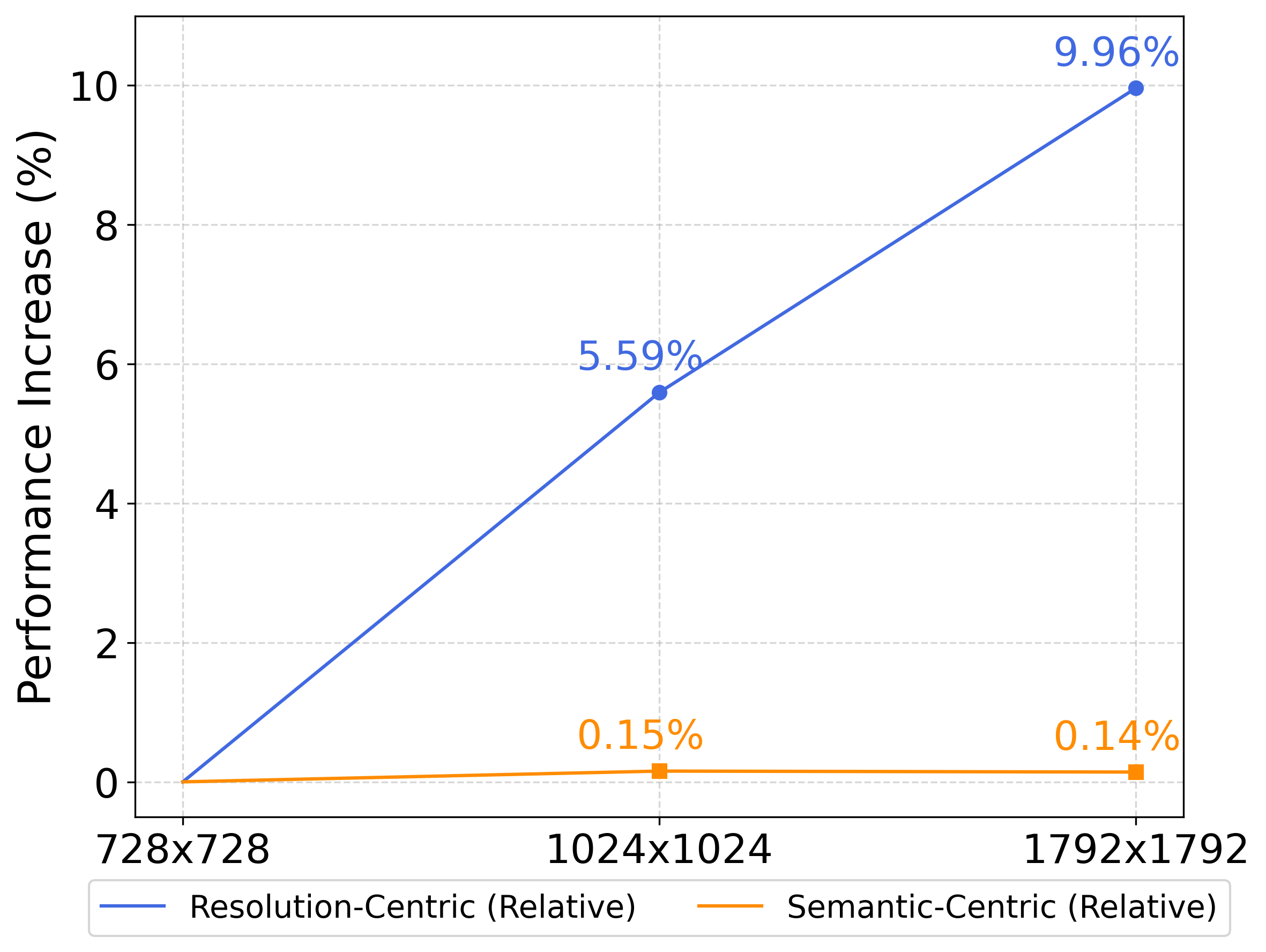}
  \caption{As the resolution increases, the performance improvement of NativeRes-LLaVA on the Resolution-Centric Benchmark is more pronounced compared to that on the Semantic-Centric Benchmarks.}
  \label{fig:RCIvsSCI}
\vspace{-15pt}
\end{wrapfigure}

As shown in Figure~\ref{fig:RCIvsSCI}, our NativeRes-LLaVA increases significantly on the Resolution-Centric type benchmarks when the input resolution increases, but remains almost unchanged on the Semantic-Centric type benchmarks. This clearly demonstrates that different benchmarks exhibit varying degrees of sensitivity to image resolution. For benchmarks that require fine-grained details, higher input image resolution leads to better performance, making native resolution support a prerequisite for the model. In contrast, for benchmarks focused on overall semantic information, input image resolution has little impact on model performance.

\subsection{Analysis of VLMs}

\textbf{How to truly achieve human-level dynamic visual perception?}
Nowadays, many methods for dynamic resolution visual encoding have emerged \cite{bai2025qwen2, 2023llava1.6, chen2024internvl, wu2024deepseek, guo2025seed15vltechnicalreport, team2025kimi, chen2025ocean, li2024mini, luo2024feast, li2025eagle, tong2024cambrian}, all sharing the common goal of realizing human-level dynamic visual perception. However, due to the challenges inherent in current multimodal benchmarks in section \ref{subsec:Bench Dilemmas}, which fail to accurately simulate the intrinsic diversity of visual data, there is a lack of quantitative evaluation on key factors such as sensitivity to aspect ratio and adaptability to extreme resolutions. 
Recently, several native-resolution vision-language models \cite{bai2025qwen2, team2025kimi, guo2025seed15vltechnicalreport, chen2025ocean} have appeared. These models adopt pure native resolution encoding strategies and have achieved state-of-the-art results across many benchmarks, bringing us a step closer to this shared goal. However, most of these models originate from major tech companies and, due to technical restrictions, their open-source ecosystems remain significantly fragmented, lacking modular training architectures like LLaVA \cite{liu2023visual}, as shown in Table \ref{tab:openness}.
More importantly, existing research has yet to establish a systematic ablation study framework to compare the advantages and disadvantages of different dynamic visual encoding strategies.
All these factors have led to the current landscape of dynamic resolution visual encoding being highly diverse and fragmented, with many competing approaches but no unified consensus.

\subsection{RC-Bench}
\label{subsec:RC-Bench}

In order to solve the above dilemma at the benchmark level, we present \textbf{RC-Bench}, a \textbf{R}esolution-\textbf{C}entric multimodal \textbf{Bench}mark that considers the core visual complexities inherent in the real world. 

\textbf{Address the challenge of imbalanced image distribution.} We constructed a dataset with enhanced uniformity in image area and aspect ratio by integrating existing public datasets \cite{fu2024ocrbenchv2, mathew2021docvqa, mathew2022infographicvqa} with proprietary data resources, as illustrated in Figure~\ref{img: Area and Ratio}. The construction process involved initially selecting high-quality images from public datasets, followed by manual verification to ensure their readability and content complexity. To address underrepresented regions, proprietary data were incorporated. Furthermore, image padding and resizing techniques were employed to augment distributions of images with extreme aspect ratios or sizes (e.g., extremely small or wide images), thereby ensuring distributional consistency between the training and evaluation phases. Ultimately, this process yielded 1750 high-quality, visually rich images, uniformly distributed across seven predefined resolution levels and five aspect ratio categories.

\textbf{Generation of question-answer pairs (QAs).} We adopted a hybrid strategy combining automated generation with manual screening. Initially, candidate QAs were generated from images using GPT-4o. These candidates were subsequently reviewed by a human annotation team to ensure that questions were unambiguous, exhibited strong visual grounding (i.e., could not be answered independently of the image). The question types primarily focused on requiring the model to recognize textual content within the images and provide accurate answers. These images spanned a diverse range of categories, including invoices, charts, natural scenes, handwritten documents, posters, and interface screenshots. The corresponding answer types included numbers, dates, names, phrases, and complete sentences.

\textbf{Evaluation methodology.} We employed distinct metrics tailored to different answer types. Specifically, for concise answers such as numbers, names, and code snippets, Exact Match (EM) was utilized. For numerical answers from images that include units, we standardized multiple equivalent expressions to accommodate human annotation preferences and enhance the robustness of the EM evaluation. For more complex answers, such as long addresses or complete sentences, partial matching was assessed using the Average Normalized Levenshtein Similarity (ANLS) metric \cite{biten2019scenetextvisualquestion}.

It is worth noting that RC-Bench differs from existing general benchmarks in two key aspects: (1) we balance the distribution of image resolutions and aspect ratios, and (2) we focus on a detailed analysis of how specific image attributes ( Resolution and Aspect ratio ) affect the accuracy of VLM responses. We provide model scores across resolution and aspect ratio dimensions to better assess the robustness of VLMs’ resolution strategies when faced with diverse visual data. More details are shown in the Appendix \ref{sec:rcbench}.

\section{Method}

In this section, we introduce NativeRes-LLaVA, a training framework for native visual encoding capable of effectively handling images of various resolutions and aspect ratios.

\subsection{Model architecture}
\begin{figure*}[!t]
    \centering
    \vspace{-5pt}
    \vspace{-1cm}
    \includegraphics[width=1.0\textwidth]{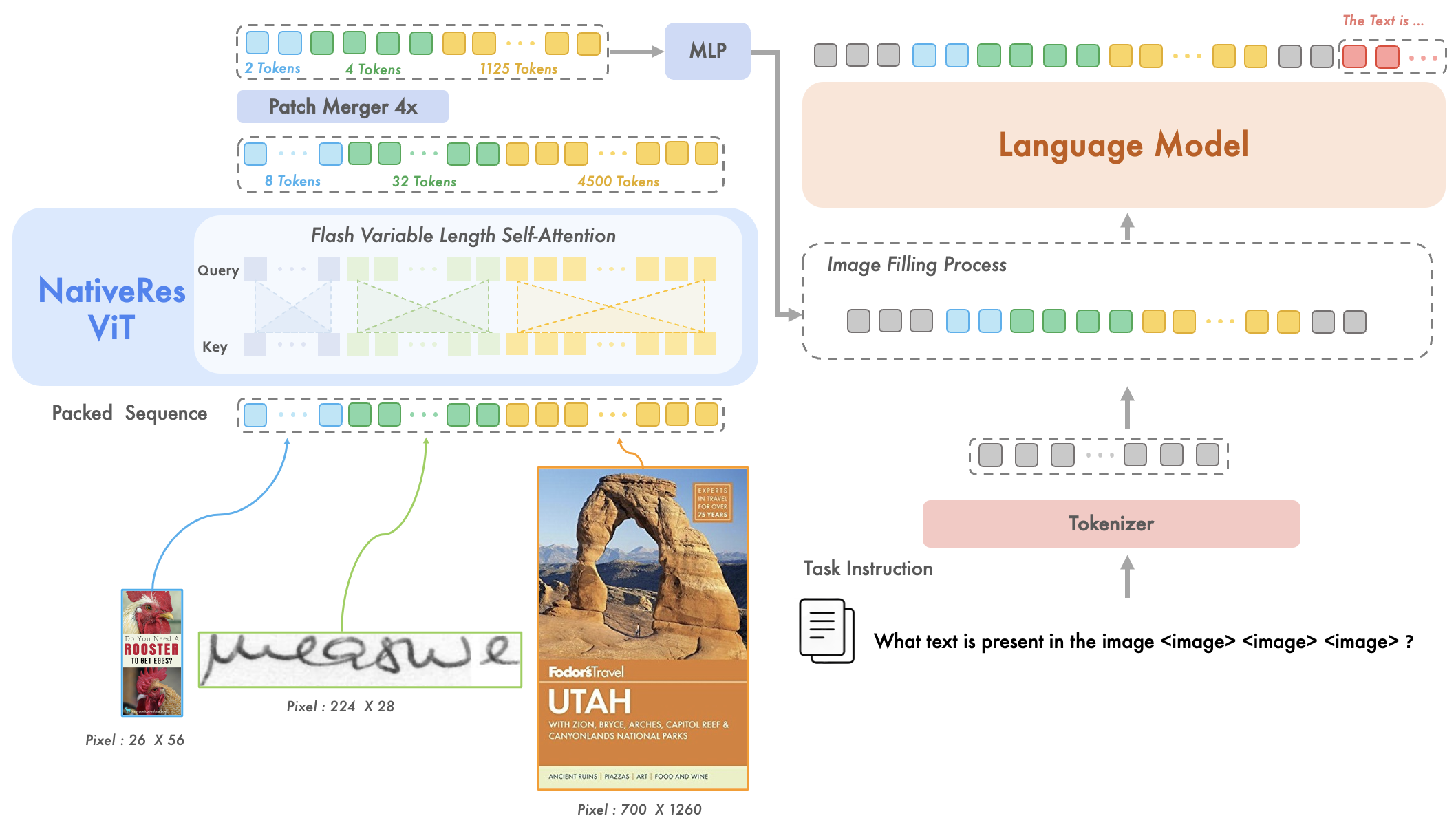}
    \vspace{-0.3cm}
    \caption{
    \textbf{The architecture of NativeRes-LLaVA.}}
    \label{fig:pipeline}
    \vspace{-25pt}
\end{figure*}

As depicted in Fig. \ref{fig:pipeline}, NativeRes-LLaVA comprises four main modules: (1) a native resolution vision encoder with 2D Rotary Position Embedding (RoPE) \cite{su2024roformer}, capable of effectively processing images with various resolutions and aspect ratios, (2) a compression module designed to further compress visual tokens from the vision encoder, (3) a two-layer Multilayer Perceptron (MLP) , and (4) an advanced LLM.


\subsection{Native-resolution visual encoding}

Inspired by Qwen2-VL \cite{wang2024qwen2}, we introduce a native-resolution encoding mechanism \cite{dehghani2023patchnpacknavit} in our proposed NativeRes-LLaVA, which can process images of arbitrary resolutions and dynamically convert them into a variable number of visual tokens. The vision encoder natively supports dynamic image resolutions and adopts 2D Rotary Positional Encoding (2D RoPE \cite{su2024roformer}) for positional encoding, enabling flexible adaptation to images of varying sizes. To improve computational efficiency, average pooling is applied over adjacent $2 \times 2$ feature patches. For instance, an image with a resolution of $336 \times 336$, when encoded by a ViT with a patch size of 14, will first be converted into 576 visual tokens. Then, after passing through a $4 \times $ patch merger, it will be compressed into 144 visual tokens before being fed into the large language model (LLM).

\subsection{Multimodal sequence packing}
Traditional ViT (such as CLIP \cite{radford2021learning} and SigLip \cite{zhai2023sigmoidlosslanguageimage}) usually process images of arbitrary resolution and aspect ratio into fixed-length patch sequences to facilitate GPU parallel computing. However, the native resolution method generates variable-length patch sequences. To achieve parallel computing, a strategy similar to text processing is often adopted: unifying the sequence length through padding. However, in order to be compatible with the longest sequence in the batch, shorter sequences need to be padded, resulting in computational redundancy.

To solve this problem, we used NaViT's Patch n' Pack \cite{dehghani2023patchnpacknavit} to pack multiple patches of different images into one sequence. Multiple images can be processed simultaneously in a sub-sequence calculation. During the training process, for each image $\bm{I_{i}}$ of different resolutions in a batch $\mathcal{B}=\{\bm{I}_1,\bm{I}_2,...,\bm{I}_K\}$, we preprocess it to a patch sequence $\bm{E}_i\in\mathbb{R}^{N_{i} \times P^{2}C}$, and directly concatenate all patches to obtain a packed sequence $\bm{P}_{s}$.
As shown in Figure \ref{fig:pipeline}, the Variable Length Flash Attention strategy \cite{dao2023flashattention2} is adopted in ViT to separate and optimize the Attention computation by leveraging the cumulative sequence lengths, which delineate the boundaries of individual patch sequences within the packed sequence $\bm{P}_{s}$, thereby ensuring that the Attention calculation of each patch sequence  $\bm{E}_i$  is isolated from each other.



\section{Experiment}
\label{Experiment}
\subsection{Implementation details}

\textbf{Model setting.} In this study, we use Qwen2-7B-Instruct \cite{yang2024qwen2} as the LLM backbone and a native resolution ViT initialized from Qwen2-VL-2B \cite{wang2024qwen2} as the Vision Encoder. All experiments were conducted on 8 NVIDIA A100-80G GPUs, and to ensure fair comparative analysis, we maintained training configurations identical to those of LLaVA-NeXT \cite{2023llava1.6} and LLaVA \cite{liu2023visual}. To avoid out-of-memory errors, we cap the maximum number of tokens per image for different settings and datasets. The detailed experimental settings are elaborated below.

\textbf{Stage1: VLM pre-training.} During the initial pre-training stage, we freeze both the LLM and Vision Encoder components while exclusively fine-tuning the Visual Projector parameters. We leverage the LLaVA-Pretrain dataset \cite{liu2023visual} comprising 558K image-text pairs and train for 1 epoch with a global batch size of 128. We utilize the AdamW optimizer \cite{loshchilov2017decoupled} with a cosine learning rate scheduler  \cite{loshchilov2016sgdr} at a maximum learning rate of 1e-3.

\textbf{Stage2: Visual Instruction Tuning.} For the visual instruction tuning stage, we employ two configurations with datasets from LLaVA-1.5 and LLaVA-NeXT. The first uses LLaVA-mix665k dataset \cite{liu2023visual} with frozen Vision Encoder while fine-tuning the MLP Projector and LLM. The second uses LLaVA-NeXT-Data (779K samples with added OCR datasets) \cite{2023llava1.6} and fine-tunes all components, applying a 2e-6 learning rate for the Vision Encoder. Both configurations use a global batch size of 32 and train for 1 epoch, with AdamW optimizer, cosine learning rate scheduler, and a maximum learning rate of 1e-5.

\subsection{Experimental setup}

\textbf{General Resolution-Centric benchmarks.}
The RC-type benchmarks include TextVQA (VQA$^\mathrm{T}$) \cite{singh2019towards}, OCRBench (OCR) \cite{liu2024ocrbench}, DocVQA (VQA$^\mathrm{D}$) \cite{mathew2021docvqa}, ChartQA (VQA$^\mathrm{C}$) \cite{masry2022chartqa},  InfographicVQA (VQA$^\mathrm{I}$) \cite{mathew2022infographicvqa}, HR-Bench (HR) \cite{wang2025divide} and MM-Vet (MMV) \cite{yu2023mm}. These benchmarks are sensitive to image resolution, as resizing or cropping operations cause detail loss and geometric distortion that impact performance. We use these to evaluate model capabilities in RC-type tasks like text recognition, geometric feature interpretation, and fine-grained detail extraction.

\textbf{General Semantic-Centric benchmarks.}
The SC-type benchmarks include AI2D \cite{kembhavi2016diagram}, MathVista (Math) \cite{lu2023mathvista}, SEED-Bench-Image (SEED) \cite{li2023seed}, MME-Cognition (MME$^\mathrm{C}$) \cite{fu2023mme}, MME-Perception (MME$^\mathrm{P}$) \cite{fu2023mme}, MMBench-CN (MMB$^\mathrm{C}$) \cite{liu2024mmbench}, MMBench-EN (MMB$^\mathrm{E}$) \cite{liu2024mmbench} and POPE \cite{li2023evaluating}. These assess semantic understanding capabilities that remain relatively robust to resolution changes. Through these benchmarks, we evaluate model performance in  SC-type tasks like scene comprehension, object detection and classification.

\textbf{Evaluation protocol.}
We comprehensively evaluate the model from the following aspects: (1) The size of the model's training data; (2) The maximum resolution supported by the model; (3) The types of  dynamic-resolution visual encoding strategy the model supports; (4) The model's capability in RC-type tasks; (5) The model's capability in SC-type tasks.(5) The model's capability in RC-Bench.

\subsection{Main results on general benchmarks}
\label{subsec:Main-results}

We conducted tests on 13 general benchmarks. Models in Table~\ref{tab:performance_v1} were pre-trained on LLaVA-Pretrain dataset and fine-tuned on LLaVA-mix665k dataset (1.22M samples total), which lacks RC-type examples. Therefore, most methods are only tested on a small number of relevant benchmarks. Table~\ref{tab:performance_v2} presents models trained on enriched datasets, substantially improving RC-type task performance. Our approach demonstrates three key advantages: \textbf{(1) Superior performance without specialized data.} As shown in Table~\ref{tab:performance_v1}, NativeRes-LLaVA achieved best results on TextVQA (72.4\%) and MM-Vet (46.6\%) despite limited training data, confirming its inherent advantage in handling dynamic resolutions. \textbf{(2) Dramatic improvements with minimal specialized data.} As shown in Table~\ref{tab:performance_v2}, with just 1.34M data including RC-type samples, NativeRes-LLaVA significantly outperformed models trained on larger datasets, surpassing Cambrian-1 on OCRBench (705 vs. 624), multiple LLaVA-NeXT variants on DocVQA (89.7\% vs. 78.2\%), and data-intensive TextHawk on InfographicVQA (61.0\% vs. 50.6\%), while achieving best results across TextVQA (74.0\%), ChartQA (79.0\%), HR-Bench (61.3\%), and MM-Vet (46.6\%/1.22M and 44.5\%/1.34M). \textbf{(3) Maintained competitive performance on SC-type tasks.} Combining Table~\ref{tab:performance_v1} and Table~\ref{tab:performance_v2}, on SC-type benchmarks that are not sensitive to image resolution, our model still achieved competitive results and outperformed most models.

\begin{table}[!h]
\centering
\vspace{-10pt}
\caption{Performance comparison of models pre-trained on LLaVA-Pretrain dataset and fine-tuned on LLaVA-mix665k dataset. MaxRes indicates the maximum supported input resolution, while RS denotes the Resolution Strategy: Fixed Resolution (Fixed), Upsampling Methods (Up), Cropping-based (Crop), Hybrid vision encoders (Hybrid) and Native Resolution Encoding (Native).}
\vspace{3pt}
\resizebox{0.9\textwidth}{!}{%
\setlength{\tabcolsep}{2pt}  
\begin{tabular}{@{}cccc|cc|cccccc@{}} 
\toprule
\multirow{3}{*}{Method} & \multirow{3}{*}{LLM} & \multirow{3}{*}{MaxRes} & \multirow{3}{*}{RS} & \multicolumn{2}{c|}{\multirow{2}{*}{Resolution-Centric}} & \multicolumn{6}{c}{\multirow{2}{*}{Semantic-Centric}} \\ 
 &  &  &  & \multicolumn{2}{c|}{} & \multicolumn{6}{c}{} \\ 
 &  &  &  & VQA$^\mathrm{T}$ & MMV & Math & SEED & MME$^\mathrm{P}$ & MMB$^\mathrm{C}$ & MMB$^\mathrm{E}$ & POPE  \\ \midrule 
LLaVA-1.5~\cite{liu2024improved} & Vicuna-7B & 336×336 & Fixed & 58.2 & 31.1 & 25.2 & 66.1 & 1511 & 58.3 & 64.3 & 87.3 \\ 
LLaVA-1.5~\cite{liu2024improved} & Vicuna-13B & 336×336 & Fixed & 61.3 & 36.1 & 27.6 & 68.2 & 1531 & 63.6 & 67.7 & 87.1 \\ 
S\textasciicircum{}2~\cite{shi2024we} & Vicuna-7B & 1008×1008 & Up & 61.0 & 32.4 & 25.3 & 67.9 & - & - & 66.2 & - \\ 
S\textasciicircum{}2~\cite{shi2024we} & Vicuna-13B & 1008×1008 & Up & 63.1 & 36.4 & 27.8 & 68.9 & - & - & 67.9 & - \\ 
LLaVA-1.5-Qwen~\cite{liu2024improved} & Qwen2-7B-Instruct & 336×336 & Fixed & - & - & 33.6 & 69.4 & - & - & 72.0 & - \\ 
LLaVA-HR~\cite{luo2024feast} & Vicuna-7B & 1024×1024 & Hybrid & 67.1 & 31.2 & - & 64.2 & 1555 & - & - & 87.6 \\ 
LLaVA-HR~\cite{luo2024feast} & Vicuna-13B & 1024×1024 & Hybrid & 68.1 & 34.8 & - & 64.5 & 1541 & - & - & 87.8 \\ 
LLaVA-UHD~\cite{guo2024llava} & Vicuna-13B & 672×1008 & Crop & 67.7 & - & - & - & 1535 & 64.8 & 68.0 & \textbf{89.1} \\ 
PIIP-LLaVA~\cite{wang2025parameter} & Vicuna-7B & 1024×1024 & Hybrid & 67.1 & 31.4 & - & 69.4 & - & - & 67.0 & 88.2 \\ 
PIIP-LLaVA~\cite{wang2025parameter} & Vicuna-13B & 1024×1024 & Hybrid & 69.2 & 36.8 & - & 70.5 & - & - & 66.5 & 87.6 \\ 
LLaVA-NeXT~\cite{2023llava1.6} & Qwen2-7B-Instruct & 768×768 & Crop & 63.3 & 39.7 & 31.5 & 72.6 & 1549 & 74.4 & \textbf{73.3} & 88.8 \\ 
\midrule \midrule
NativeRes-LLaVA & Qwen2-7B-Instruct & 1260×1260 & Native & \textbf{72.4} & \textbf{46.6} & \textbf{36.6} & \textbf{73.1} & \textbf{1566} & \textbf{75.7} & 73.2 & 88.6 \\ 
\bottomrule
\end{tabular}%
}
\vspace{-10pt}
\label{tab:performance_v1}
\end{table}

\begin{table}[h]
\vspace{-10pt}
\caption{Performance of models training on datasets of varying scales. \#Data indicates the combined volume of train data. In the  gray part, LLaVA-OneVision and Qwen2-VL-7B can respectively represent the reference results of LLaVA-NeXT and NativeRes-LLaVA after training with more data.}
\vspace{3pt}
\Large
\resizebox{\columnwidth}{!}{%
\setlength{\tabcolsep}{2pt}  
\begin{tabular}{@{}ccccc|ccccccc|cccccccc@{}}
\toprule
\multirow{3}{*}{Method} & \multirow{3}{*}{LLM} & \multirow{3}{*}{Data} & \multirow{3}{*}{MaxRes} & \multirow{3}{*}{RS} 
& \multicolumn{7}{c|}{\multirow{2}{*}{Resolution-Centric}} 
& \multicolumn{8}{c}{\multirow{2}{*}{Semantic-Centric}} \\
 &  &  &  &  & \multicolumn{7}{c|}{} & \multicolumn{8}{c}{} \\
 &  &  &  &  & VQA$^\mathrm{T}$ & OCR & VQA$^\mathrm{D}$ & VQA$^\mathrm{C}$ & VQA$^\mathrm{I}$ & HR & MMV 
 & AI2D & Math & SEED & MME$^\mathrm{C}$ & MME$^\mathrm{P}$ & MMB$^\mathrm{C}$ & MMB$^\mathrm{E}$ & POPE \\ \midrule
UReader~\cite{ye2023ureader} & LLaMA2-7B & 86M & 896×1120 & Crop 
& 57.6 & - & 65.4 & 59.3 & 42.2 & - & - 
& - & - & - & - & - & - & - & - \\
Monkey~\cite{li2024monkey} & Vicuna-7B & 1.40B & 896×1344 & Crop 
& 67.6 & 514 & 66.5 & 65.1 & 36.1 & 38.0 & - 
& 62.6 & 33.5 & 64.3 & - & - & - & 59.8 & 67.6 \\
Mini-Gemini-HD~\cite{li2024mini} & Vicuna-7B & 2.7M & 1536×1536 & Hybrid 
& 68.4 & 477 & 65.0 & 47.4 & - & 50.1 & 41.3 
& 68.2 & 32.2 & 66.9 & 319 & 1546 & - & 65.8 & 86.8 \\
Mini-Gemini-HD~\cite{li2024mini} & Vicuna-13B & 2.7M & 1536×1536 & Hybrid 
& 70.2 & 501 & 70.0 & - & - & - & \textbf{50.5} 
& - & 37.0 & - & 320 & {\ul 1597} & - & 68.6 & 87.0 \\
TextHawk~\cite{yu2024texthawk} & InternLM-7B & 115M & 1344×1344 & Crop 
& - & - & 76.4 & 66.6 & {\ul 50.6} & - & - 
& - & - & 69.2 & - & 1500 & - & 74.6 & - \\
LLaVA-NeXT~\cite{2023llava1.6} & Vicuna-7B & 1.34M & 672×672 & Crop 
& 64.9 & 532 & 63.6 & 54.8 & - & 47.9 & - 
& 66.6 & 34.4 & 70.2 & 323 & 1519 & - & 67.4 & - \\
LLaVA-NeXT~\cite{2023llava1.6} & Vicuna-13B & 1.34M & 672×672 & Crop 
& 67.1 & 537 & 77.5 & 62.2 & - & - & - 
& 70.0 & 35.1 & 71.9 & 317 & 1575 & - & 70.0 & - \\
LLaVA-NeXT~\cite{2023llava1.6} & LLaMA3-8B & 1.34M & 672×672 & Crop 
& - & - & {\ul 78.2} & 69.5 & - & - & - 
& 71.6 & 37.5 & - & 368 & \textbf{1604} & - & 72.1 & - \\
TokenPacker~\cite{li2024tokenpacker} & Vicuna-13B & 2.7M & 1344×1344 & Crop 
& 70.6 & 521 & 70.0 & - & - & - & - 
& - & - & - & 350 & 1574 & - & 68.7 & 88.1 \\
SliME~\cite{zhang2024beyond} & LLaMA3-8B & 2.0M & 2016×2016 & Crop 
& 65.6 & - & - & - & - & - & 36.8 
& - & 43.6 & - & 346 & 1573 & 71.0 & 75.4 & 86.0 \\
Cambrian-1~\cite{tong2024cambrian} & LLaMA3-8B & 9.5M & 1024×1024 & Hybrid 
& 71.7 & {\ul 624} & 77.8 & {\ul 73.3} & - & - & - 
& 73.0 & \textbf{49.0} & \textbf{74.7} & - & 1547 & - & 75.9 & - \\
MG-LLaVA~\cite{zhao2024mg} & LLaMA3-8B & 2.56M & 768×768 & Hybrid 
& 68.1 & - & 49.0 & - & - & - & - 
& 75.6 & - & 71.5 & - & - & - & 76.6 & - \\
LLaVA-UHD-v2~\cite{zhang2024llava} & Qwen2-7B-Instruct & 1.42M & 1008×672 & Crop 
& 70.6 & 577 & 72.9 & 70.4 & - & 59.9 & - 
& 75.5 & 39.1 & 73.6 & - & - & - & 77.1 & - \\
LLaVA-NeXT~\cite{2023llava1.6} & Qwen2-7B-Instruct & 1.22M & 768×768 & Crop 
& 63.3 & 416 & 42.6 & 25.3 & 28.3 & 58.1 & 39.7 
& 64.5 & 31.5 & 72.6 & 329 & 1549 & 74.4 & 73.3 & \textbf{88.8} \\
LLaVA-NeXT~\cite{2023llava1.6} & Qwen2-7B-Instruct & 1.34M & 768×768 & Crop 
& 67.3 & 566 & 73.4 & 71.4 & 37.8 & 54.5 & 43.0 
& {\ul 77.3} & 40.5 & {\ul 74.6} & 357 & 1562 & {\ul 76.3} & \textbf{78.2} & 88.4 \\ \midrule \midrule
NativeRes-LLaVA & Qwen2-7B-Instruct & 1.22M & 1792×1792 & Native 
& {\ul 72.4} & 584 & 67.7 & 44.4 & 40.2 & {\ul 60.4} & {\ul 46.6} 
& 66.3 & 36.6 & 73.1 & {\ul 448} & 1566 & 75.7 & 73.2 & {\ul 88.6} \\
NativeRes-LLaVA & Qwen2-7B-Instruct & 1.34M & 1792×1792 & Native 
& \textbf{74.0} & \textbf{705} & \textbf{89.7} & \textbf{79.0} & \textbf{61.0} & \textbf{61.3} & 44.5 
& \textbf{78.2} & {\ul 43.9} & 74.1 & \textbf{483} & 1568 & \textbf{76.9} & {\ul 77.6} & 87.7 \\ \midrule
{\color[HTML]{C0C0C0} LLaVA-OneVision~\cite{li2024llava}} & {\color[HTML]{C0C0C0}Qwen2-7B-Instruct} & {\color[HTML]{C0C0C0}9.35M} & {\color[HTML]{C0C0C0}2304×2304} & {\color[HTML]{C0C0C0}Crop} 
& {\color[HTML]{C0C0C0}-} & {\color[HTML]{C0C0C0}622} & {\color[HTML]{C0C0C0}87.5} & {\color[HTML]{C0C0C0}80.0} & {\color[HTML]{C0C0C0}68.8} & {\color[HTML]{C0C0C0}-} & {\color[HTML]{C0C0C0}57.5} 
& {\color[HTML]{C0C0C0}81.4} & {\color[HTML]{C0C0C0}63.2} & {\color[HTML]{C0C0C0}75.4} & {\color[HTML]{C0C0C0}418} & {\color[HTML]{C0C0C0}1580} & {\color[HTML]{C0C0C0}-} & {\color[HTML]{C0C0C0}80.8} & {\color[HTML]{C0C0C0}-} \\
{\color[HTML]{C0C0C0}Qwen2-VL-7B~\cite{wang2024qwen2}} & {\color[HTML]{C0C0C0}Qwen2-7B-Instruct} & {\color[HTML]{C0C0C0}1.4T$^\mathrm{Tok}$} & {\color[HTML]{C0C0C0}-} & {\color[HTML]{C0C0C0}Native} 
& {\color[HTML]{C0C0C0}84.3} & {\color[HTML]{C0C0C0}866} & {\color[HTML]{C0C0C0}94.5} & {\color[HTML]{C0C0C0}88.4} & {\color[HTML]{C0C0C0}76.5} & {\color[HTML]{C0C0C0}-} & {\color[HTML]{C0C0C0}62.0} 
& {\color[HTML]{C0C0C0}83.0} & {\color[HTML]{C0C0C0}58.2} & {\color[HTML]{C0C0C0}-} & {\color[HTML]{C0C0C0}-} & {\color[HTML]{C0C0C0}-} & {\color[HTML]{C0C0C0}80.5} & {\color[HTML]{C0C0C0}83.0} & {\color[HTML]{C0C0C0}-} \\ \bottomrule
\end{tabular}%
}
\vspace{-10pt}
\label{tab:performance_v2}
\end{table}

\subsection{Ablation study}
Based on the findings in Section \ref{subsec:Main-results}, we identify three promising dynamic visual encoding strategies (Cropping-based, Hybrid vision encoders, and Native Resolution Encoding) and perform thorough ablation studies on general benchmarks and RC-Bench to assess their comparative strengths.

\subsubsection{Results on general benchmarks}

\textbf{Enhancing Model Performance via Native Resolution Encoding.}
In Tables~\ref{tab:performance_v1} and~\ref{tab:performance_v2}, NativeRes-LLaVA demonstrates significantly greater effectiveness than our baseline, LLaVA-NeXT, on the RC-type tasks. LLaVA-NeXT employs SigLip-384 \cite{zhai2023sigmoidlosslanguageimage} as its ViT. To verify that the performance gains of NativeRes-LLaVA are primarily attributable to the Native Resolution resolution strategy rather than the use of a more powerful ViT (Qwen2-VL-ViT), we replaced the ViT module in LLaVA-NeXT with the ViT from Qwen2-VL \cite{wang2024qwen2}, and fixed its input resolution to match that of SigLip-384. As shown in Table~\ref{tab:ablation}, this substitution leads to substantial improvements overall. However, performance remains clearly inferior to that of NativeRes-LLaVA—particularly on RC-type tasks such as DocVQA, InfoVQA, HR-Bench, and RC-Bench highlighting the superiority of the Native Resolution strategy over the cropping-based resolution strategy.

\textbf{High MaxRes significantly improves the model performance.}
To assess the impact of maximum resolution (MaxRes) on the model’s visual understanding capabilities, we restricted the input resolution of NativeRes-LLaVA by resizing images such that each image yields around 729 tokens (equivalent to 378×378 MaxRes) before Patch Merging. As shown in Table~\ref{tab:ablation}, this limitation leads to a notable performance drop on RC-type tasks compared with other native strategies of high MaxRes. Furthermore, when the Native Resolution Strategy is adopted, a higher MaxRes achieves better results, underscoring the critical role of high resolution in such evaluations. Interestingly, performance on Semantic-Centric tasks remains largely unaffected and, in some cases, even improves. This observation suggests that these benchmarks primarily emphasize overall image semantics while overlooking fine-grained visual details. These results validate the resolution dilemmas discussed in Section \ref{subsec:Bench Dilemmas} and support the effectiveness of our proposed classification of multimodal benchmarks into Semantic-Centric (SC) and Resolution-Centric (RC) tasks.

\begin{table}[h]
\vspace{-10pt}
\caption{Ablation study to verify the contribution of NativeRes strategy and importance of MaxRes.}
\vspace{3pt}
\Large
\resizebox{\columnwidth}{!}{%
\setlength{\tabcolsep}{3pt}  
\begin{tabular}{@{}ccccc|cccccccc|cccccccc@{}} 
\toprule
\multirow{3}{*}{Method} & \multirow{3}{*}{\#Data} & \multirow{3}{*}{ViT} & \multirow{3}{*}{MaxRes} & \multirow{3}{*}{RS} & \multicolumn{8}{c|}{\multirow{2}{*}{{Resolution-Centric}}} & \multicolumn{8}{c}{\multirow{2}{*}{{Semantic-Centric}}} \\ 
 &  &  &  &  & \multicolumn{8}{c|}{} & \multicolumn{8}{c}{} \\ 
 &  &  &  &  & VQA$^\mathrm{T}$ & OCR & VQA$^\mathrm{D}$ & VQA$^\mathrm{C}$ & VQA$^\mathrm{I}$ & HR & MMV &RC & AI2D & Math & SEED & MME$^\mathrm{C}$ & MME$^\mathrm{P}$ & MMB$^\mathrm{C}$ & MMB$^\mathrm{E}$ & POPE\\ \midrule 
LLaVA-NeXT & \multirow{5}{*}{1.22M} & SigLip-384 & 768×768 & Crop & 63.3 & 416 & 42.6 & 25.3 & 28.3 & {\ul 58.1} & 39.7 & 22.3 & 64.5 & 31.5 & {\ul 72.6} & 329 & 1549 & 73.3 & 74.4 & \textbf{88.8} \\
LLaVA-NeXT &  & Qwen2-VL-ViT & 728×728 & Crop & 69.0 & 546 & 50.9 & 36.9 & 31.6 & 47.9 & 44.7 & 42.4 & \textbf{66.6} & {\ul 36.4} & 71.6 & 437 & 1486 & \textbf{74.3} & \textbf{76.6} & 88.0 \\
NativeRes-LLaVA &  & Qwen2-VL-ViT & 378×378 & Fixed & 60.2 & 494 & 33.0 & 30.9 & 27.3 & 48.3 & 40.9 & 40.1 & 66.2 & 35.8 & 72.5 & \textbf{480} & \textbf{1577} & {\ul 73.9} & {\ul 76.1} & 87.0 \\
NativeRes-LLaVA &  & Qwen2-VL-ViT & 728×728 & Native & {\ul 71.3} & {\ul 567} & {\ul 58.2} & {\ul 44.2} & {\ul 39.5} & 56.4 & \textbf{47.0} & {\ul 49.6} & {\ul 66.4} & 36.2 & 72.3 & 444 & {\ul 1572} & 73.2 & 75.7 & {\ul 88.6} \\
NativeRes-LLaVA &  & Qwen2-VL-ViT & 1260×1260 & Native & \textbf{72.4} & \textbf{584} & \textbf{67.7} & \textbf{44.4} & \textbf{40.2} & \textbf{60.4} & {\ul 46.6} & \textbf{51.9} & 66.3 & \textbf{36.6} & \textbf{73.1} & {\ul 448} & 1566 & 73.2 & 75.7 & {\ul 88.6} \\ \midrule \midrule
LLaVA-NeXT & \multirow{5}{*}{1.34M} & SigLip-384 & 768×768 & Crop & 67.3 & 566 & 73.4 & 71.4 & 37.8 & 54.5 & 43.0 & 27.7 & 77.3 & 40.5 & \textbf{74.6} & 357 & 1562 & 76.3 & \textbf{78.2} & \textbf{88.4} \\
LLaVA-NeXT &  & Qwen2-VL-ViT & 728×728 & Crop & 70.2 & 666 & 77.5 & 76.2 & 36.7 & {\ul 55.1} & 41.4 & 47.9 & 77.9 & \textbf{43.9} & 72.9 & 446 & \textbf{1584} & {\ul 76.9} & 77.5 & 87.5 \\
NativeRes-LLaVA &  & Qwen2-VL-ViT & 378×378 & Fixed & 59.8 & 608 & 57.7 & 71.2 & 30.0 & 50.1 & 41.8 & 45.4 & 77.8 & 42.5 & 73.2 & 406 & 1539 & 75.2 & 76.8 & 87.0 \\
NativeRes-LLaVA &  & Qwen2-VL-ViT & 728×728 & Native & {\ul 72.7} & {\ul 685} & {\ul 80.2} & {\ul 78.7} & {\ul 41.5} & 51.4 & {\ul 44.0} & {\ul 53.6} & {\ul 78.1} & 43.3 & 73.9 & \textbf{483} & 1555 & \textbf{77.0} & {\ul 77.7} & {\ul 87.7} \\
NativeRes-LLaVA &  & Qwen2-VL-ViT & 1792×1792 & Native & \textbf{74.0} & \textbf{705} & \textbf{89.7} & \textbf{79.0} & \textbf{61.0} & \textbf{61.3} & \textbf{44.5} & \textbf{60.1} & \textbf{78.2} & \textbf{43.9} & {\ul 74.1} & \textbf{483} & {\ul 1568} & {\ul 76.9} & 77.6 & {\ul 87.7} \\ \bottomrule
\end{tabular}%
}
\label{tab:ablation}
\vspace{-10pt}
\end{table}

\subsubsection{Results on RC-Bench}

Given that RC-Bench exhibits a balanced distribution across the Area and Ratio dimensions, we evaluate the robustness of vision-language models (VLMs) in dynamic resolution encoding strategies not only by reporting the average accuracy (acc), but also by introducing the Coefficient of Variation (CV) along both dimensions as a complementary metric. As a dimensionless statistic that measures relative dispersion, CV is calculated by dividing the standard deviation $\sigma$ by the mean $\mu$, thereby capturing the degree of variability relative to the average level. By incorporating CV, we gain a more fine-grained understanding of model performance stability under varying resolution conditions, which facilitates the identification of more robust encoding strategies.

Table \ref{myvlm} provides evidence that, under constrained training data conditions, the native resolution strategy substantially outperforms cropping-based approaches on RC-Bench in terms of accuracy. Moreover, it exhibits greater robustness to both area-level variations—as measured by the coefficient of variation of area accuracy (ACV)—and changes in aspect ratio—as reflected in the coefficient of variation of ratio accuracy (RCV). Across all metrics, it consistently surpasses both cropping-based and fixed-resolution strategies.

As shown in Table \ref{open-source-vlm}, to strengthen the persuasiveness of our conclusions, we selected several representative open-source models employing the Cropping, Hybrid, and Native strategies, while keeping their parameter sizes approximately consistent. The experimental results show that VLMs using the Native strategy consistently outperform those using the other two approaches, both in terms of accuracy and robustness to variations in Area and Ratio. These findings further substantiate our main conclusion.
Comprehensive visualization results are presented in Appendix \ref{sec:supp-rcbench} to supplement and support the main findings.
\begin{table}[h]
\vspace{-8pt}
\centering
\begin{minipage}{0.45\linewidth}
\vspace{-10pt}
\caption{Ablation study of Fixed, Crop, and Native resolution strategies on RC-Bench. ACV and RCV values are multiplied by 10\textsuperscript{2}.}
\vspace{3pt}
\Large
\resizebox{\linewidth}{!}{%
\setlength{\tabcolsep}{3pt}
\begin{tabular}{@{}cccc|ccc@{}}
\toprule
Method & \#Data & MaxRes & RS & \multicolumn{1}{l}{Acc.} & \multicolumn{1}{l}{ACV$\downarrow$
} & \multicolumn{1}{l}{RCV$\downarrow$
} \\ \midrule
LLaVA-NeXT-SigLip & \multirow{5}{*}{1.22M} & 768×768 & Crop & 22.3 & 40.9 & 34.5 \\
LLaVA-NeXT-QwenViT &  & 728×728 & Crop & 42.4 & 27.5 & 27.6 \\
NativeRes-LLaVA &  & 378×378 & Fixed & 40.1 & 32.6 & 32.4 \\
NativeRes-LLaVA &  & 728×728 & Native & 49.6 & 21.8 & 15.4 \\
NativeRes-LLaVA &  & 1260×1260 & Native & \textbf{51.9} & \textbf{18.0} & \textbf{12.5} \\ \midrule
LLaVA-NeXT-SigLip & \multirow{5}{*}{1.34M} & 768×768 & Crop & 27.7 & 28.6 & 36.6 \\
LLaVA-NeXT-QwenViT &  & 728×728 & Crop & 47.9 & 23.5 & 25.0 \\
NativeRes-LLaVA &  & 378×378 & Fixed & 45.4 & 25.5 & 25.2 \\
NativeRes-LLaVA &  & 728×728 & Native & 53.6 & 19.1 & 16.8 \\
NativeRes-LLaVA &  & 1792×1792 & Native & \textbf{60.1} & \textbf{13.4} & \textbf{7.2} \\ \bottomrule
\label{myvlm}
\end{tabular}
}
\end{minipage}%
\hfill
\begin{minipage}{0.48\linewidth}
\vspace{-10pt}
\caption{Comparison of open source models using different resolution strategies on RC-Bench. ACV and RCV values are multiplied by 10\textsuperscript{2}.}
\vspace{3pt}
\Large
\resizebox{\linewidth}{!}{%
\setlength{\tabcolsep}{3pt}
\begin{tabular}{@{}cccc|ccc@{}}
\toprule
Method & LLM & \#Data & RS & Acc. & ACV$\downarrow$ & RCV$\downarrow$ \\ \midrule
Cambrian-1~\cite{tong2024cambrian} & LLaMA3-8B-Instruct & 9.5M & Hybrid & 44.1 & 19.3 & 30.1 \\ \cmidrule(){1-4}
LLaVA-OneVision~\cite{li2024llava}  & Qwen2-7B-Instruct & 9.35M & \multirow{4}{*}{Crop} & 55.3 & 13.4 & 11.4 \\
Intern-VL-2.5~\cite{chen2024expanding}  & InternLM2.5-7B-chat & 142B$^\mathrm{Tok}$ &  & 65.0 & 9.7 & 5.7 \\
Intern-VL-3~\cite{zhu2025internvl3} & Qwen2.5-7B & 200B$^\mathrm{Tok}$ &  & 60.4 & 28.3 & 5.9 \\
{\color[HTML]{C0C0C0}DeepSeek-VL2~\cite{wu2024deepseek}} & {\color[HTML]{C0C0C0}28B-A4B} & {\color[HTML]{C0C0C0}800B$^\mathrm{Tok}$} &  & {\color[HTML]{C0C0C0}65.3} & {\color[HTML]{C0C0C0}16.4} & {\color[HTML]{C0C0C0}3.8} \\ \cmidrule(){1-4}
NativeRes-LLaVA & Qwen2-7B-Instruct & 1.34M & \multirow{5}{*}{Native} & 60.1 & 13.4 & 7.2 \\
Qwen2-VL~\cite{wang2024qwen2} & Qwen2-7B-Instruct & 1.4T$^\mathrm{Tok}$ &  & 77.6 & \textbf{5.8} & \textbf{2.0} \\
Qwen2.5-VL~\cite{bai2025qwen2} & Qwen2.5-7B-Instruct & 4.1T$^\mathrm{Tok}$ &  & \textbf{80.4} & 6.3 & 2.1 \\
{\color[HTML]{C0C0C0}Kimi-VL~\cite{team2025kimi}}  & {\color[HTML]{C0C0C0}16B-A3B} & {\color[HTML]{C0C0C0}4.4T$^\mathrm{Tok}$} &  & {\color[HTML]{C0C0C0}77.6} & {\color[HTML]{C0C0C0}7.9} & {\color[HTML]{C0C0C0}1.7} \\
{\color[HTML]{C0C0C0}Seed1.5-VL~\cite{guo2025seed15vltechnicalreport}}  & {\color[HTML]{C0C0C0}A20B} & {\color[HTML]{C0C0C0}3T+$^\mathrm{Tok}$} &  & {\color[HTML]{C0C0C0}75.1} & {\color[HTML]{C0C0C0}7.4} & {\color[HTML]{C0C0C0}3.3} \\ 
\bottomrule
\label{open-source-vlm}
\end{tabular}%
}
\end{minipage}
\vspace{-20pt}
\end{table}

\subsubsection{Cropping-based Vs. Native Resolution Visual Encoding}
\begin{figure}[!h]
\centering
\begin{subfigure}{0.32\textwidth}
  \centering
  \includegraphics[width=0.95\linewidth]{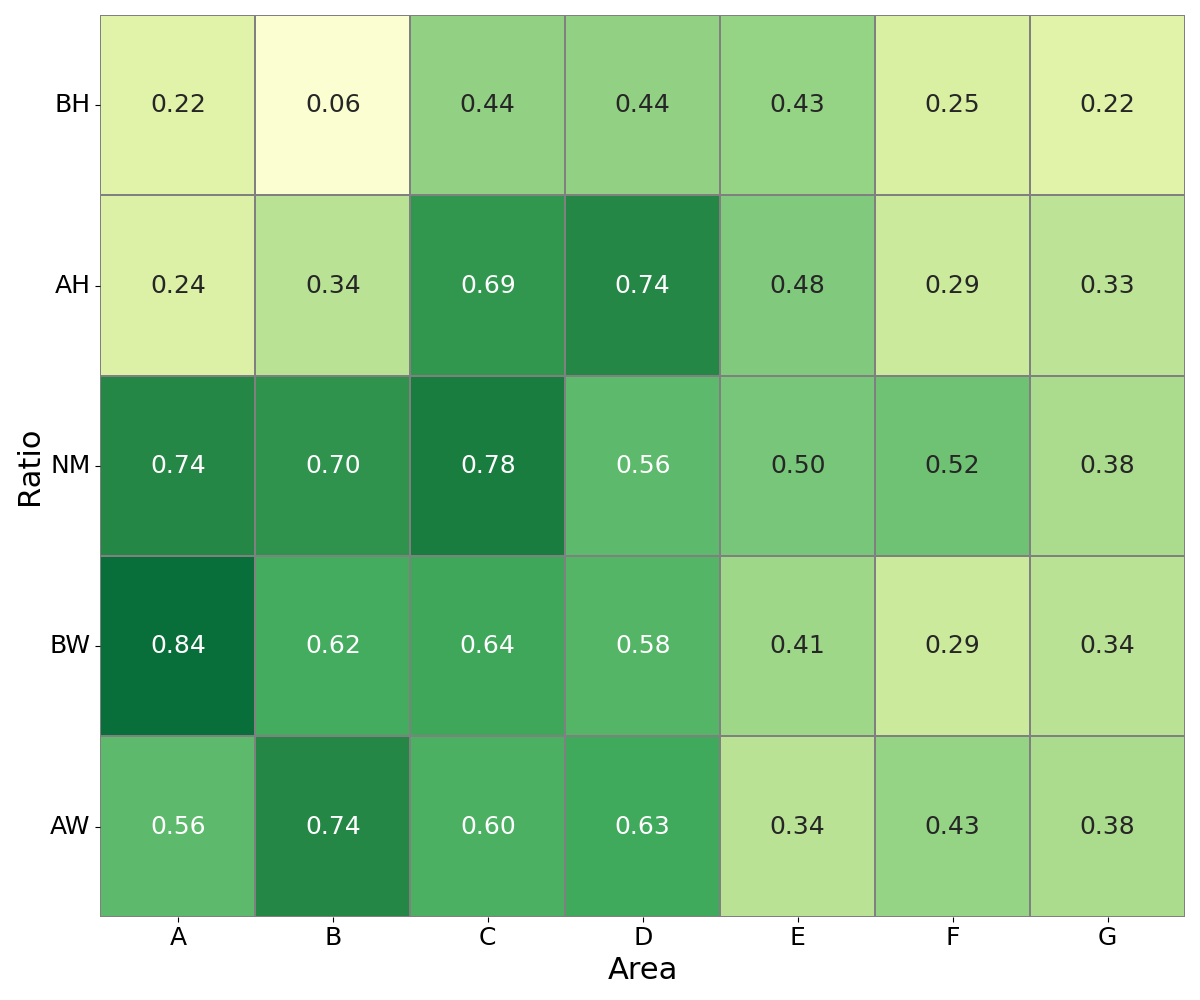}
  \caption{\footnotesize $728 \times 728$, Crop}
  \label{fig:crop-based}
\end{subfigure}
\hfill
\begin{subfigure}{0.32\textwidth}
  \centering
  \includegraphics[width=0.95\linewidth]{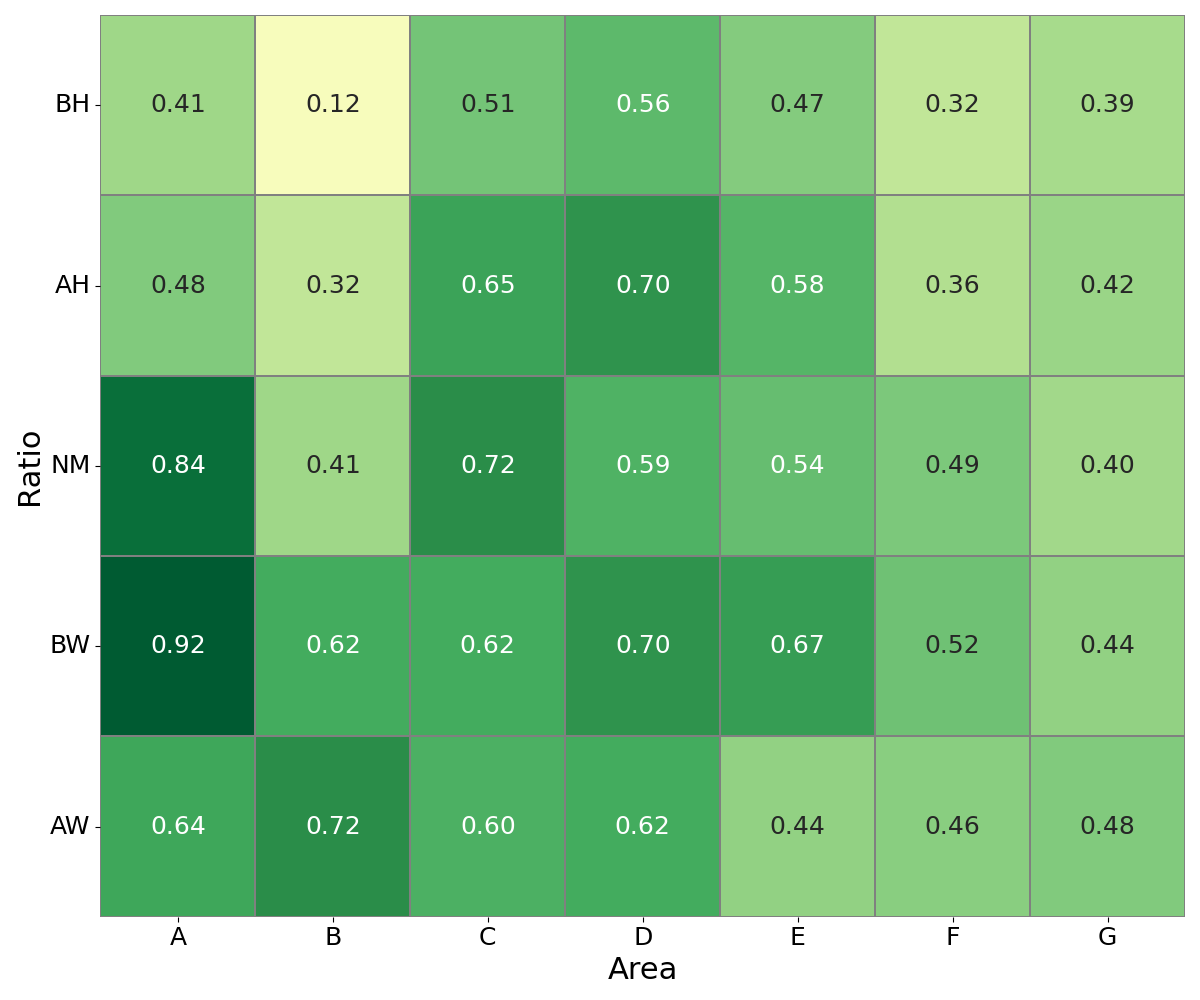}
  \caption{\footnotesize $728 \times 728$, Native}
  \label{fig:native-res}
\end{subfigure}
\hfill
\begin{subfigure}{0.32\textwidth}
  \centering
  \includegraphics[width=0.95\linewidth]{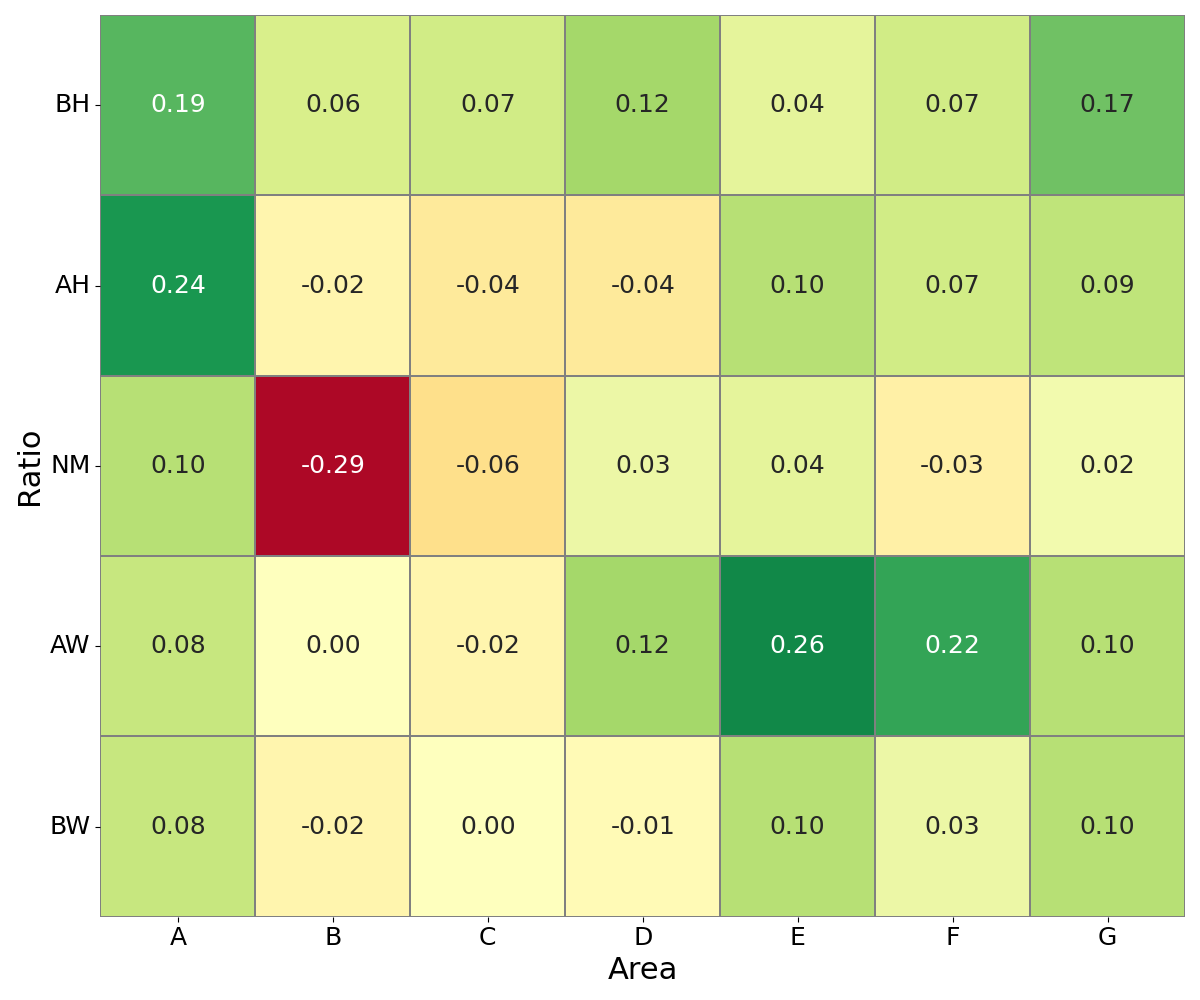}
  \caption{\footnotesize Accuracy Difference}
  \label{fig:delta}
\end{subfigure}

\vspace{-8pt}
\caption{\textbf{Comparison of Visual Encoding Strategy on the RC-Bench.} The figure illustrates the change in accuracy of the Native Resolution method (b) compared to the Cropping-based method (a) at a $728 \times 728$ resolution. (c), Accuracy Difference, shows the value in each cell calculated by subtracting the accuracy of the Cropping-based method from the accuracy of the Native Resolution method (i.e., (b) -- (a)).}
\label{fig:res-comparison}
\vspace{-10pt}
\end{figure}

\begin{table}[h]
\centering
\caption{Ablation study of Cropping-based and Native resolution strategies on RC-Bench. ACV and RCV values are multiplied by 10\textsuperscript{2}.}
{\small 
\begin{tabular}{@{}cccc|ccc@{}}
\toprule
Method & \#Data & MaxRes & RS & Acc. & ACV$\downarrow$ & RCV$\downarrow$ \\ 
\midrule
LLaVA-NeXT-QwenViT & \multirow{2}{*}{1.34M} & 728×728 & Crop & 47.9 & 23.5 & 25.0 \\
NativeRes-LLaVA & & 728×728 & Native & 53.6 & 19.1 & 16.8 \\ 
\bottomrule
\vspace{-20pt}
\end{tabular}
}

\label{tab:smallfont}
\end{table}
In visual encoding for dynamic resolution tasks, the Cropping-based strategy is the current mainstream approach. To thoroughly investigate the performance differences between this method and the Native Resolution method, we designed a rigorous set of comparative experiments. All evaluations were conducted on the RC-Bench. To eliminate confounding variables, both encoding strategies were implemented on a unified Qwen2-VL-ViT backbone and configured with the same maximum resolution limit ( $728 \times 728$ ) .

We conducted a heatmap analysis on the RC-Bench results to compare the performance of Cropping-based versus Native Resolution visual encoding strategies in Table \ref{tab:smallfont}. As illustrated in the figure~\ref{fig:res-comparison}, the heatmap is a $7 \times 5$ grid with Area on the X-axis and Ratio on the Y-axis. More visualization results are presented in Appendix \ref{sec:supp-rcbench}.

In Figure~\ref{fig:res-comparison} (c), \textbf{(1) Green cells} indicate that the Native method has a higher accuracy (positive values).\textbf{(2) Red cells} indicate that the Crop method has a higher accuracy (negative values).\textbf{(3) Yellow cells} indicate that the accuracy of both methods is similar.

\paragraph*{Key Findings}
An analysis of the accuracy difference heatmap in Figure~\ref{fig:res-comparison} (c) (Native Resolution Accuracy - Cropping-based Accuracy) reveals the following key conclusions:

\begin{itemize}
    \item \textbf{Performance Drop in a Specific Case:} The Native Resolution method shows a significant performance degradation (accuracy difference of -0.29) in only one specific configuration: (NM, B). This indicates that the Cropping-based method performs substantially better in this particular task.

    \item \textbf{On-Par Performance in Common Scenarios:} Across most other common resolution and aspect ratio combinations, the performance of the Native Resolution method is largely on par with the Cropping-based method, with accuracy differences close to zero, as represented by the numerous light-colored cells in Figure~\ref{fig:res-comparison}(c).

    \item \textbf{Superior Robustness in Extreme Scenarios:} Critically, the Native Resolution method demonstrates a clear advantage when processing images with extreme aspect ratios (e.g., the BH row) or extreme areas (e.g., the G column). The dark green cells in Figure~\ref{fig:res-comparison}(c) (e.g., +0.24 at (AH, A) and +0.26 at (AW, E)) provide strong evidence for this. This strongly suggests that the Native Resolution method possesses superior robustness when faced with diverse and unconventional visual data.
\end{itemize}

\begin{wrapfigure}{r}{0.7\textwidth}
\vspace{-20pt}
  \centering
  \includegraphics[width=0.7\textwidth]{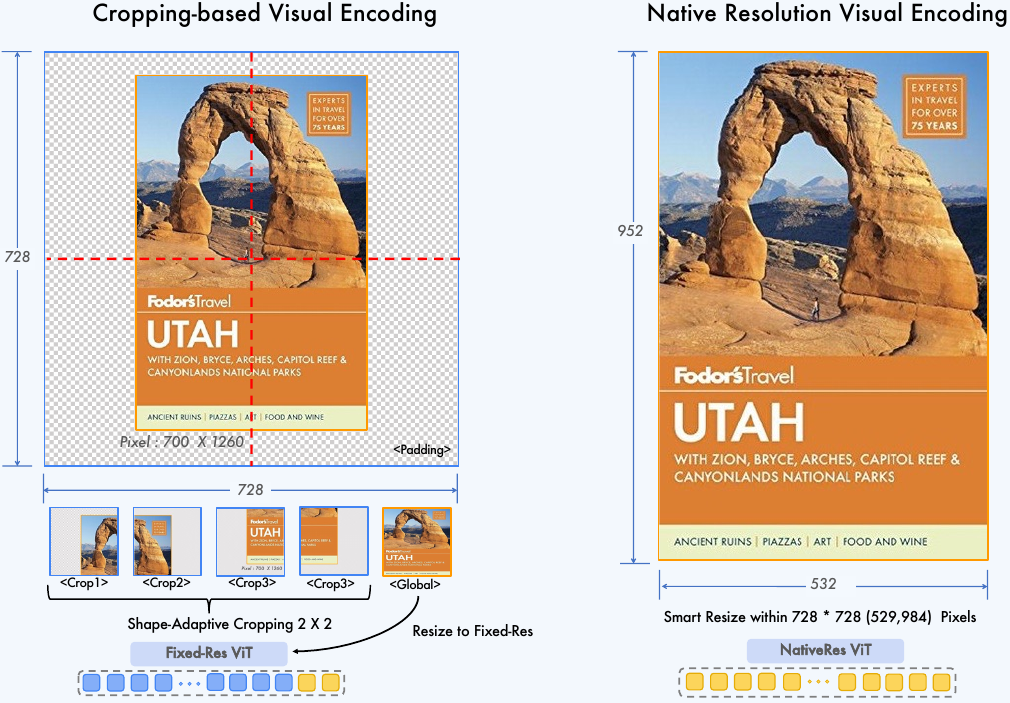}
  \caption{An illustration of the mechanism between Cropping-based and Native Resolution Visual Encoding.}
  \label{fig:RCvsSC}
\vspace{-20pt}
\end{wrapfigure}

\paragraph*{Mechanism Analysis}
We observe an anomalous performance discrepancy at the (NM, B) configuration, where the Cropping-based method surpasses the Native Resolution method. This phenomenon is explained by the inherent mechanics of each approach and the specific properties of (NM, B) data, which consists of small-sized images ($Area \in [100^2, 384^2]$) with a balanced aspect ratio ($w/h \in [1:2, 2:1]$). 

The design of the Cropping-based method typically involves supplying the vision encoder with a down-sampled thumbnail for global context, as  illustrated in Figure~\ref{fig:RCvsSC}. However, in the specific case of the (NM, B) task, the source image dimensions align almost perfectly with the vision encoder's standard fixed input resolution (e.g., $384 \times 384$). Critically, this is the same resolution format used during the model's extensive pre-training phase.

\begin{itemize}

    \item \textbf{Alignment with Pre-training} As a result, the Cropping-based method circumvents any need for significant down-sampling or resizing for the (NM, B) input. The data is presented in a format nearly identical to the pre-training corpus, creating an ideal condition for the model to achieve maximum efficacy.

    \item \textbf{Generalization vs. Specialization} Conversely, the Native Resolution method, despite its architectural flexibility for varied inputs, does not match the performance of the highly specialized Cropping-based approach in this \textbf{"sweet spot"} scenario. This particular performance dip underscores a key trade-off: while the Cropping-based method is optimized for a specific data format that mirrors pre-training, the Native Resolution method demonstrates superior generalization and robustness across a more diverse and challenging spectrum of visual inputs, which is its primary design advantage.
\end{itemize}

\subsubsection{Model performance at different LLM scales}
We conducted ablation experiments across different model scales, specifically evaluating the performance of NativeRes-LLaVA at 1B, 2B, and 7B scales (after integrating the Visual Encoder) on all benchmarks. LLaVA-NeXT served as the baseline throughout. All models were trained on 1.34M samples, comprising LLaVA-Pretrain and LLaVA-NeXT-Data. As shown in Table~\ref{tab:ablation_on_scale}, we observed a significant performance improvement in NativeRes-LLaVA as model scale increased. Notably, the 2B version of NativeRes-LLaVA already outperformed the 7B LLaVA-NeXT on extremely resolution-sensitive benchmarks such as TextVQA, OCRBench, DocVQA, InfographicVQA, and RC-Bench. These results further validate the superiority of the Native Resolution strategy over the Cropping-based strategy.

\begin{table}[h]
\vspace{-10pt}
\caption{Ablation experiments on different LLM scales.}
\vspace{3pt}
\Large
\resizebox{\columnwidth}{!}{%
\setlength{\tabcolsep}{3pt}  
\begin{tabular}{@{}cccc|cccccccc|cccccccc@{}}
\toprule
\multirow{3}{*}{Method} & \multirow{3}{*}{LLM} & \multirow{3}{*}{MaxRes} & \multirow{3}{*}{RS} & \multicolumn{8}{c|}{\multirow{2}{*}{Resolution-Centric}} & \multicolumn{8}{c}{\multirow{2}{*}{Semantic-Centric}} \\
 &  &  &  & \multicolumn{8}{c|}{} & \multicolumn{8}{c}{} \\
 &  &  &  & VQA$^\mathrm{T}$ & OCR & VQA$^\mathrm{D}$ & VQA$^\mathrm{C}$ & VQA$^\mathrm{I}$ & HR & MMV &RC & AI2D & Math & SEED & MME$^\mathrm{C}$ & MME$^\mathrm{P}$ & MMB$^\mathrm{C}$ & MMB$^\mathrm{E}$ & POPE\\ \midrule
LLaVA-NeXT & Qwen2-7B-Instruct        & 768×768 & Crop & 67.3 & 566 & 73.4 & {\ul 71.4} & 37.8 & {\ul 54.5} & {\ul 43.0} & 27.7 & {\ul 77.3} & {\ul 40.5} & \textbf{74.6} & 357 & {\ul 1562} & {\ul 76.3} & \textbf{78.2} & \textbf{88.4} \\
NativeRes-LLaVA & Qwen2-0.5B-Instruct & 1792×1792 & Native & 64.4 & 601 & 75.3 & 62.6 & 31.4 & 51.3 & 32.7 & 46.0 & 54.1 & 28.6 & 54.6 & 257 & 1309 & 51.2 & 52.1 &  {\ul 88.1} \\
NativeRes-LLaVA & Qwen2-1.5B-Instruct & 1792×1792 & Native & {\ul 71.0} & {\ul 630} & {\ul 85.4} & 70.7 & {\ul 42.7} & 54.3 & 37.2 & {\ul 56.3} & 66.9 & 35.4 & 61.0 & {\ul 381} & 1401 & 62.6 & 66.4 & 87.3 \\
NativeRes-LLaVA & Qwen2-7B-Instruct   & 1792×1792 & Native & \textbf{74.0} & \textbf{705} &\textbf{ 89.7} & \textbf{79.0} & \textbf{61.0} & \textbf{61.3} & \textbf{44.5} & \textbf{60.1} & \textbf{78.2} & \textbf{43.9} & {\ul 74.1} & \textbf{483} & \textbf{1568} & \textbf{76.9} & {\ul 77.6} & {\ul 87.7} \\ \bottomrule
\end{tabular}%
}
\label{tab:ablation_on_scale}
\end{table}

\subsubsection{Model performance at different LLMs}
Our Native Resolution strategy is compatible with any LLM while maintaining outstanding performance. To verify this, we replaced Qwen2-7B-Instruct \cite{yang2024qwen2} with Vicuna-7B \cite{chiang2023vicuna}, which has significantly lower language capability. As shown in Table~\ref{tab:ablation_on_llm}, thanks to our Native Resolution strategy, the performance of Vicuna-7B on extremely resolution-sensitive benchmarks remains significantly superior to that of LLaVA-NeXT using Qwen2-7B-Instruct. This further demonstrates the superiority of the Native Resolution strategy.

\begin{table}[h]
\vspace{-10pt}
\caption{Ablation experiments on different LLM.}
\vspace{3pt}
\Large
\resizebox{\columnwidth}{!}{%
\setlength{\tabcolsep}{3pt}
\begin{tabular}{@{}ccccc|cccccccc|cccccccc@{}}
\toprule
\multirow{3}{*}{Method} & \multirow{3}{*}{LLM} & \multirow{3}{*}{\#Data} & \multirow{3}{*}{MaxRes} & \multirow{3}{*}{RS}
& \multicolumn{8}{c|}{\multirow{2}{*}{Resolution-Centric}}
& \multicolumn{8}{c}{\multirow{2}{*}{Semantic-Centric}} \\
 &  &  &  &  & \multicolumn{8}{c|}{}
& \multicolumn{8}{c}{} \\
 &  &  &  &  & VQA$^\mathrm{T}$ & OCR & VQA$^\mathrm{D}$ & VQA$^\mathrm{C}$ & VQA$^\mathrm{I}$ & HR & MMV & RC
   & AI2D & Math & SEED & MME$^\mathrm{C}$ & MME$^\mathrm{P}$ & MMB$^\mathrm{C}$ & MMB$^\mathrm{E}$ & POPE \\
\midrule
LLaVA-NeXT & Qwen2-7B-Instruct & \multirow{3}{*}{1.22M} & 768×768 & Crop
    & 63.3 & 416 & 42.6 & 25.3 & 28.3 & {\ul 58.1} & 39.7 & 22.3
    & {\ul 64.5} & {\ul 31.5} & {\ul 72.6} & 329 & {\ul 1549} & \textbf{73.3} & {\ul 74.4} & \textbf{88.8} \\
NativeRes-LLaVA & Vicuna-7B &  & 1260×1260 & Native
    & {\ul 67.5} & {\ul 515} & {\ul 52.7} & {\ul 30.5} & {\ul 31.2} & 48.4 & {\ul 41.3} & {\ul 37.9}
    & 58.9 & 29.6 & 69.0 & {\ul 375} & 1511 & 59.8 & 68.2 & 87.8 \\
NativeRes-LLaVA & Qwen2-7B-Instruct &  & 1260×1260 & Native
    & \textbf{72.4} & \textbf{584} & \textbf{67.7} & \textbf{44.4} & \textbf{40.2} & \textbf{60.4} & \textbf{46.6} & \textbf{51.9}
    & \textbf{66.3} & \textbf{36.6} & \textbf{73.1} & \textbf{448} & \textbf{1566} & {\ul 73.2} & \textbf{75.7} & {\ul 88.6} \\
\midrule
LLaVA-NeXT & Qwen2-7B-Instruct & \multirow{3}{*}{1.34M} & 768×768 & Crop
    & 67.3 & 566 & 73.4 & 71.4 & 37.8 & {\ul 54.5} & 43.0 & 27.7
    & {\ul 77.3} & {\ul 40.5} & \textbf{74.6} & 357 & {\ul 1562} & {\ul 76.3} & \textbf{78.2} & \textbf{88.4} \\
NativeRes-LLaVA & Vicuna-7B &  & 1792×1792 & Native
    & {\ul 72.9} & {\ul 666} & {\ul 82.5} & {\ul 72.8} & {\ul 41.5} & 52.8 & {\ul 44.3} & {\ul 52.6}
    & 72.5 & 36.9 & 71.8 & 370 & 1479 & 63.5 & 72.0 & 87.3 \\
NativeRes-LLaVA & Qwen2-7B-Instruct &  & 1792×1792 & Native
    & \textbf{74.0} & \textbf{705} & \textbf{89.7} & \textbf{79.0} & \textbf{61.0} & \textbf{61.3} & \textbf{44.5} & \textbf{60.1}
    & \textbf{78.2} & \textbf{43.9} & {\ul 74.1} & \textbf{483} & \textbf{1568} & \textbf{76.9} & {\ul 77.6} & {\ul 87.7} \\
\bottomrule
\end{tabular}%
}
\label{tab:ablation_on_llm}
\end{table}



\section{Limitations and future work}

\textbf{1. Model and Training Strategy.}
A primary limitation stems from the constraints on our computational resources. This necessitated two main compromises in our methodology. 

First, our model does not yet achieve state-of-the-art (SOTA) performance. We plan to address this by conducting large-scale training on more extensive datasets, with the goal of reaching performance levels comparable to leading models such as Qwen-VL-Series~\cite{wang2024qwen2} and Seed1.5-VL~\cite{guo2025seed15vltechnicalreport}.

Second, rather than training a Vision Transformer (ViT) from scratch, we opted to initialize it with the pretrained weights from Qwen2-VL~\cite{wang2024qwen2}. This pragmatic choice aligns well with our framework, which is specifically designed to process high-resolution inputs natively.

Looking ahead, a key direction for future work is to investigate effective strategies for training a ViT from scratch at native resolution, with the goal of creating a model that is fully tailored to and seamlessly integrated with our proposed architecture.

\begin{figure}[!t]
\centering
\begin{subfigure}{0.32\textwidth}
  \centering
  \includegraphics[width=0.95\linewidth]{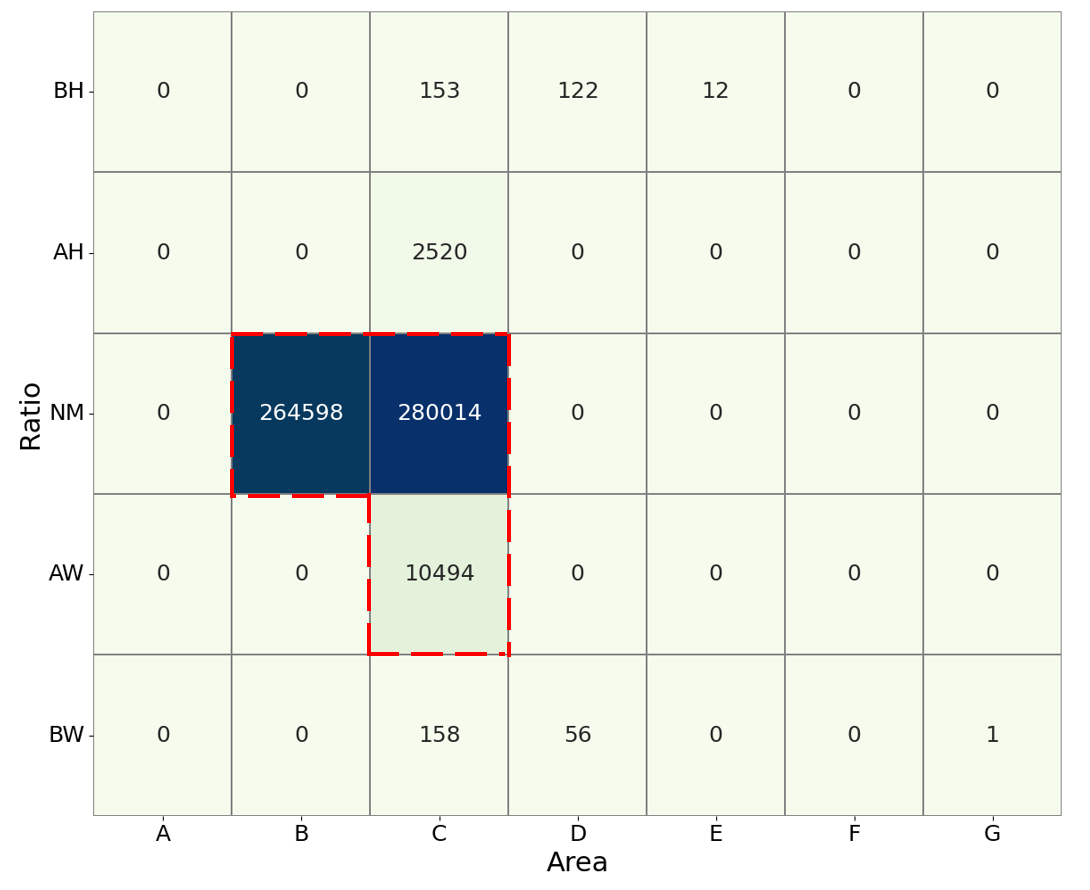}
  \caption{\footnotesize LLaVA-Pretrain-558K}
  \label{fig:LLaVA-Pretrain-558K}
\end{subfigure}
\hfill
\begin{subfigure}{0.32\textwidth}
  \centering
  \includegraphics[width=0.95\linewidth]{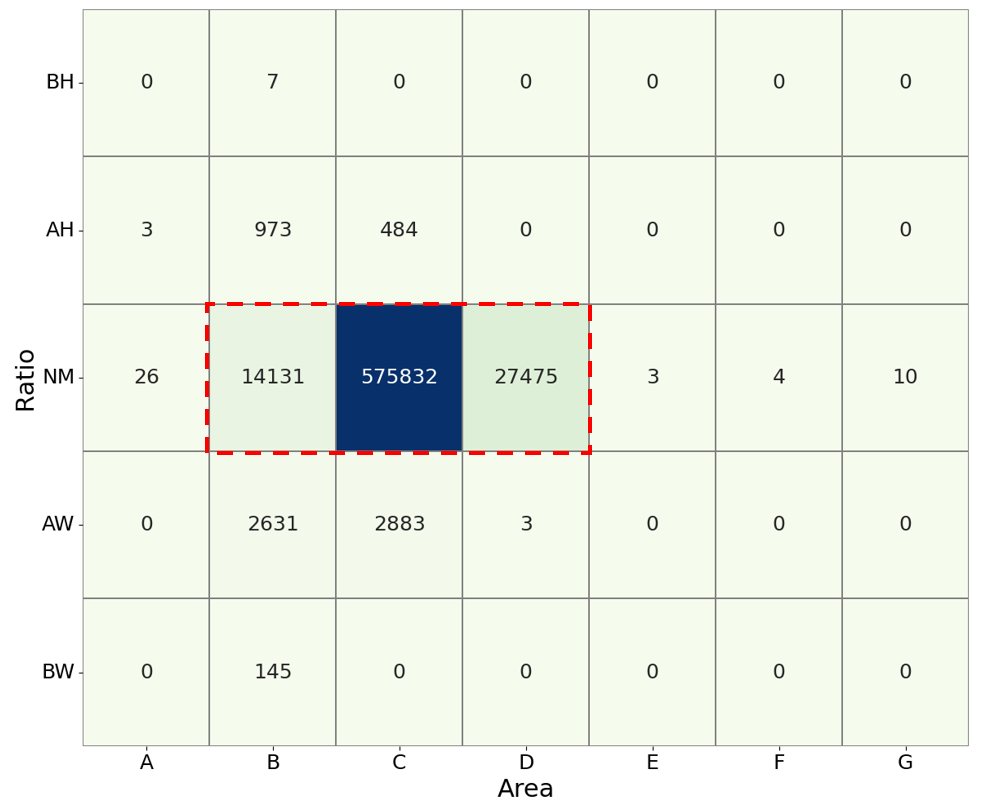}
  \caption{\footnotesize LLaVA-SFT-665K}
  \label{fig:LLaVA-SFT-665K}
\end{subfigure}
\hfill
\begin{subfigure}{0.32\textwidth}
  \centering
  \includegraphics[width=0.95\linewidth]{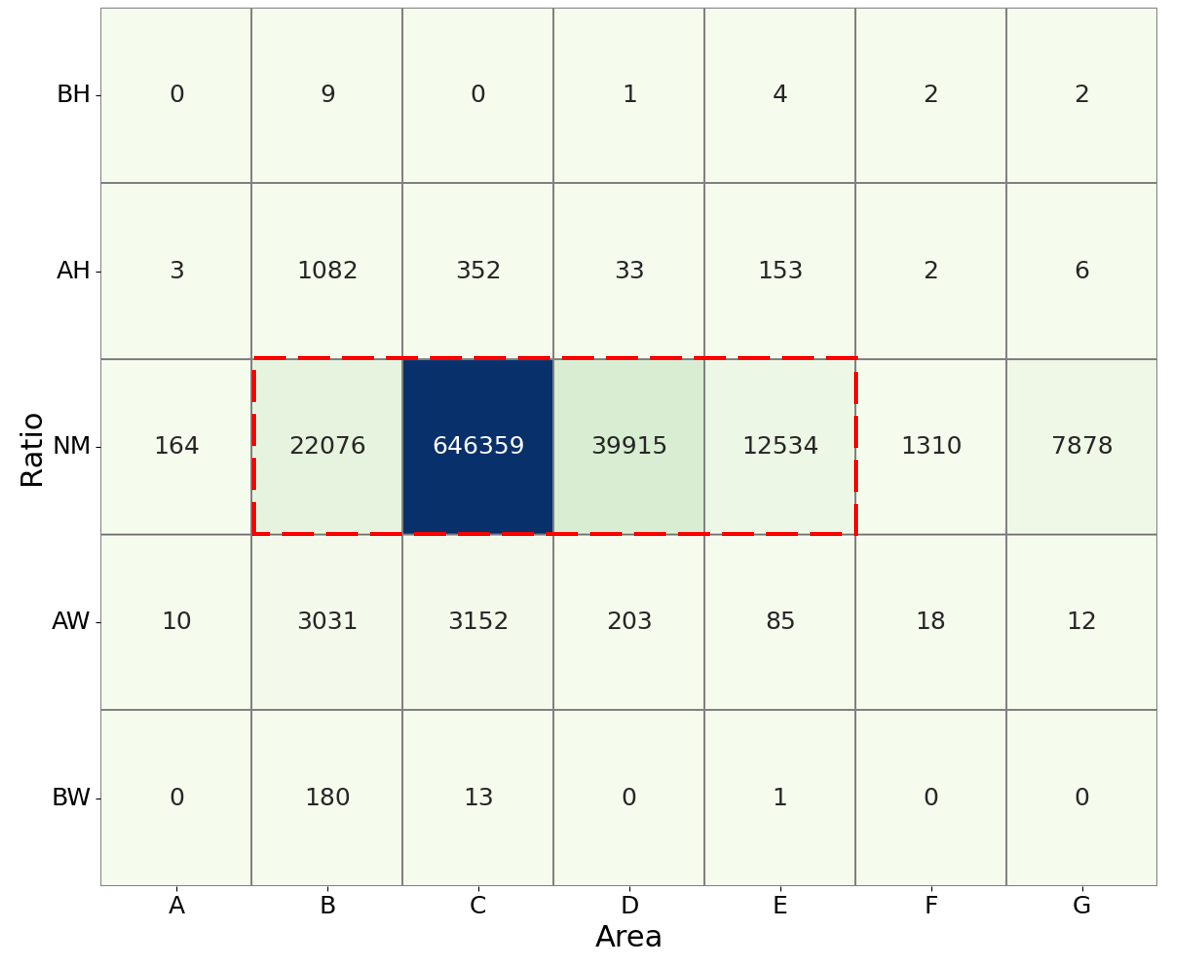}
  \caption{\footnotesize LLaVA-NeXT-SFT-779K}
  \label{fig:LLaVA-NeXT-SFT-779K}
\end{subfigure}

\vspace{-8pt}
\caption{Visualization of the image resolution and aspect ratio distribution used during the pretraining and fine-tuning stages}
\label{fig:dataset-distribution}
\vspace{-20pt}

\end{figure}
\textbf{2. Training Data and Resolution Generalization.}
The second set of limitations stems from the training data and the model's generalization capabilities. 
As illustrated in Figure~\ref{fig:dataset-distribution}, we visualize the resolution and aspect ratio distribution of the datasets used in both the pretraining and fine-tuning stages. 

We have observed that the predominant resolutions and aspect ratios of images in existing multimodal datasets are insufficient to fully leverage our model's high-resolution capabilities. Although our architecture supports inputs up to $1792 \times 1792$ pixels, this figure reflects the limitations of available training data rather than a hard architectural constraint. Configuring the model for even higher resolutions would be ineffective without datasets containing samples at a commensurate scale. 

This indicates that our model may not have been effectively trained on the full spectrum of resolutions it can handle, posing a significant challenge for future data curation and training strategies. Consequently, and also due to spatial constraints in this paper, we have not yet performed a detailed investigation into the problem of resolution generalization. 

Future research will focus on a meticulous analysis of how the distribution of image resolutions and aspect ratios in training data affects model performance on tasks requiring varied input dimensions. This investigation will be crucial for understanding and improving the model's generalization abilities across diverse visual contexts.

\textbf{3. Computational Bottleneck in High-Resolution Image Processing.}
Native resolution visual encoding faces an inherent challenge when processing high-resolution images. Specifically, for a Vision Transformer (ViT) operating at its native resolution, the computational complexity of the self-attention mechanism scales quadratically with the number of input patches, denoted as $O(N^2)$, where $N$ represents the number of patches. 

This quadratic scaling imposes a substantial computational and memory burden during the prefill stage. A key direction for our future research is to optimize this process, aiming to mitigate the computational costs associated with high-resolution inputs without compromising model performance.

It is noteworthy that recent models, such as Qwen2.5-VL~\cite{bai2025qwen2}, employ window attention to alleviate this issue. We plan to adapt our Qwen2.5-VL-ViT~\cite{bai2025qwen2} framework to further investigate and address this challenge.

\section{Conclusion}

In this paper,  we identified critical resolution dilemmas in the domain of VLMs. To address these dilemmas, we introduced RC-Bench, a resolution-centric benchmark designed to systematically evaluate visual encoding strategies under diverse visual conditions. Furthermore, we proposed NativeRes-LLaVA, a training framework for native-resolution visual encoding that preserves fine-grained image details. Extensive experiments across multiple benchmarks validate the effectiveness of native-resolution visual encoding and underscore the importance of resolution-aware design in VLMs. We hope our work will inspire further research into scalable and high-fidelity native visual encoding strategies for real-world applications.

\newpage

{
    \small
    \bibliographystyle{plain}
    \bibliography{ref}
}







\newpage
\appendix

\section{Technical Appendices and Supplementary Material}

\subsection{Data distribution of existing Benchmarks and RC-Bench}
\label{sec:rcbench}
\begin{figure}[h]
\centering

\begin{subfigure}{0.33\textwidth}
\centering
\includegraphics[width=\linewidth]{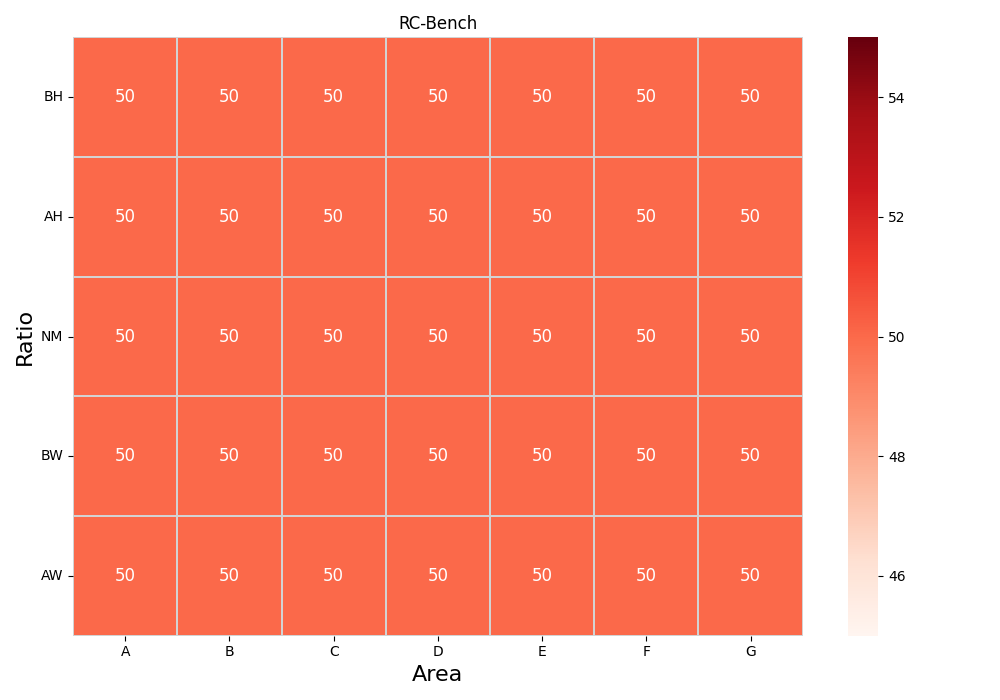}
\caption{RC-Bench}
\label{RC-Bench1}
\end{subfigure}%
\hfill
\begin{subfigure}{0.33\textwidth}
\centering
\includegraphics[width=\linewidth]{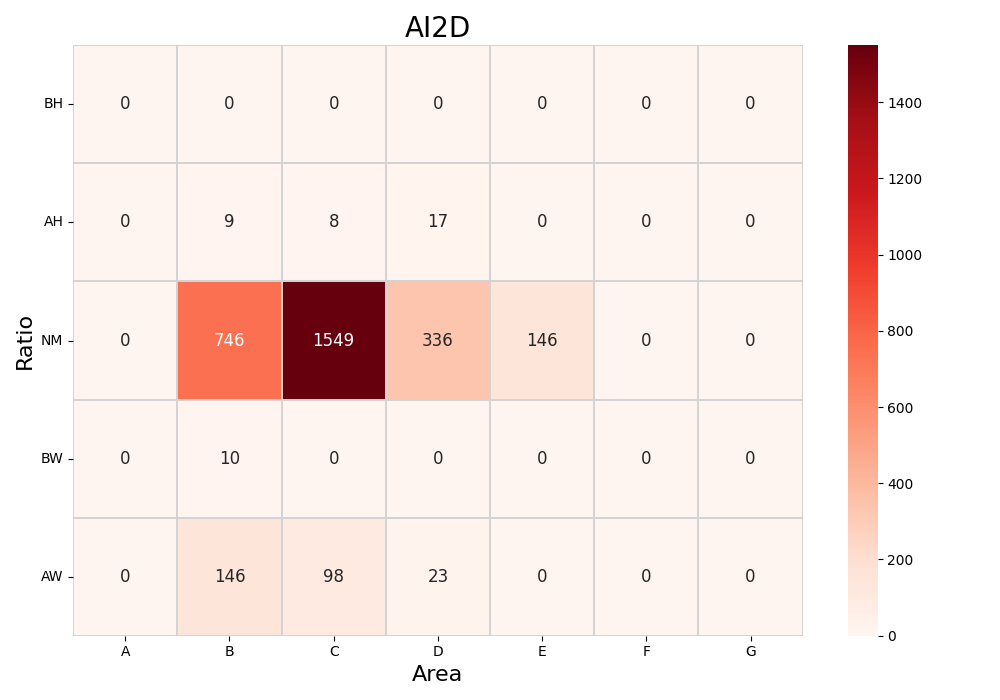}
\caption{AI2D}
\label{ai2d1}
\end{subfigure}%
\hfill
\begin{subfigure}{0.33\textwidth}
\centering
\includegraphics[width=\linewidth]{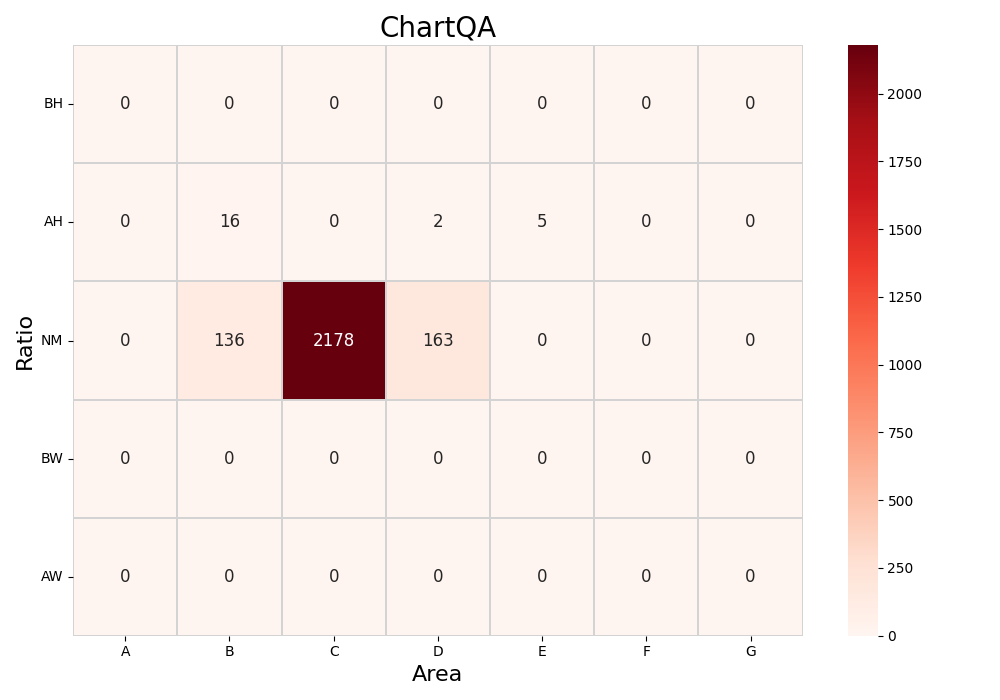}
\caption{ChartQA}
\label{ChartQA2}
\end{subfigure}

\vspace{10pt}

\begin{subfigure}{0.33\textwidth}
\centering
\includegraphics[width=\linewidth]{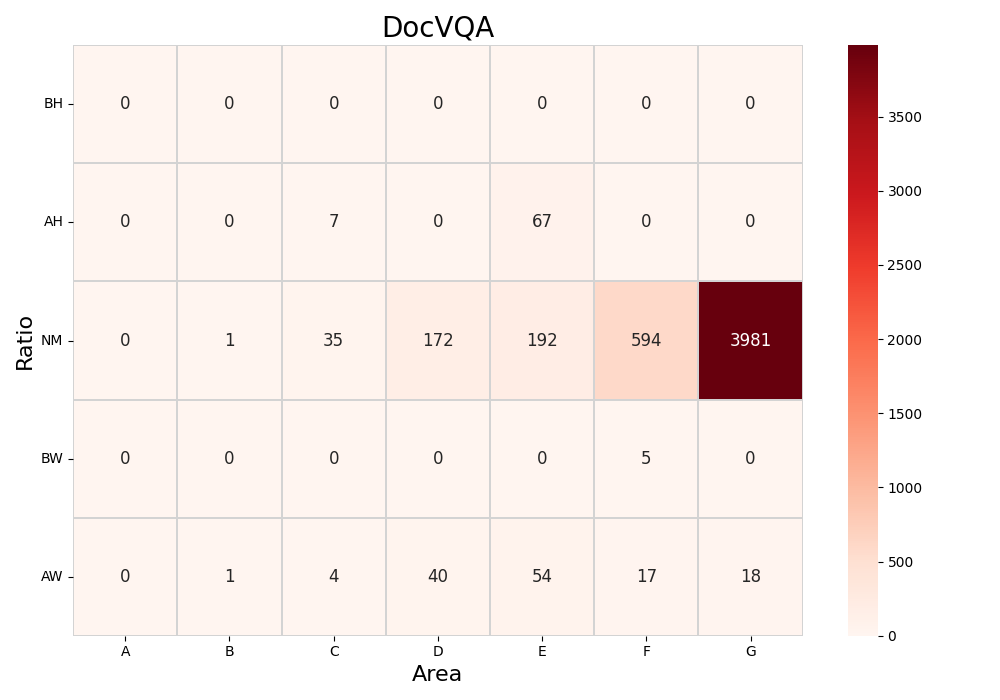}
\caption{DocVQA}
\label{DocVQA3}
\end{subfigure}%
\hfill
\begin{subfigure}{0.33\textwidth}
\centering
\includegraphics[width=\linewidth]{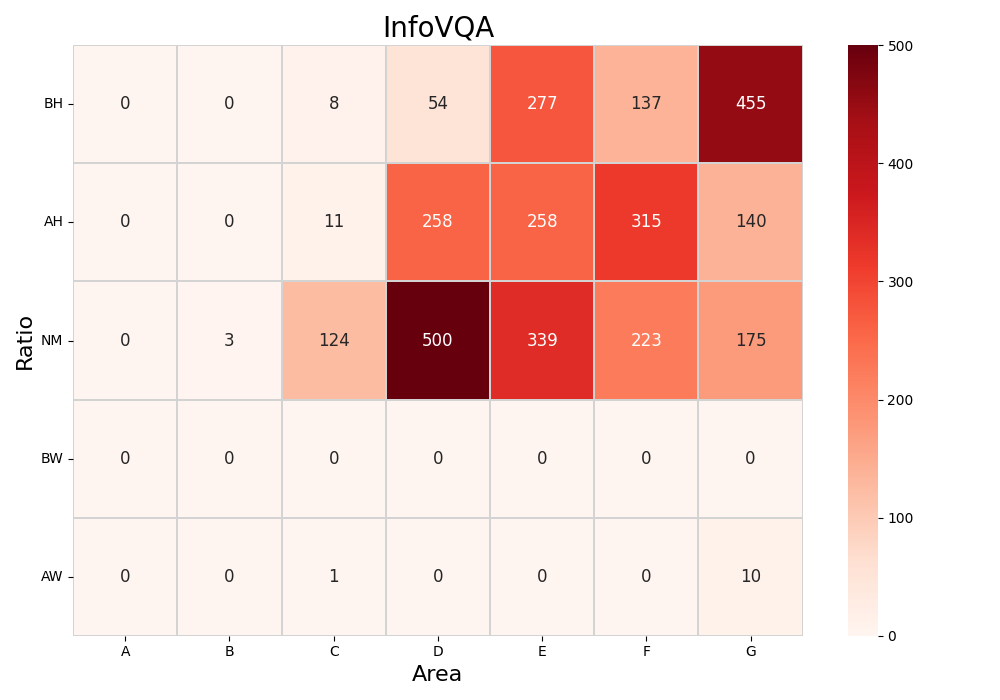}
\caption{InfoVQA}
\label{InfoVQA4}
\end{subfigure}%
\hfill
\begin{subfigure}{0.33\textwidth}
\centering
\includegraphics[width=\linewidth]{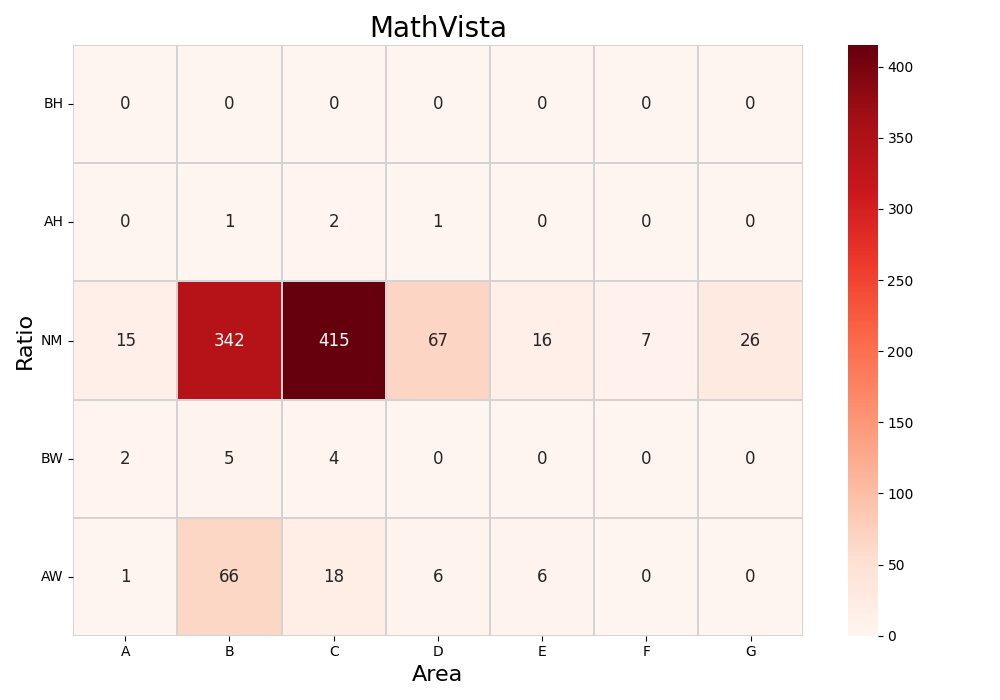}
\caption{MathVista}
\label{MathVista5}
\end{subfigure}

\vspace{10pt}

\begin{subfigure}{0.33\textwidth}
\centering
\includegraphics[width=\linewidth]{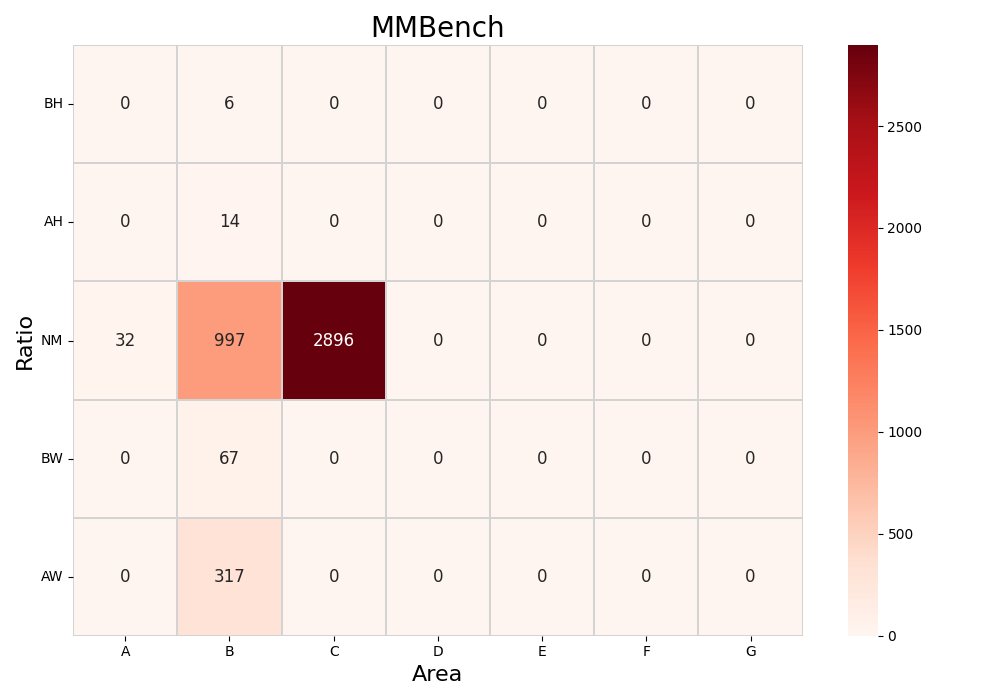}
\caption{MMBench}
\label{MMBench6}
\end{subfigure}%
\hfill
\begin{subfigure}{0.33\textwidth}
\centering
\includegraphics[width=\linewidth]{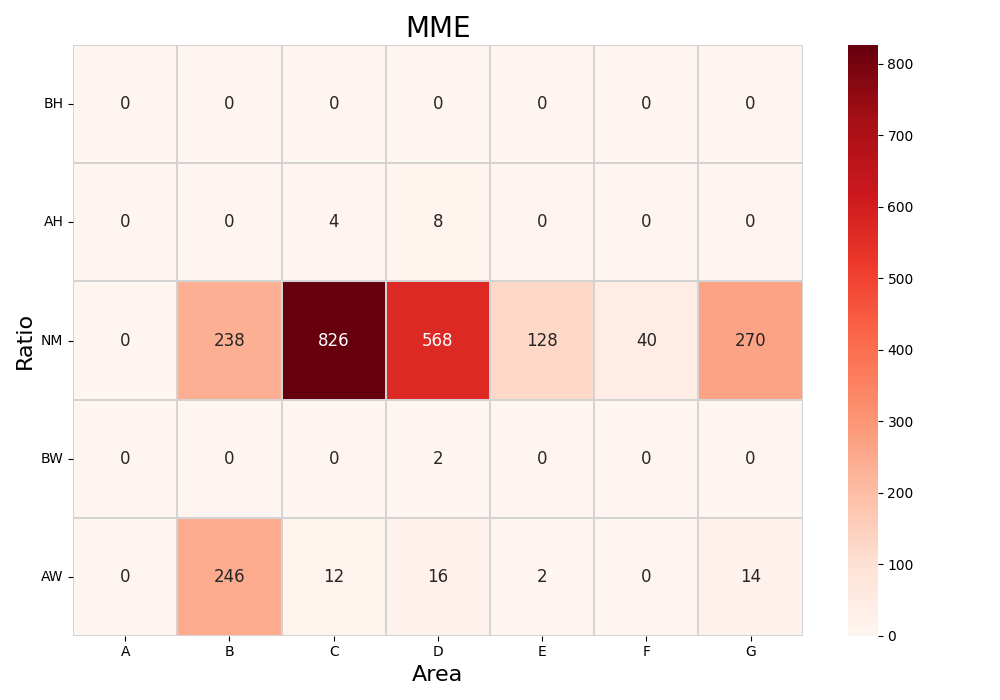}
\caption{MME}
\label{MME7}
\end{subfigure}%
\hfill
\begin{subfigure}{0.33\textwidth}
\centering
\includegraphics[width=\linewidth]{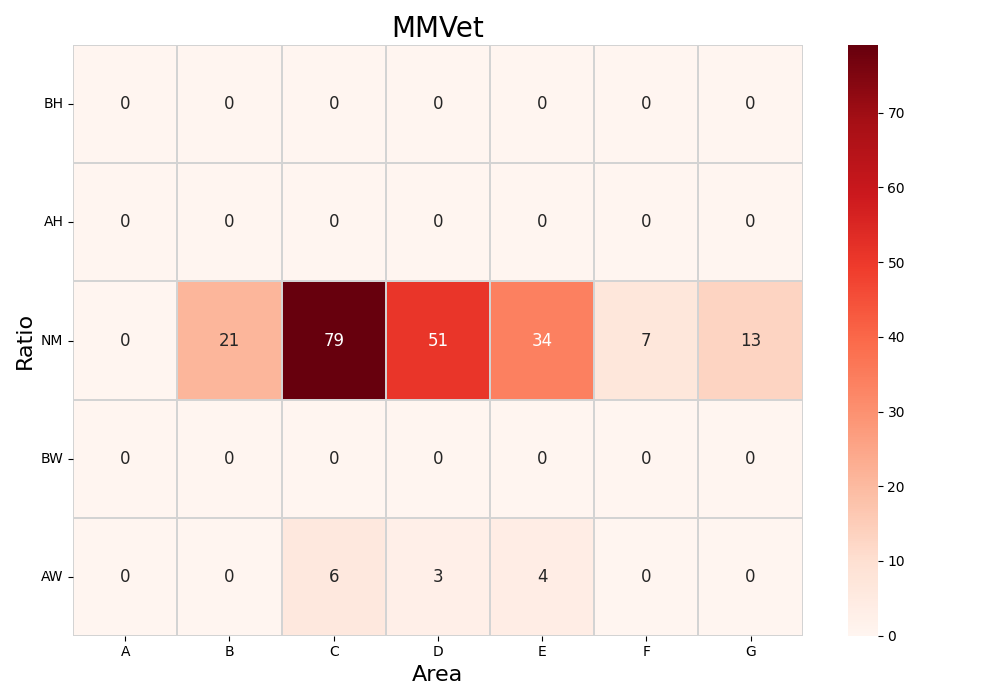}
\caption{MM-Vet}
\label{MMVet8}
\end{subfigure}

\vspace{10pt}

\begin{subfigure}{0.33\textwidth}
\centering
\includegraphics[width=\linewidth]{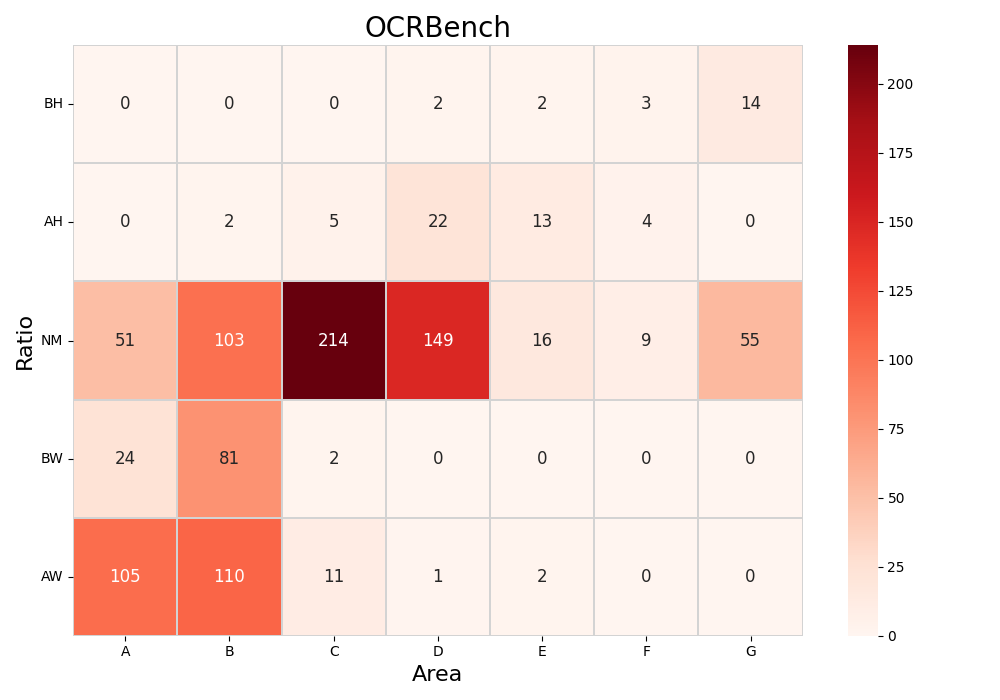}
\caption{OCRBench}
\label{OCRBench9}
\end{subfigure}%
\hfill
\begin{subfigure}{0.33\textwidth}
\centering
\includegraphics[width=\linewidth]{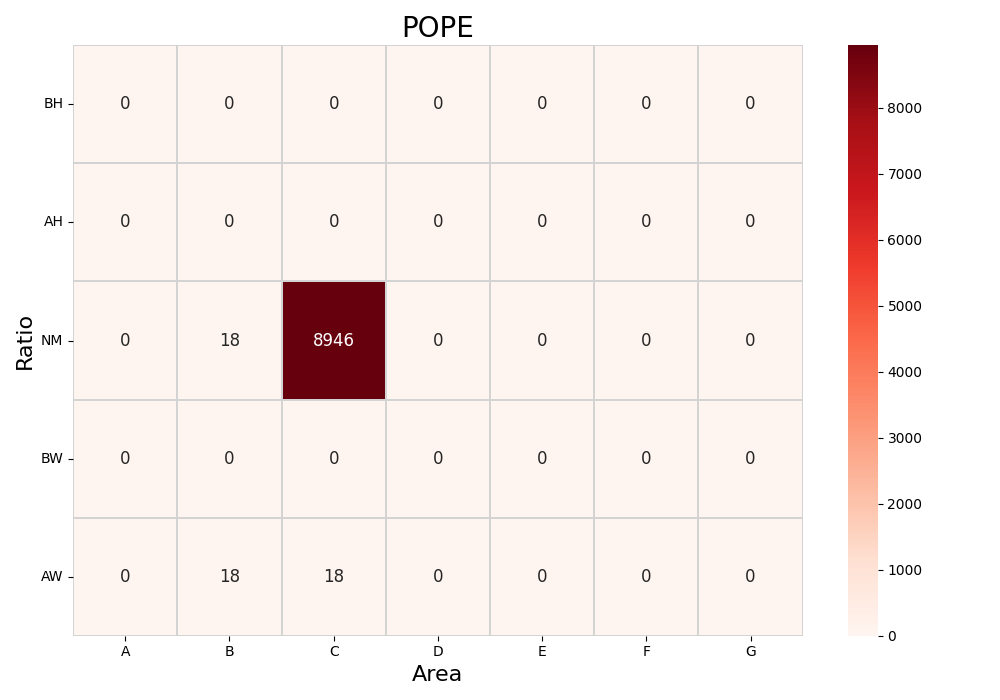}
\caption{POPE}
\label{Pope10}
\end{subfigure}%
\hfill
\begin{subfigure}{0.33\textwidth}
\centering
\includegraphics[width=\linewidth]{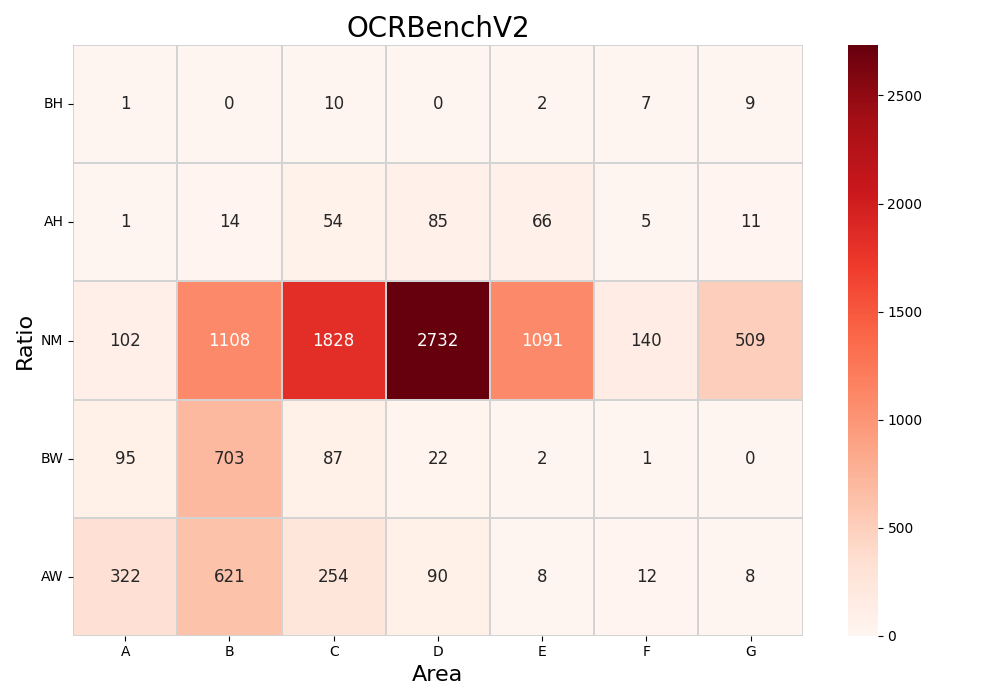}
\caption{OCRBench-v2}
\label{OCRBenchv2-11}
\end{subfigure}%

\caption{Data distribution of existing Benchmarks}
\label{fig:all_benchmarks}
\end{figure}

Here, we present a detailed analysis of the image resolution in existing benchmarks. The results are visualized in the form of a heatmap in Figure \ref{fig:all_benchmarks}, where the X-axis represents the Area \ref{Area Define}, indicating different resolution distributions ranging from small to large (labeled A through G). The Y-axis corresponds to the Ratio\ref{Ratio Define}, representing various aspect ratio distributions spanning from BW to BH. Among them, (a) is the RC-Bench proposed by ourselves.

\textbf{Area-Based Resolution Levels}

The classification of image resolution is based on the fundamental unit of $384 \times 384$ pixels. Images are categorized by their total area (width $\times$ height) into seven distinct levels (A--G), as shown below:

\begin{table}[h!]
\centering
\caption{Image resolution levels based on $384 \times 384$ pixel units}
\vspace{3pt}
\begin{tabular}{@{}lll@{}}
\toprule
\textbf{Level} & \textbf{Area Range (pixels$^2$)} & \textbf{Description} \\ \midrule
A & $1 \sim 100 \times 100$ & Very small images, e.g., icons \\
B & $100 \times 100 \sim 384 \times 384$ & Small images \\
C & $384 \times 384 \sim (2 \cdot 384)^2$ & Medium-sized images \\
D & $(2 \cdot 384)^2 \sim (3 \cdot 384)^2$ & Medium-large images \\
E & $(3 \cdot 384)^2 \sim (4 \cdot 384)^2$ & Large images \\
F & $(4 \cdot 384)^2 \sim (5 \cdot 384)^2$ & Extra-large images \\
G & $(5 \cdot 384)^2 \sim \infty$ & Ultra-large images \\ \bottomrule
\end{tabular}
\label{Area Define}
\end{table}

Each level forms a disjoint interval, ensuring:
\begin{align}
    \text{Area}_{i} \cap \text{Area}_{i+1} = \emptyset,\quad \text{Area}_{i} \cup \text{Area}_{i+1} \subseteq \mathbb{R}^{+}
\label{eq_Area}
\end{align}

\text{Aspect Ratio Categories}

Images can also be classified by their aspect ratio (width:height) into five categories:

\begin{table}[h!]
\centering
\caption{Image aspect ratio classification}
\vspace{3pt}
\begin{tabular}{@{}lll@{}}
\toprule
\textbf{Category} & \textbf{Aspect Ratio Range} & \textbf{Description} \\ \midrule
NM (Normal) & $2{:}1 \sim 1{:}2$ & Near-square or balanced images \\
AW (Almost Wide) & $4{:}1 \sim 2{:}1$ & Moderately wide images \\
BW (Very Wide) & $>\!4{:}1$ & Extremely wide images (e.g., panoramas) \\
AH (Almost High) & $1{:}2 \sim 1{:}4$ & Moderately tall images \\
BH (Very High) & $<\!1{:}4$ & Extremely tall images (e.g., vertical scrolls) \\ \bottomrule
\end{tabular}
\label{Ratio Define}
\end{table}

The aspect ratio is defined as:

\begin{align}
    \text{Aspect Ratio} = \frac{w}{h}, \quad \text{where } w,h \in \mathbb{Z}^{+}
\label{eq_Ratio}
\end{align}

\newpage

\newpage

\subsection{Training Settings of VLM Pre-Training}
\label{appendix:details}
The training details and hyperparameters for VLM pre-training are presented in Tab.~\ref{tab:mllmsetting}. We first conducted VLM pre-training on the LLaVA-Pretrain dataset (558K). This was followed by Visual Instruction Tuning, with Visual SFT 1 and Visual SFT 2 performed on the LLaVA-mix665k and LLaVA-NeXT-Data (779K) datasets, respectively. Compared to Visual SFT 1, the dataset used for Visual SFT 2 is specifically optimized for OCR tasks and further fine-tuned for the Visual Encoder. As a result, the trained model demonstrates enhanced Resolution-Centric capabilities.


\begin{table}[h]
\vspace{-10pt}
\caption{Training details and hyper-parameters for VLM pre-training and visual supervised fine-tuning.}
\vspace{3pt}
\resizebox{\columnwidth}{!}{%
\begin{tabular}{@{}l|ccc@{}}
\toprule
\textbf{Details} & Pre-Training & Visual SFT 1 & Visual SFT 2 \\ \midrule
\# Samples & 558K & 665K & 779K \\
Vision Encoder & Qwen2-VL-ViT & Qwen2-VL-ViT & Qwen2-VL-ViT \\
Visual Projector & MLP & MLP & MLP \\
LLM Backbone & Qwen2-7B-Instruct & Qwen2-7B-Instruct & Qwen2-7B-Instruct \\
Tunable Parts & Visual Projector & Visual Projector, LLM & Vision Encoder, Visual Projector, LLM \\
\# Tokens per Image & 4 $\sim$ 2048 & 4 $\sim$ 2048 & 4 $\sim$ 4096 \\
Context Length & 4096 & 4096 & 5120 \\
Sequence Packing & \checkmark & \checkmark & \checkmark \\ \midrule
Precision &  \texttt{BF16} &  \texttt{BF16} &  \texttt{BF16} \\
Global Batch Size & 128 & 32 & 32 \\
\# Training Epoch & 1 & 1 & 1 \\
\# GPUs & 8 A100-80G & 8 A100-80G & 8 A100-80G \\
LR Scheduler & linear-warmup+cosine-decay & linear-warmup+cosine-decay & linear-warmup+cosine-decay \\
Peak LR & 1e-3 & 1e-5 & 1e-5 \\
Warmup Ratio & 0.03 & 0.03 & 0.03 \\
Weight Decay & 0. & 0. & 0. \\
Vision Encoder LR & - & - & 2e-6 \\ 
\# GPU Hours & 16 & 112 & 192 \\ \bottomrule
\end{tabular}%
}
\label{tab:mllmsetting}
\end{table}

\newpage
\subsection{
More detailed Evaluation Results on RC-Bench
}
\label{sec:supp-rcbench}
\begin{figure}[h]
\centering

\begin{subfigure}{0.31\textwidth}
\centering
\includegraphics[width=\linewidth]{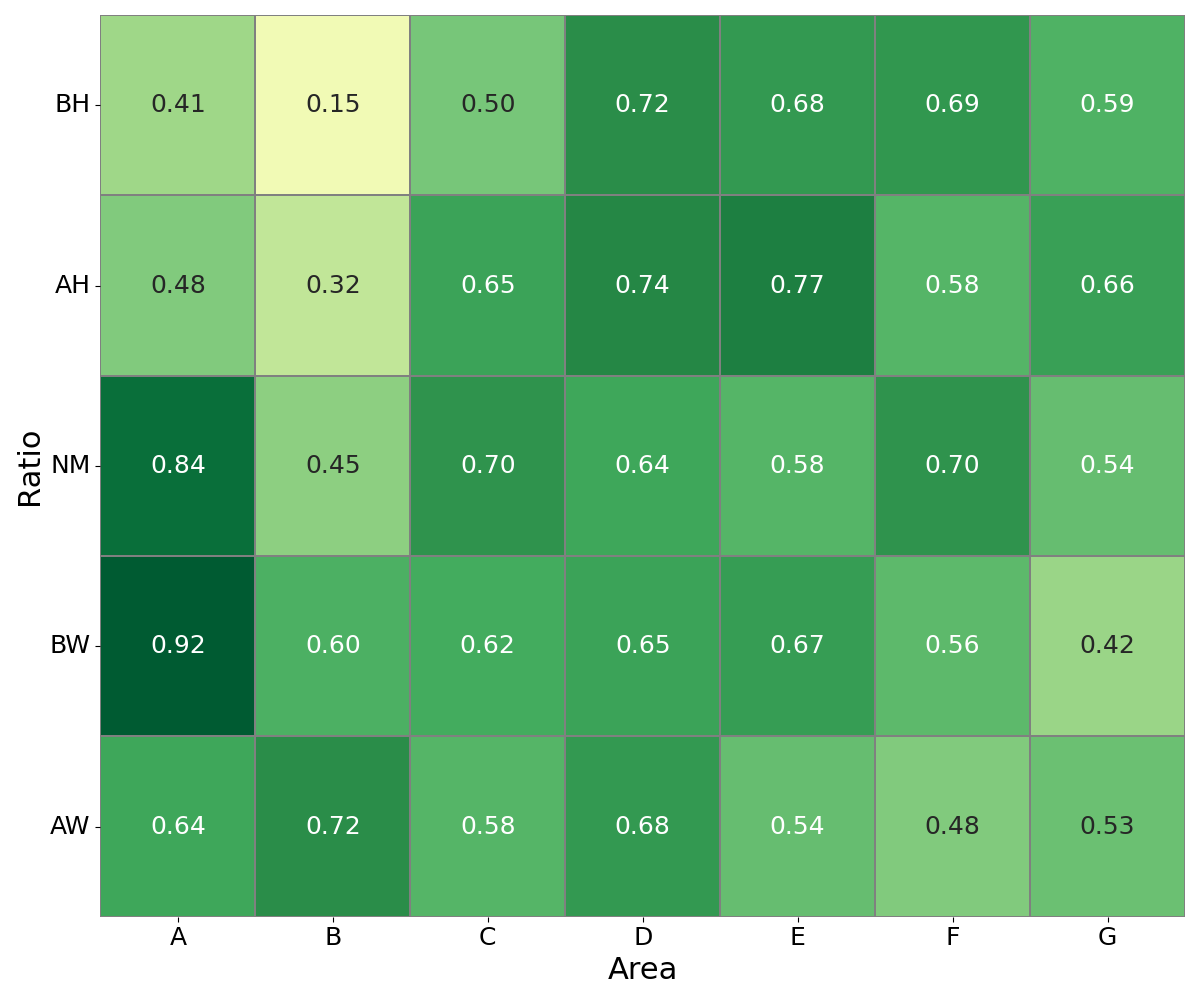}
\caption{NativeRes-LLaVA-7B}
\label{fig:nativeres-llava}
\end{subfigure}
\hfill
\begin{subfigure}{0.31\textwidth}
\centering
\includegraphics[width=\linewidth]{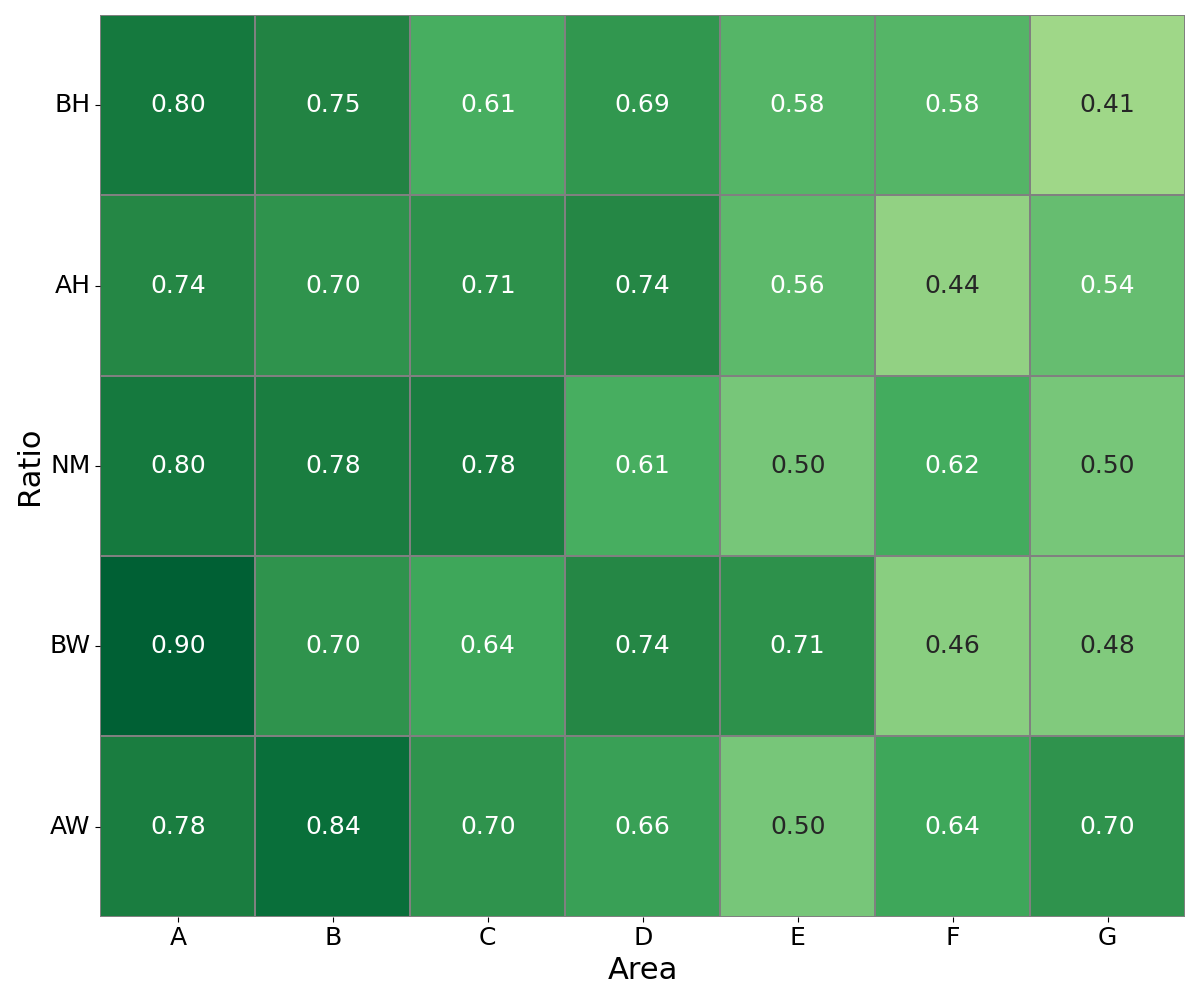}
\caption{Deepseek-VL2-A4B}
\label{fig:deepseek-vl2}
\end{subfigure}
\hfill
\begin{subfigure}{0.31\textwidth}
\centering
\includegraphics[width=\linewidth]{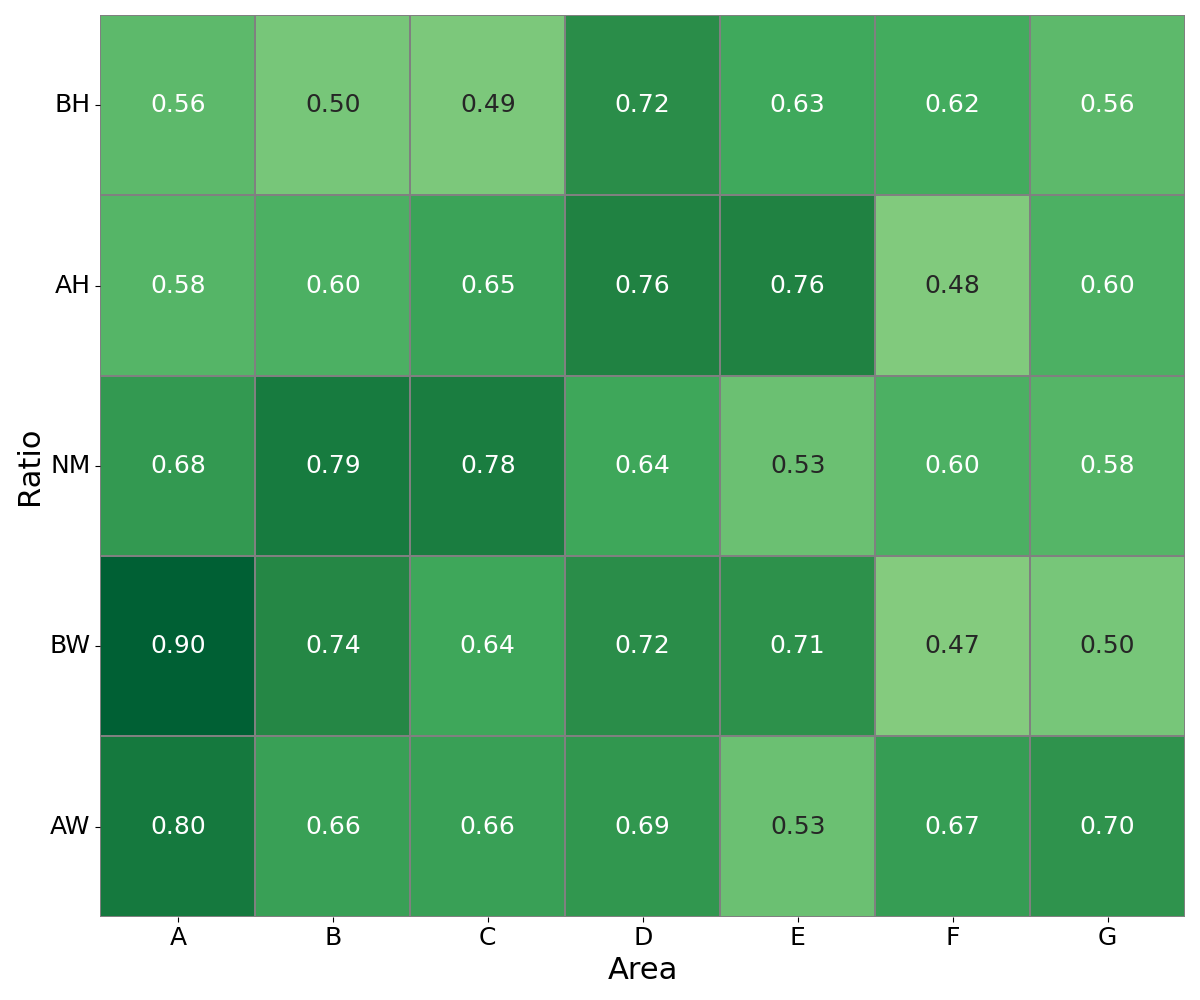}
\caption{InternVL2.5-8B}
\label{fig:InternVL2_5-8B}
\end{subfigure}

\vspace{10pt}

\begin{subfigure}{0.31\textwidth}
\centering
\includegraphics[width=\linewidth]{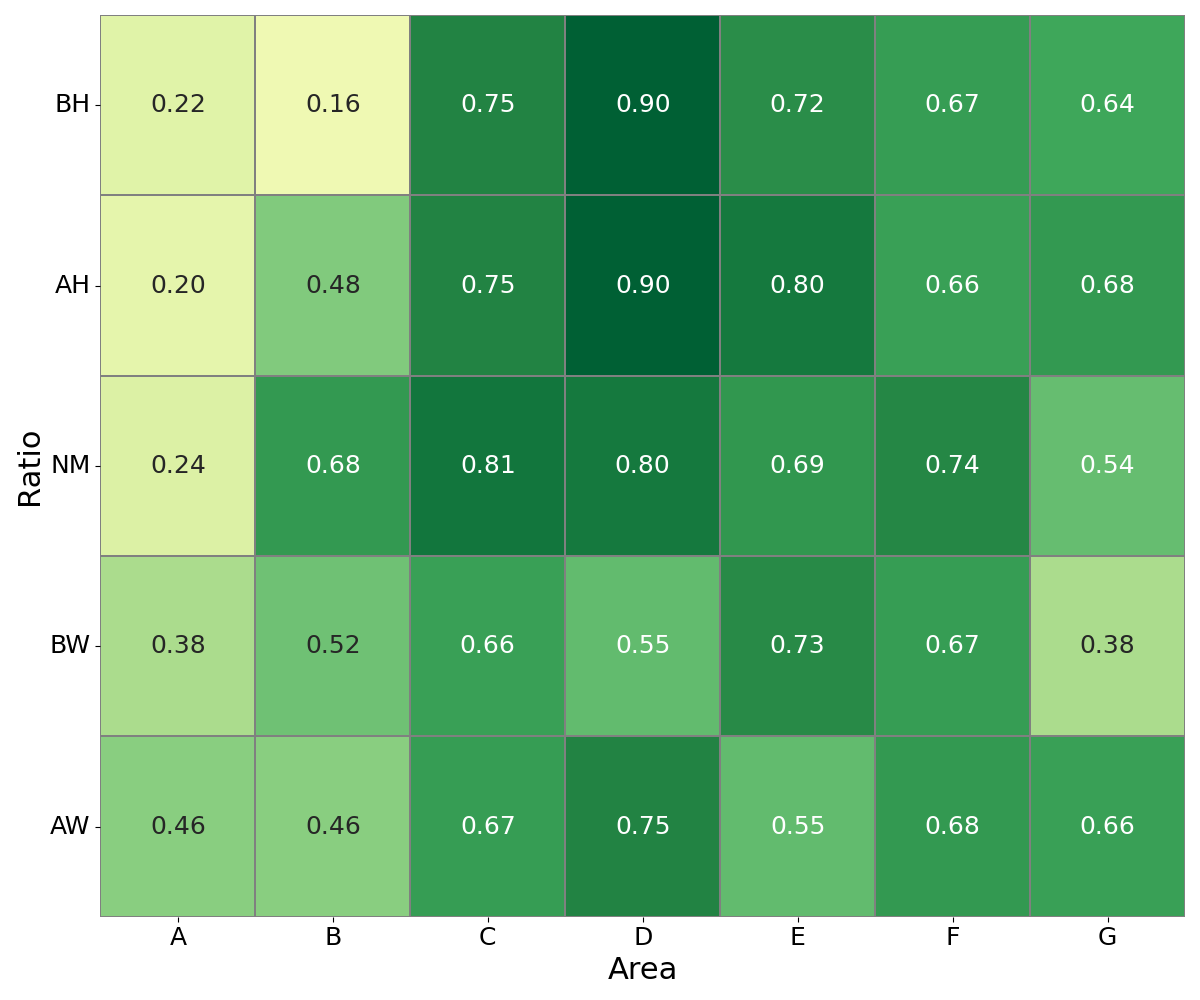}
\caption{InternVL3-8B}
\label{fig:InternVL3-8B}
\end{subfigure}
\hfill
\begin{subfigure}{0.31\textwidth}
\centering
\includegraphics[width=\linewidth]{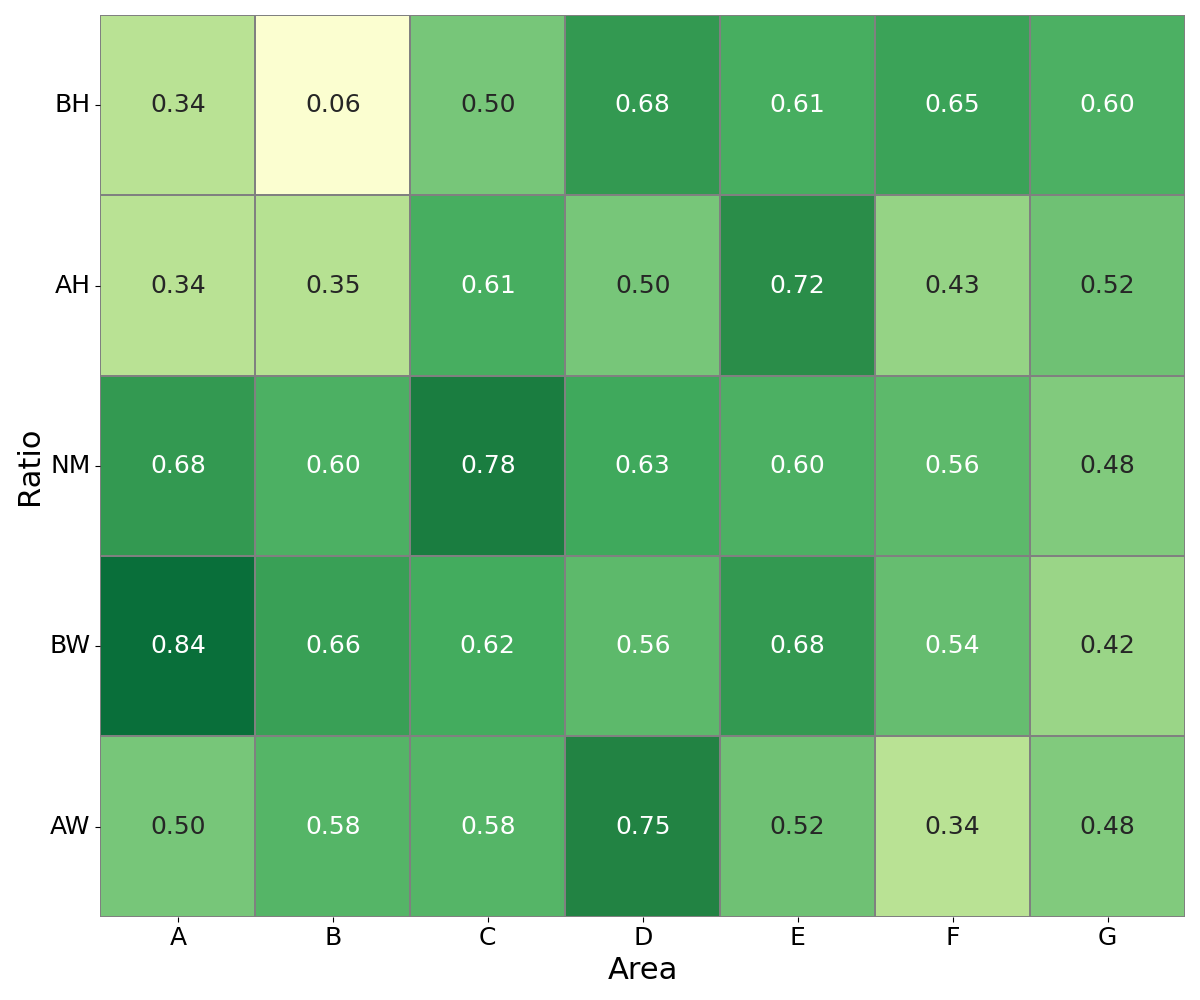}
\caption{LLaVA-OneVision-7B}
\label{fig:LLaVAOneVision}
\end{subfigure}
\hfill
\begin{subfigure}{0.31\textwidth}
\centering
\includegraphics[width=\linewidth]{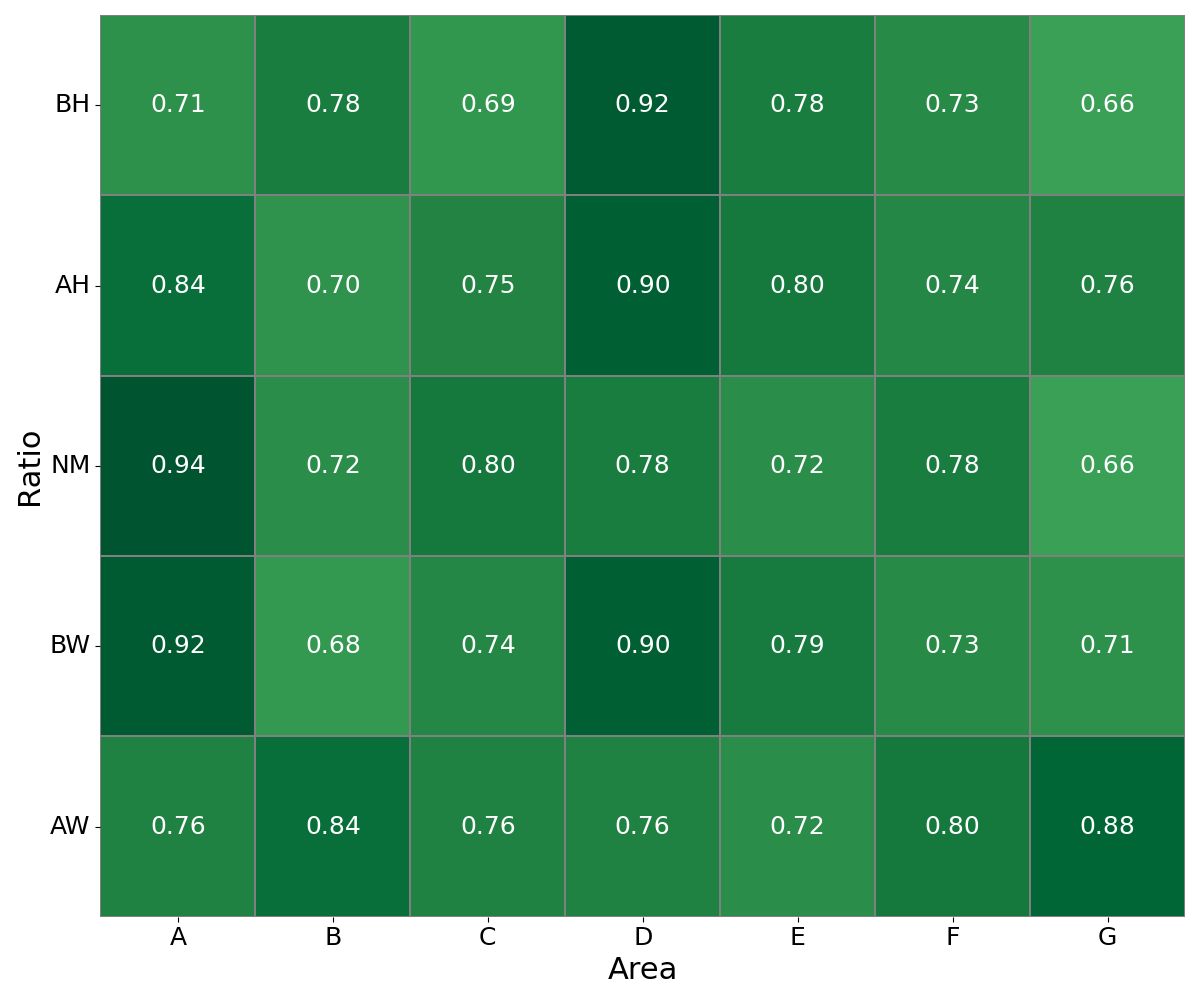}
\caption{Qwen2-VL-Instruct-7B}
\label{fig:Qwen2-VL-Instruct}
\end{subfigure}

\vspace{10pt}

\begin{subfigure}{0.31\textwidth}
\centering
\includegraphics[width=\linewidth]{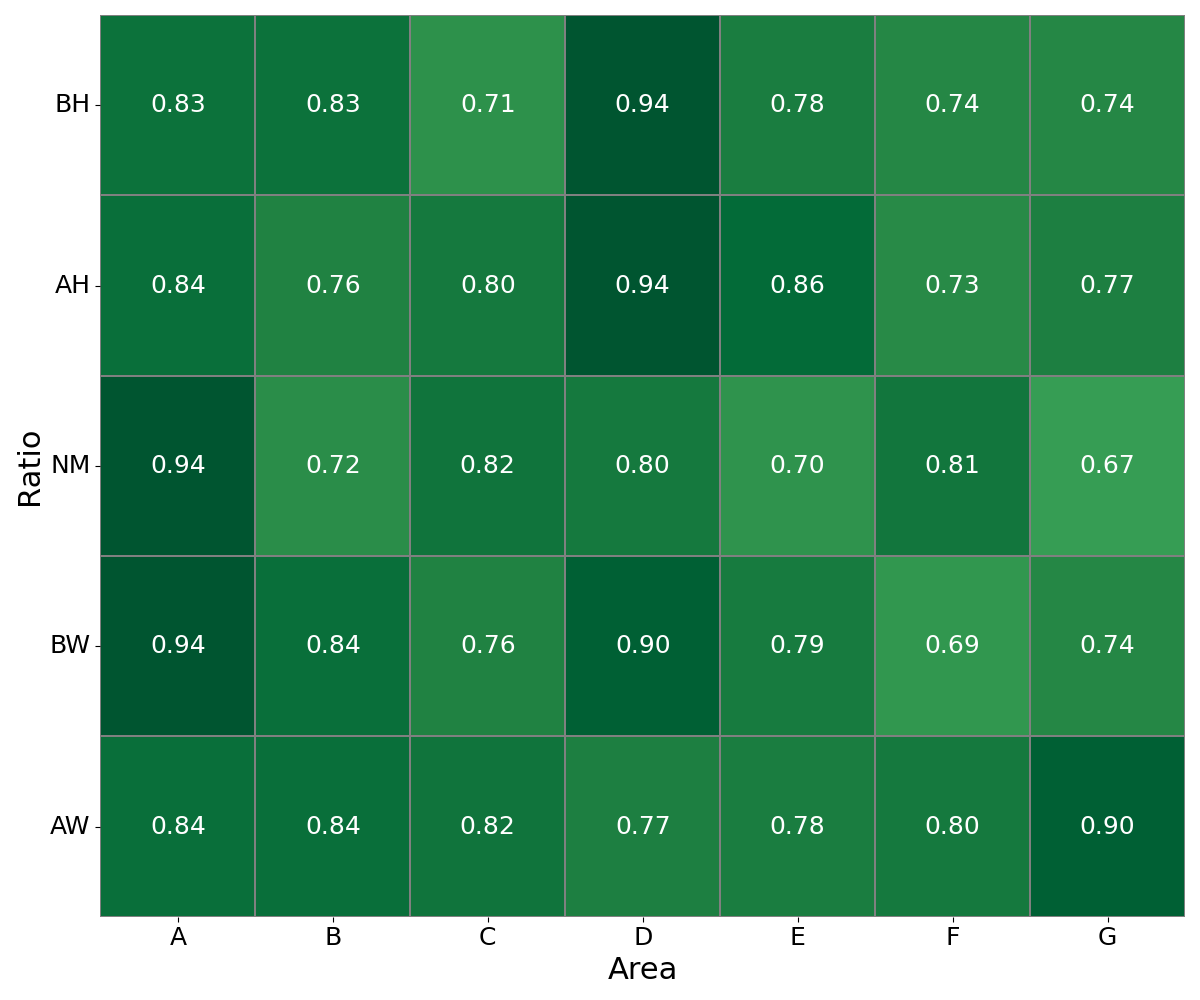}
\caption{Qwen2.5-VL-Instruct-7B}
\label{fig:Qwen2.5-VL-Instruct}
\end{subfigure}
\hfill
\begin{subfigure}{0.31\textwidth}
\centering
\includegraphics[width=\linewidth]{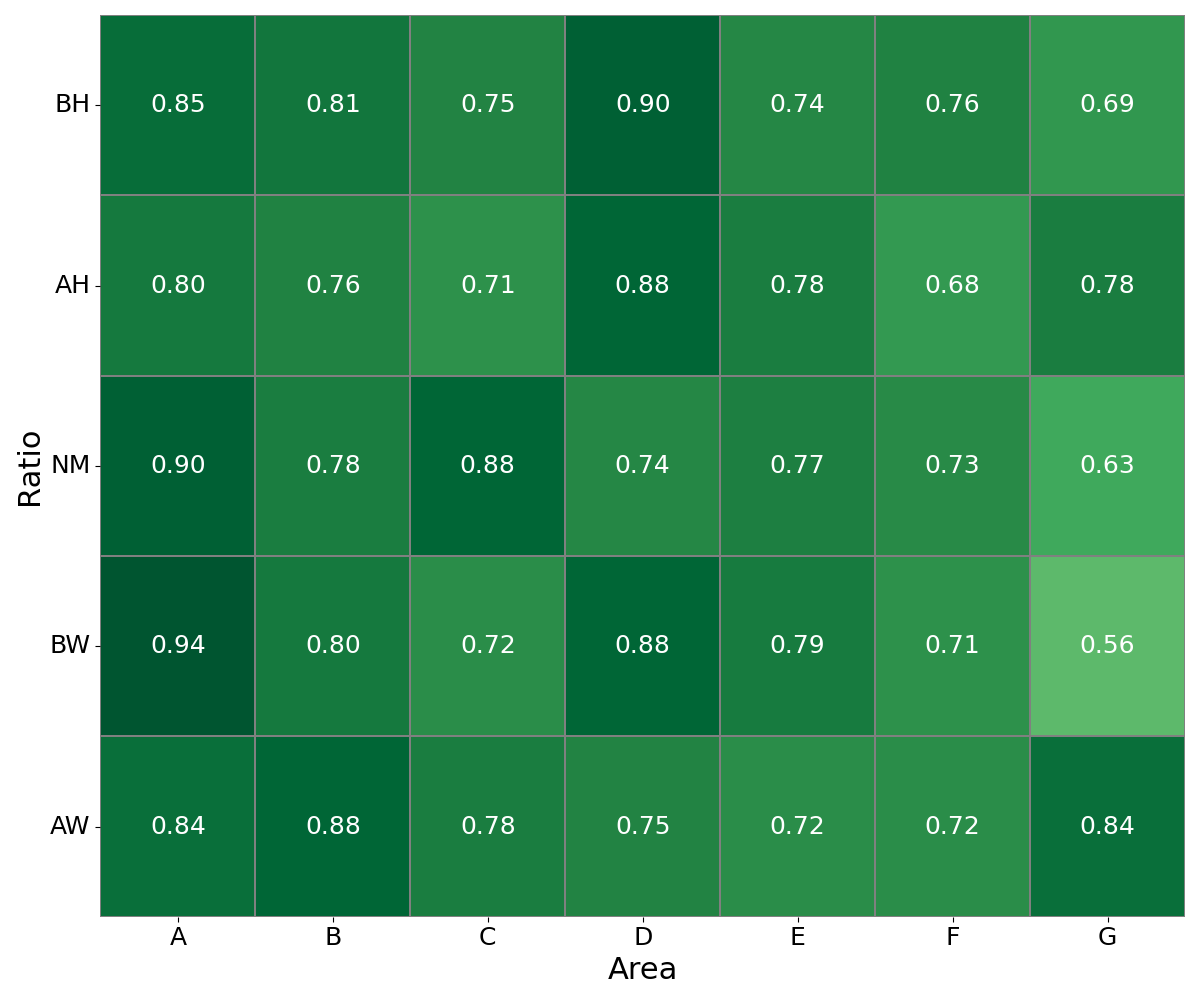}
\caption{Kimi-VL-Instruct-A3B}
\label{fig:Kimi-VL-A3B-Instruct}
\end{subfigure}
\hfill
\begin{subfigure}{0.31\textwidth}
\centering
\includegraphics[width=\linewidth]{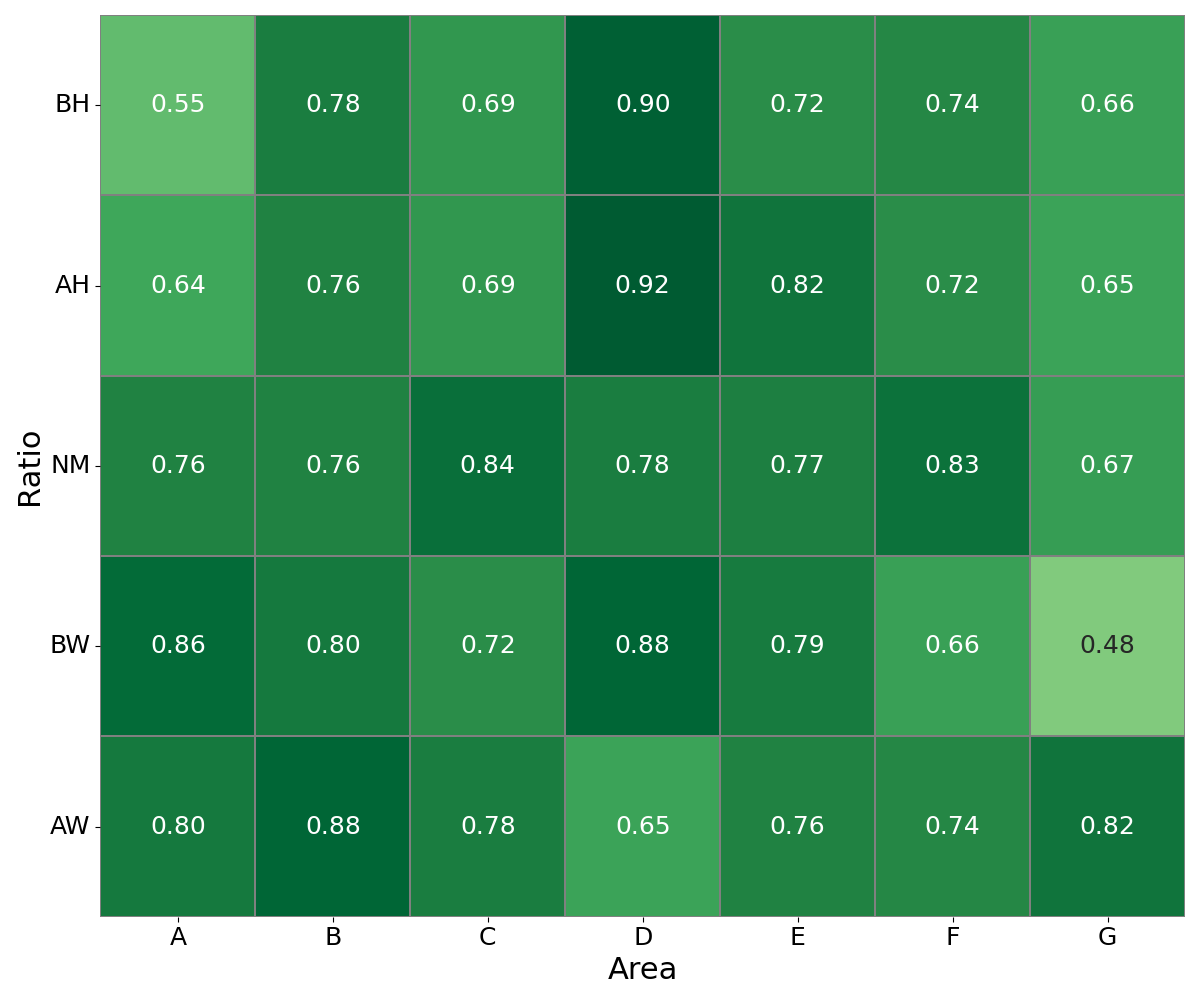}
\caption{Seed1.5-VL-A20B}
\label{fig:Seed1.5-VL}
\end{subfigure}

\vspace{10pt}

\begin{subfigure}{0.30\textwidth}
\centering
\includegraphics[width=\linewidth]{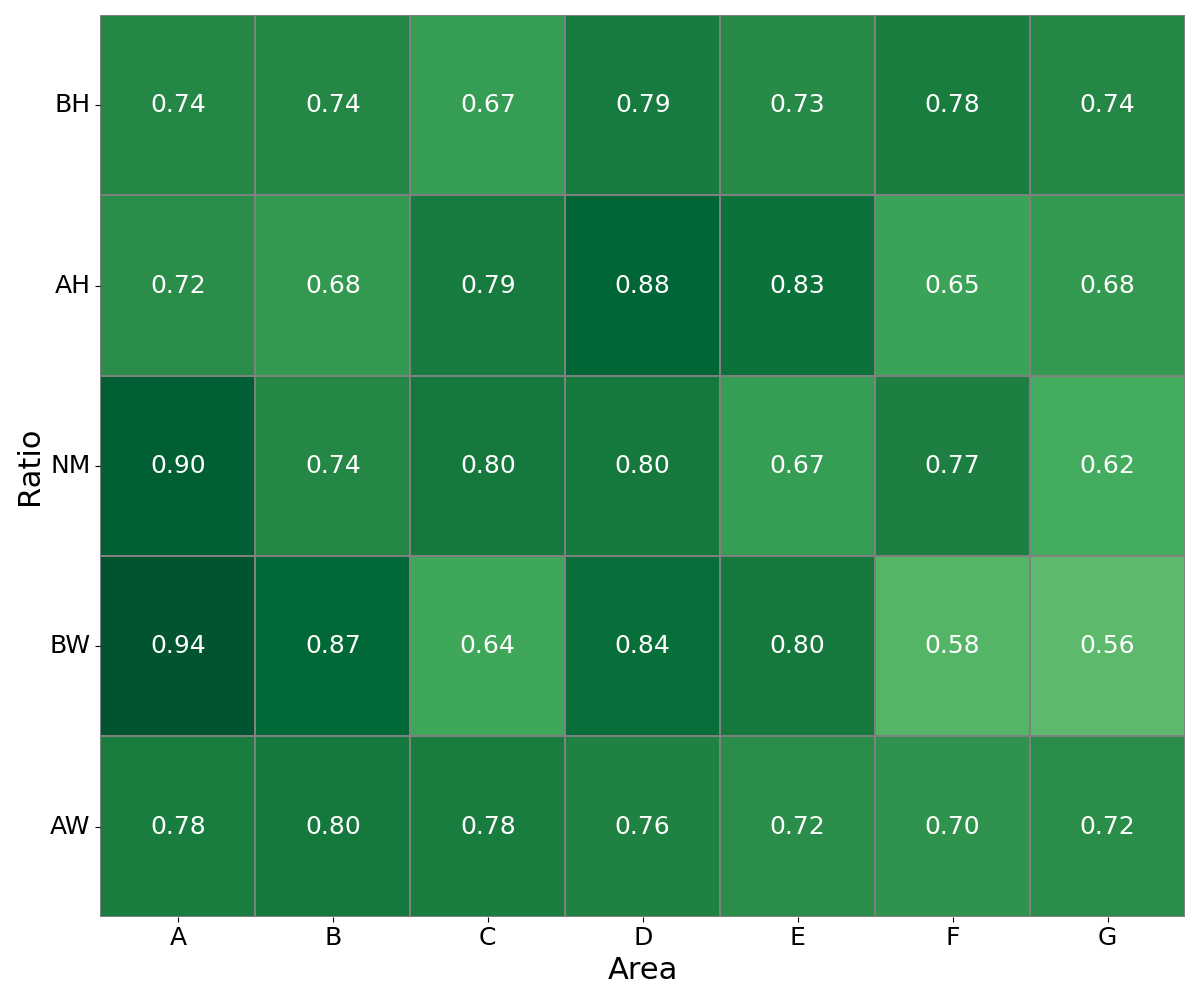}
\caption{GPT-4o}
\label{fig:GPT-4o}
\end{subfigure}%
\hfill
\begin{subfigure}{0.30\textwidth}
\centering
\includegraphics[width=\linewidth]{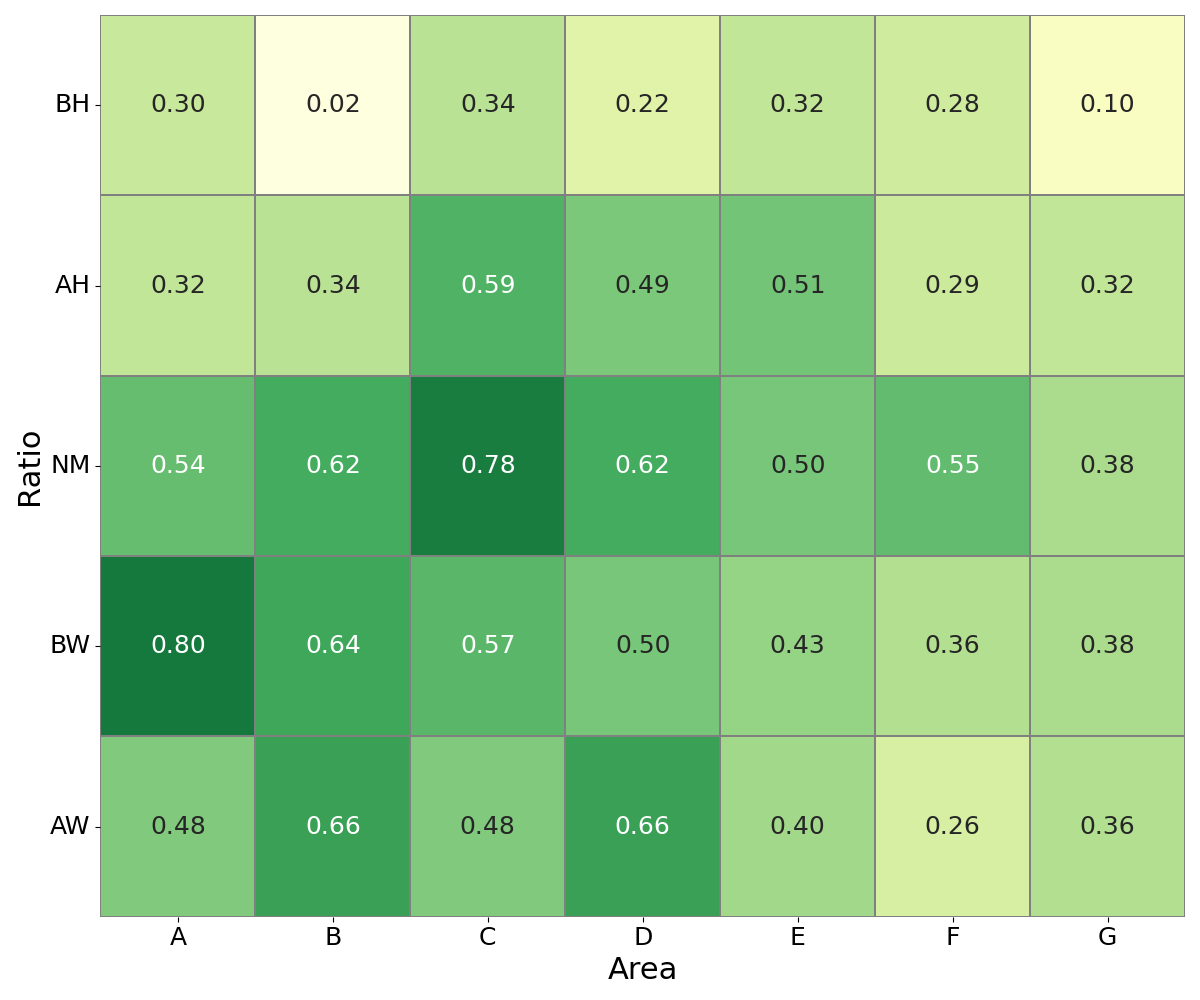}
\caption{Cambrian-8B}
\label{fig:Cambrian-8B}
\end{subfigure}

\caption{Evaluation results of vision-language models on RC-Bench.}
\label{fig:rcbench-vlm-eval}
\end{figure}

We conducted a heatmap analysis on the RC-Bench results, comparing open-source models(Figure \ref{fig:rcbench-vlm-eval}) with our models trained under limited data(Figure \ref{fig:llava-native-rcbench}). The heatmap is structured as a 7 × 5 grid, with Area on the X-axis and Ratio on the Y-axis, visually presenting the model scores across resolution and aspect ratio dimensions. This enables a more intuitive assessment of the robustness of VLMs' resolution strategies when dealing with diverse visual data.

\newpage

\begin{figure}[h]
\centering

\begin{subfigure}{0.45\textwidth}
\centering
\includegraphics[width=\linewidth]{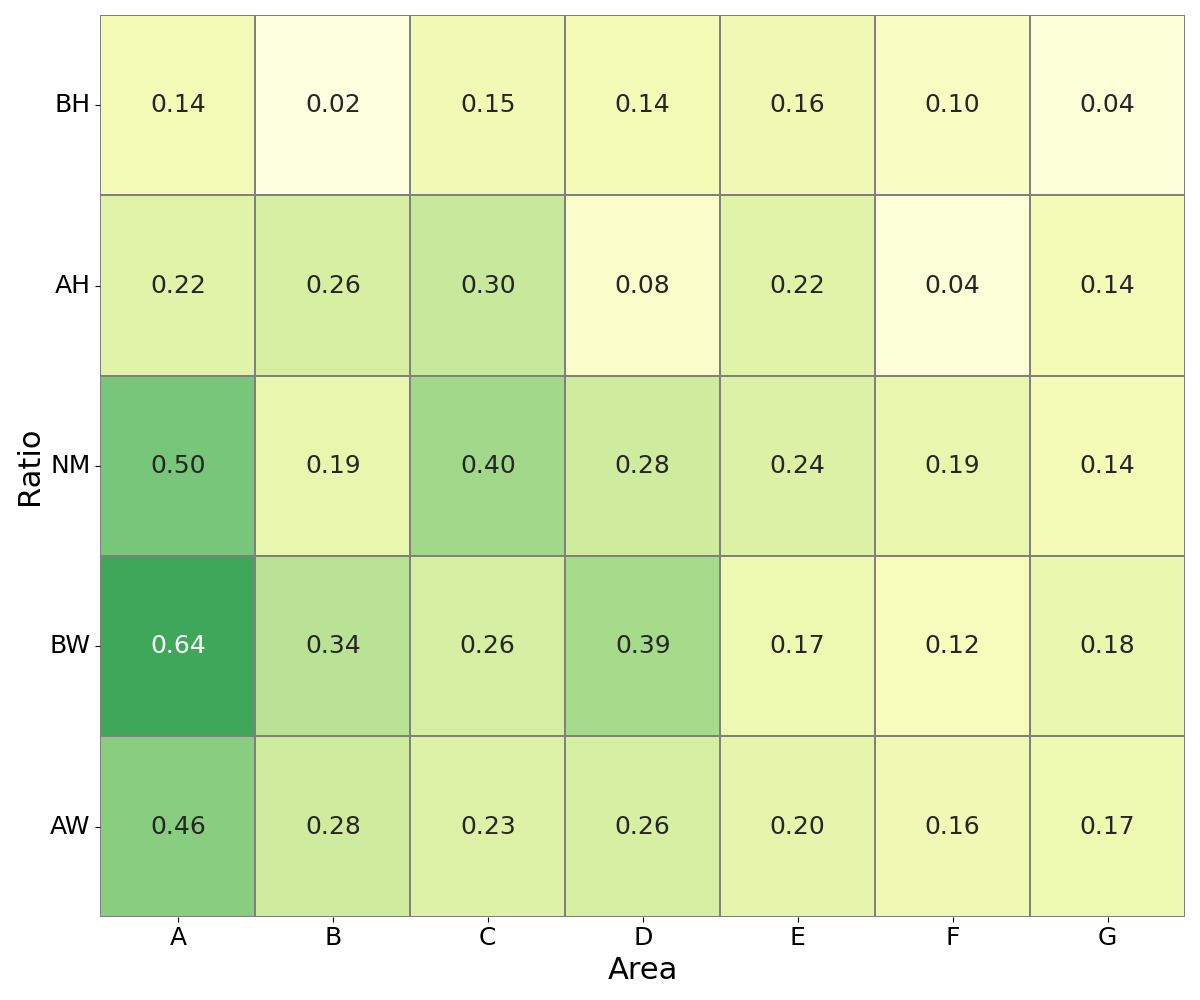}
\caption{LLaVA-NeXT-SigLip-384}
\end{subfigure}%
\hfill
\begin{subfigure}{0.45\textwidth}
\centering
\includegraphics[width=\linewidth]{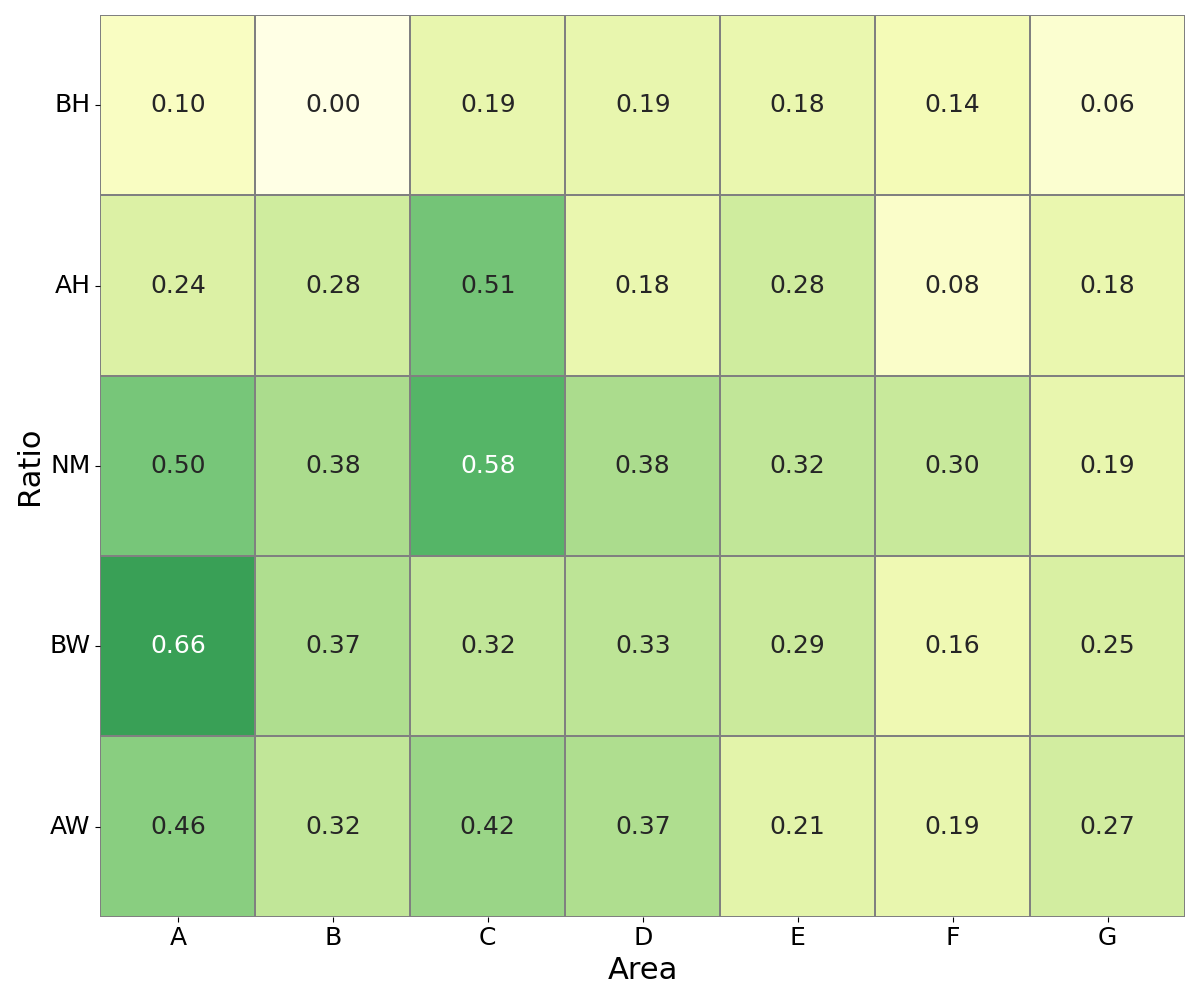}
\caption{LLaVA-NeXT-SigLip-384}
\end{subfigure}
\\[-0.5ex]
\makebox[\textwidth]{\footnotesize 768×768, Crop}

\vspace{10pt}

\begin{subfigure}{0.45\textwidth}
\centering
\includegraphics[width=\linewidth]{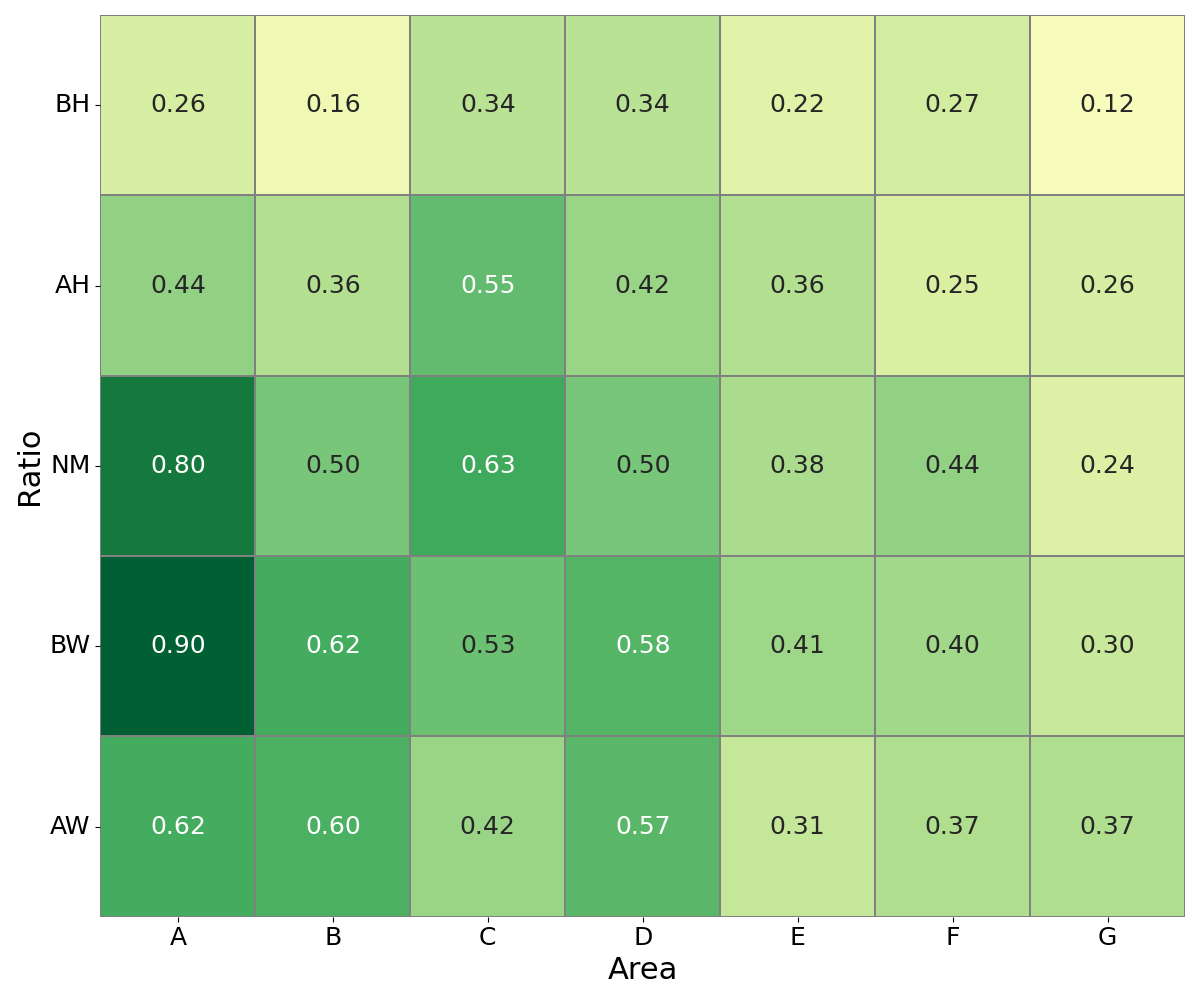}
\caption{LLaVA-NeXT-QwenViT}
\end{subfigure}%
\hfill
\begin{subfigure}{0.45\textwidth}
\centering
\includegraphics[width=\linewidth]{models/NativeResLLaVA-AnyRes.png}
\caption{LLaVA-NeXT-QwenViT}
\end{subfigure}
\\[-0.5ex]
\makebox[\textwidth]{\footnotesize 728×728, Crop}

\vspace{10pt}

\begin{subfigure}{0.45\textwidth}
\centering
\includegraphics[width=\linewidth]{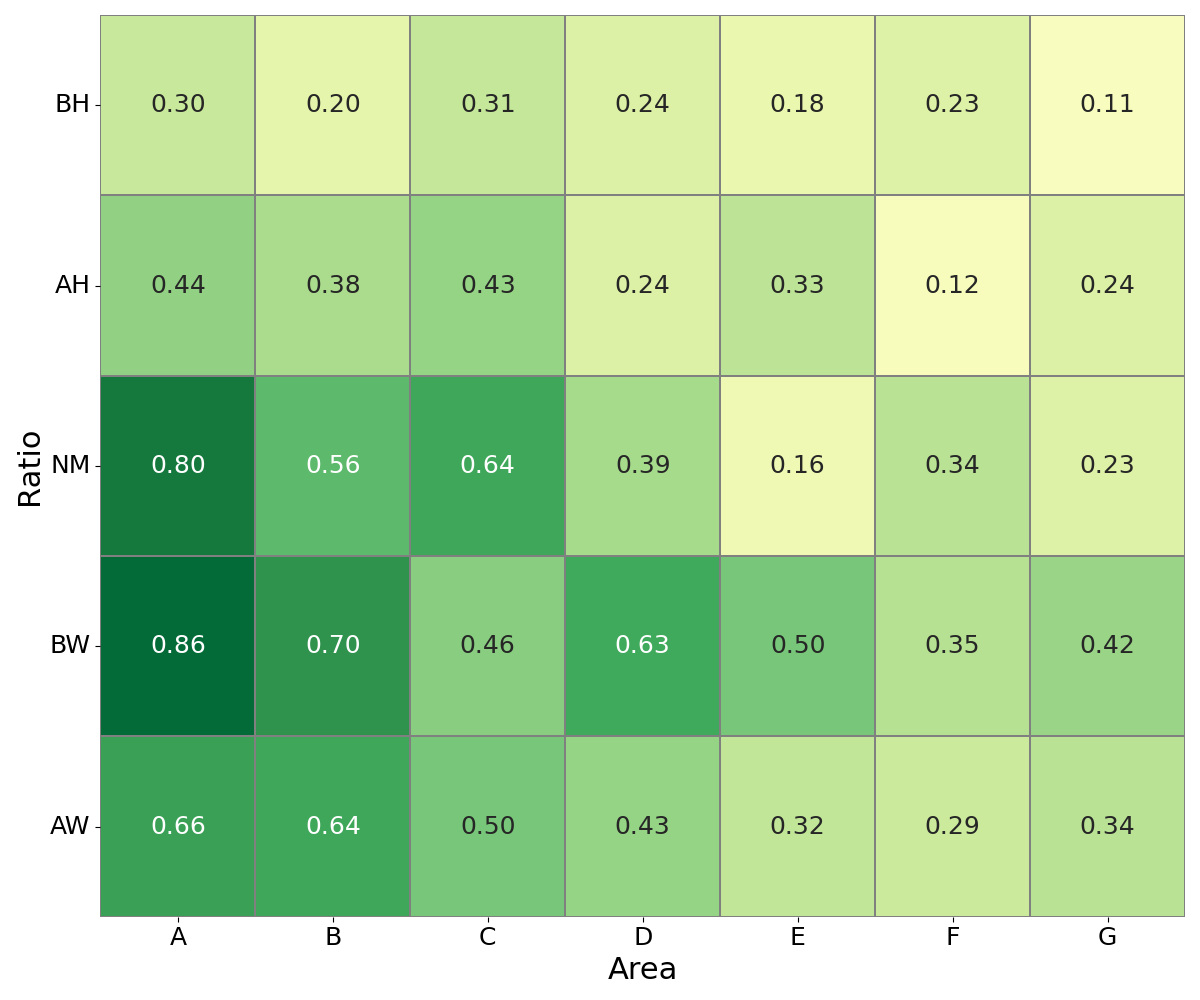}
\caption{NativeRes-LLaVA}
\end{subfigure}%
\hfill
\begin{subfigure}{0.45\textwidth}
\centering
\includegraphics[width=\linewidth]{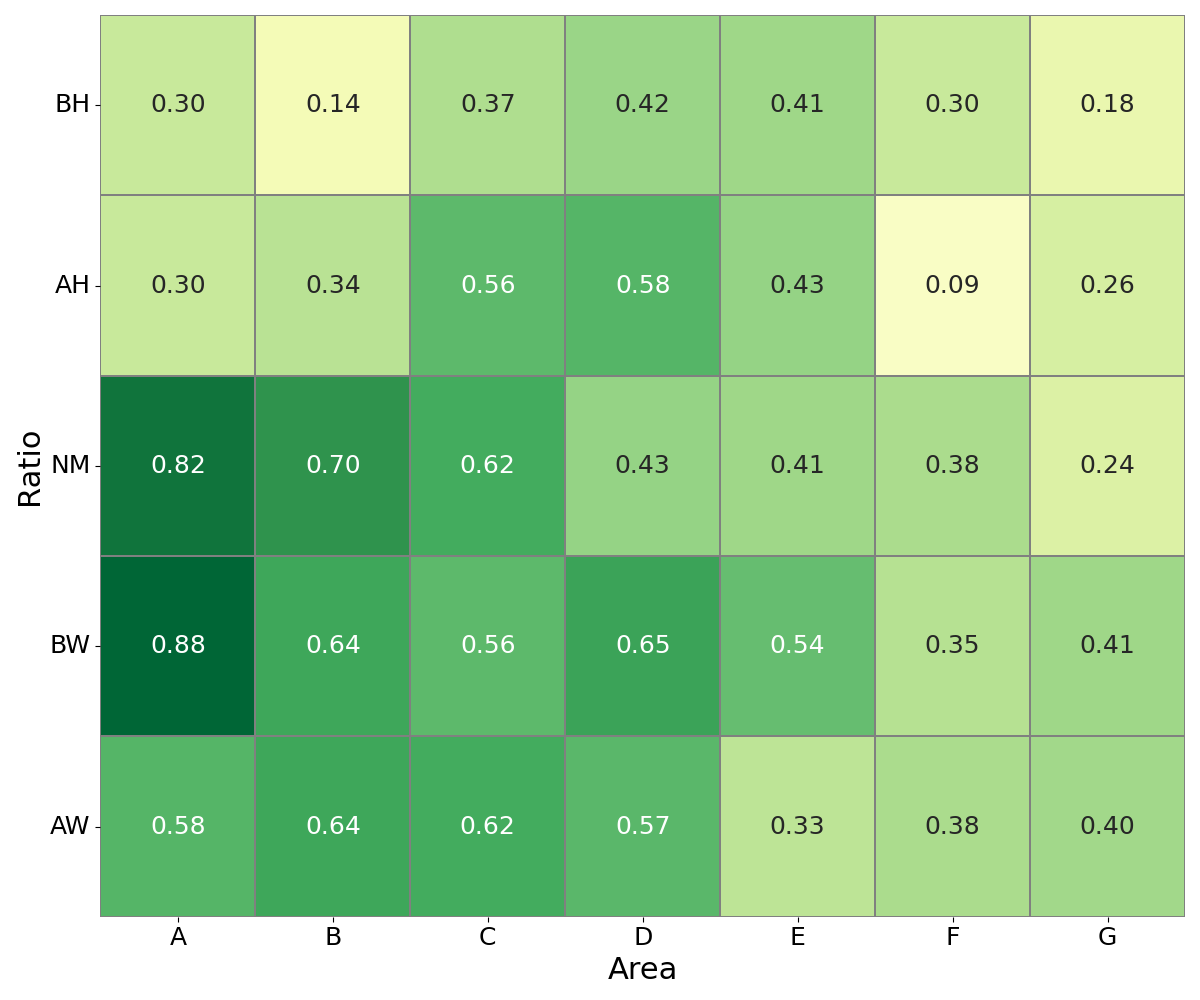}
\caption{NativeRes-LLaVA}
\end{subfigure}
\\[-0.5ex]
\makebox[\textwidth]{\footnotesize 378×378, Fixed}
\end{figure}
\vspace{10pt}

\newpage

\begin{figure}[h]
\ContinuedFloat
\centering

\begin{subfigure}{0.45\textwidth}
\centering
\includegraphics[width=\linewidth]{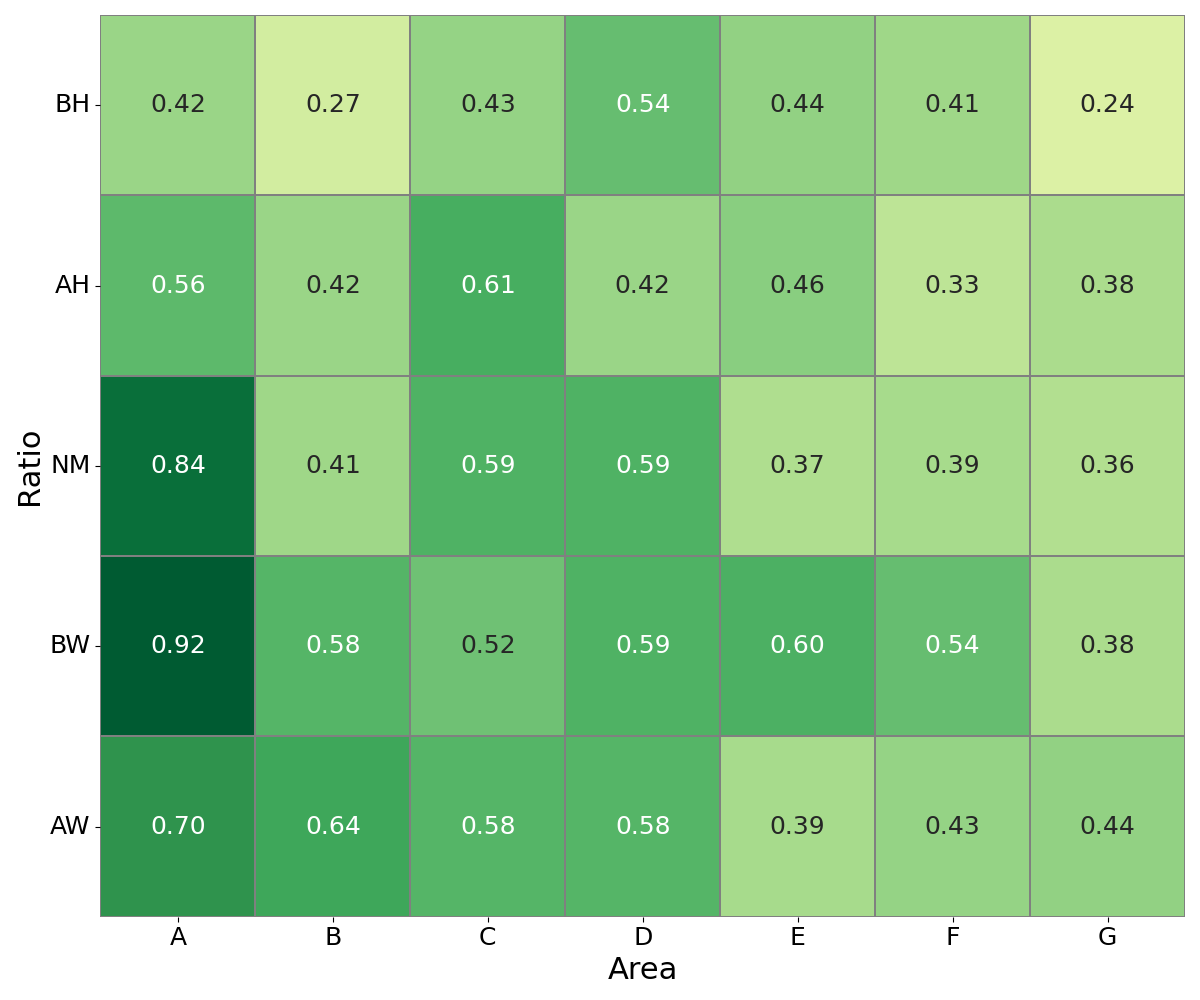}
\caption{NativeRes-LLaVA}
\end{subfigure}%
\hfill
\begin{subfigure}{0.45\textwidth}
\centering
\includegraphics[width=\linewidth]{models/4_4096_728_v2.png}
\caption{NativeRes-LLaVA}
\end{subfigure}
\\[-0.5ex]
\makebox[\textwidth]{\footnotesize 728×728, Native}

\vspace{10pt}

\begin{subfigure}{0.45\textwidth}
\centering
\includegraphics[width=\linewidth]{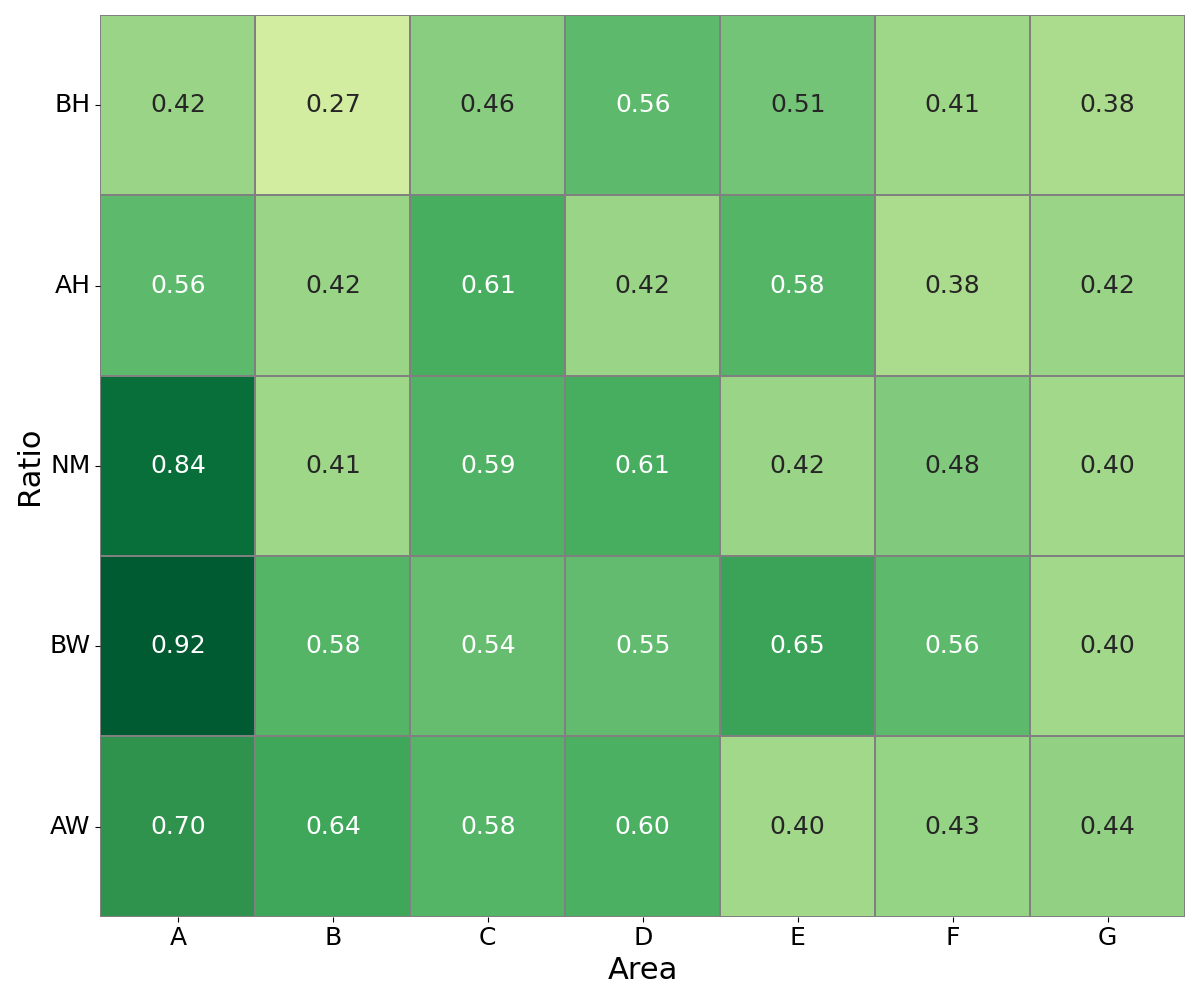}
\caption{NativeRes-LLaVA}
\parbox{\linewidth}{\centering \footnotesize 1260×1260, Native}
\end{subfigure}%
\hfill
\begin{subfigure}{0.45\textwidth}
\centering
\includegraphics[width=\linewidth]{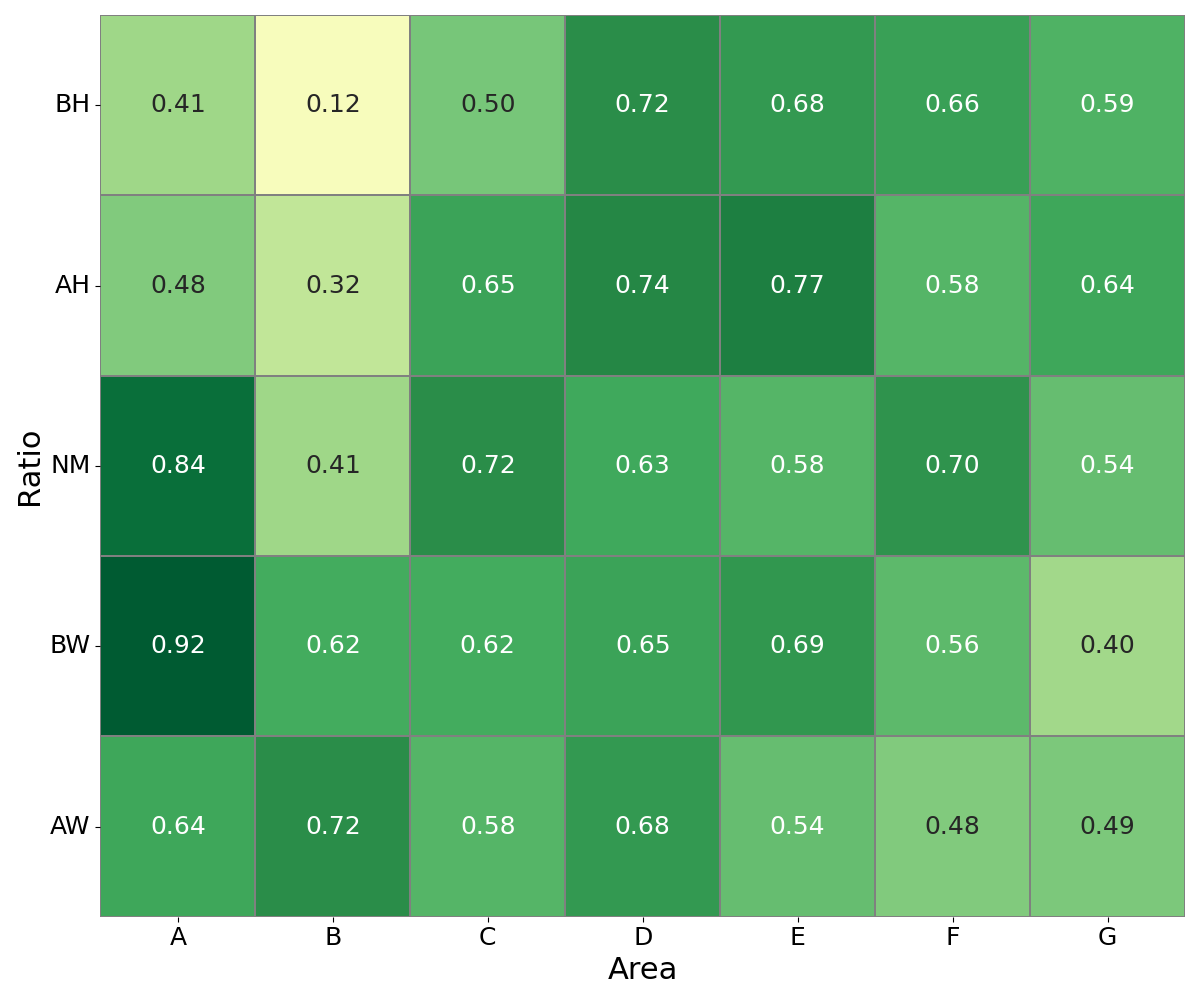}
\caption{NativeRes-LLaVA}
\parbox{\linewidth}{\centering \footnotesize 1792×1792, Native}
\end{subfigure}

\caption{Detail evaluation results of \textbf{LLaVA-NeXT} and \textbf{NativeRes-LLaVA} on RC-Bench. Training data sizes: \textbf{Left — 1.22M}, \textbf{Right — 1.34M}.}

\label{fig:llava-native-rcbench}
\end{figure}

\newpage
\subsection{
Evaluation settings in RC-Bench
}
In the evaluation of \textsc{RC-Bench}, we first adopt a heuristic approach that combines keyword-based rules with large language models (LLMs) to categorize answers into several types, including dates, numbers, identifiers, phrases, addresses, and complete sentences.

For shorter answers—such as those consisting of a single number, date, or code—we employ the \textit{Exact Match} (EM) criterion to assess the accuracy of model outputs. For longer answers, we use the \textit{Average Normalized Levenshtein Similarity} (ANLS) metric to mitigate the impact of minor textual variations on the evaluation results.

To further enhance the robustness of evaluation, we apply multiple answer normalization strategies. For answers containing unit symbols (e.g., the dollar sign ``\$'' or measurement units such as ``cm''), we extend the set of ground-truth references by including reasonable variations---for example, both ``9'' and ``9~cm'' are treated as correct. Additionally, we account for variations in the ordering of units commonly observed across different language models. If the unit appears in a semantically appropriate position, we consider the answer correct. For instance, ``193'', ``193~\$'', and ``\$~193'' are all accepted as valid answers. Moreover, to mitigate the impact of language mismatch between questions and answers, we ensure that each question is expressed in the same language as its corresponding reference answer.

\subsection{
Data Examples in RC-Bench
}
\label{sec:supp-Demo}
We provide more examples extracted from our RC-Bench. We try to cover different image categories in every Area and Aspect Ratio Range to offer a holistic overview of RC-Bench.

\newpage


\section*{Supplementary material}

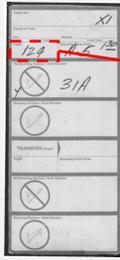 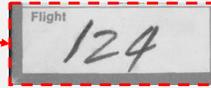

Resolution: 844 x 1783　Area: E　Ratio: AH

Q: What is the flight number on the ticket?
A: 124

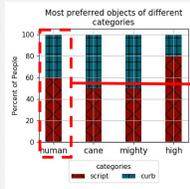 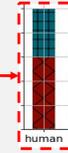

Resolution: 448 x 448　Area: C　Ratio: NM

Q: What is the first category on the x-axis?
A: human

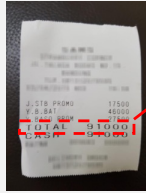 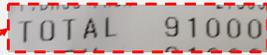

Resolution: 960 x 1280　Area: D　Ratio: NM

Q: What is the total amount on the receipt ?
A: 91000

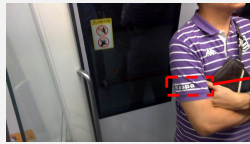 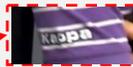

Resolution: 1280 x 720　Area: D　Ratio: NM

Q: What brand is it ?
A: Kappa

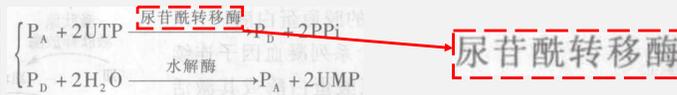

Resolution: 884 x 230　Area: C　Ratio: AW

Q: 图中第一个反应的催化酶是什么？
A: 尿苷酰转移酶

# DesignWrite
INCORPORATED

**IMPORTANT FAX MESSAGE**
**PLEASE DELIVER IMMEDIATELY**

**TO:** Dr. Rose Miketta          **FAX NUMBER:** 847-470-6717

**FROM:** Fran Karo, PhD          **DATE/TIME:** Feb 2, 1998
DesignWrite, Inc.

**NUMBER OF PAGES:** 2
(including cover sheet)

**RE:** Cycloserine technical report

Dear Dr. Miketta,

As a follow-up to my conversation with your assistant, I have enclosed the actual reference of the technical report that I am trying to get a copy of. I misinformed your assistant: I did not find the reference on MedLine, but in a 1996 review article published in Drugs & Aging titled: Cognitive Enhancers in Age-Related Cognitive Decline. I would appreciate any help you could give me in finding this report.

Thank you for your assistance.

Sincerely,

Fran Karo, PhD
Medical Writer

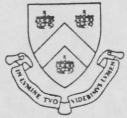

# Columbia University School of Law

ADMISSIONS OFFICE/435 WEST 116 STREET/NEW YORK, NEW YORK 10027-7297

## INSTRUCTOR'S APPRAISAL

Applicant's Name: Kristen W. McNutt     Class of 19____
(PLEASE TYPE OR PRINT)

Applicant's Address: 3800 N. Lake Shore Drive, Apt. 12D
Chicago, IL  60613

Instructor's Name: William J. Darby, M.D., Ph.D.     Academic Title: Professor and Chairman (Emeritus)
(e.g., "Professor," "Associate Professor," etc.)

TO THE APPLICANT: PLEASE TYPE OR PRINT YOUR NAME AND ADDRESS AND THE NAME AND TITLE OF THE INSTRUCTOR TO WHOM YOU WILL BE FORWARDING THIS FORM, ABOVE, AND YOUR NAME AND SOCIAL SECURITY NUMBER ON THE REVERSE SIDE. THIS FORM SHOULD BE GIVEN TO AN INSTRUCTOR WHO KNOWS YOU WELL AND WHO WILL BE WILLING TO EVALUATE YOUR POTENTIAL AS A LAW STUDENT. ALSO, PLEASE ADDRESS THE CERTIFICATION POSTCARD (USING INSTITUTIONAL OR CORPORATE ADDRESS) SO THAT YOUR REFEREE WILL KNOW THAT THE APPRAISAL WAS RECEIVED INTACT.

As stated in the "Instructions for Completing Your Application," recent legislation affords to matriculants the right of access to appraisals submitted in support of their applications if the school retains these appraisals. You may choose to waive this right. Although it is entirely your decision as to whether or not you waive your right of access to such appraisals, it is possible that some persons may be more guarded and less candid in their evaluations if they know you may read them than if you waive your right and they know that their appraisals will remain confidential. For this reason we are providing you with the option on this appraisal form of waiving this right if you so desire. Please note that your decision concerning this option will not affect our consideration of your application. If you choose to waive your right of access to this evaluation, please sign your name on the line below.

_____     _____
Date                          Applicant's Signature

TO THE INSTRUCTOR: The student named above is applying to the School of Law of Columbia University. Under recent legislation, if the applicant's signature does not appear on the above line, this form will be accessible to the applicant if the applicant matriculates and if the form is retained after matriculation. (Please see TO THE APPLICANT, above.)

If you prefer not to use the form on the reverse side, or if you wish to supplement the form, you may make your comments in any fashion you choose. Please note, however, that if the applicant has chosen to waive his or her right of access to your appraisal, this form or a facsimile, signed by the applicant, must accompany your appraisal in order for us to assure you of confidentiality.

In selecting students for the entering class, it is our objective to accept only those who are likely to reflect credit upon the institutions from which they have come by their achievements here and at the Bar. Your evaluation of this applicant, therefore, will assist us in passing upon his or her application.

If you wish to add anything to your evaluation, please write to us directly.

Please check the enclosed postcard to be certain that it can be sent to you at your institutional or corporate address.

We sincerely thank you for your cooperation.

James M. Milligan
Assistant Dean
Director of Admissions

9/81

Source: https://www.industrydocuments.ucsf.edu/docs/gmnm0227

**61.1% OF DOGS**

refused to accept a treat from that stranger

Q: What percentage of dogs refuse to accept treat from stranger?
A: 61.1%

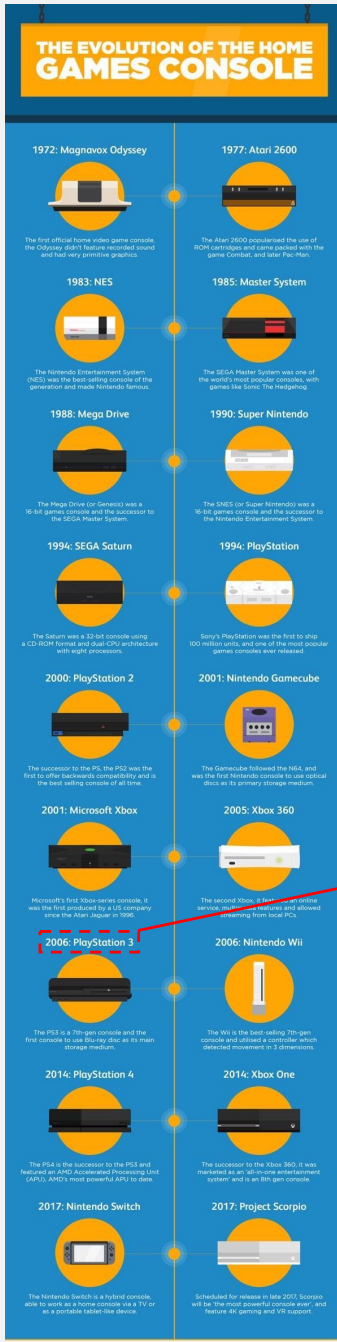

2006: PlayStation 3

Q: What is the name of the game console released in 2006, shown on the left ?
A: PlayStation 3



\end{document}